\newif\ifisTR

\isTRtrue

\documentclass{article} 

\usepackage{fullpage}

\usepackage{bm}                 
\usepackage{algorithm}          
\usepackage{algpseudocode}  

\usepackage[table]{xcolor} 

\usepackage{microtype}
\usepackage{graphicx}
\usepackage{booktabs} 
\usepackage{array}
\usepackage{placeins}

\usepackage{subcaption}
\usepackage{tcolorbox}
\usepackage{nicefrac}  
\usepackage{enumitem}
\usepackage{hyperref}
\usepackage{multirow}
\usepackage{colortbl}
\usepackage{xspace}
\usepackage{wrapfig}
\usepackage{amsmath}
\usepackage{amssymb}
\usepackage{mathtools}
\usepackage[export]{adjustbox}
\usepackage{amsthm}
\usepackage{color}
\usepackage{xcolor}
\usepackage{tikz}
\usepackage{float}
\usepackage{adjustbox}
\usepackage{tabularx}

\usepackage[capitalize,noabbrev]{cleveref}

\usepackage{overpic}

\theoremstyle{plain}

\definecolor{mygray}{gray}{0.85}
\definecolor{LightBlue}{cmyk}{0.06, 0.03, 0.01, 0.0}

\usepackage[textsize=tiny]{todonotes}

\usepackage[round,semicolon,sort]{natbib}

\usepackage{amsmath,amsfonts,bm}

\newcommand{\F}{\mathrm{F}}

\def\eqref#1{equation~\ref{#1}}

\def\1{\bm{1}}

\def\vxi{{\bm{\xi}}}

\def\vv{{\bm{v}}}

\def\mA{{\bm{A}}}
\def\mB{{\bm{B}}}
\def\mC{{\bm{C}}}

\def\mG{{\bm{G}}}

\def\mI{{\bm{I}}}

\def\mM{{\bm{M}}}

\def\mO{{\bm{O}}}

\def\mQ{{\bm{Q}}}

\def\mU{{\bm{U}}}
\def\mV{{\bm{V}}}
\def\mW{{\bm{W}}}
\def\mX{{\bm{X}}}
\def\mY{{\bm{Y}}}
\def\mZ{{\bm{Z}}}

\def\mLambda{{\bm{\Lambda}}}
\def\mSigma{{\bm{\Sigma}}}

\DeclareMathAlphabet{\mathsfit}{\encodingdefault}{\sfdefault}{m}{sl}
\SetMathAlphabet{\mathsfit}{bold}{\encodingdefault}{\sfdefault}{bx}{n}

\usepackage{mathrsfs}

\newcommand{\E}{\mathbb{E}}

\newcommand{\R}{\mathbb{R}}

\def\mLambda{{\mathbf{\Lambda}}}
\def\mSigma{{\mathbf{\Sigma}}}

\newtheorem{theorem}{Theorem}[section]

\newtheorem{lemma}[theorem]{Lemma}

\newtheorem{definition}[theorem]{Definition}
\newtheorem{assumption}[theorem]{Assumption}

\usepackage[textsize=tiny]{todonotes}
\definecolor{LightBlue}{HTML}{F3F3F3}

\usepackage[capitalize,noabbrev]{cleveref}

\newcommand{\ShowComments}{no}

\ifthenelse{\equal{\ShowComments}{yes}}{
\newcommand{\LEI}[1]{{\color{orange}LEI: #1}}
\newcommand{\yaoqing}[1]{\textcolor{orange}{[Yaoqing:\ #1]}}
\newcommand{\shuhua}[1]{\textcolor{red}{[Shuhua:\ #1]}}

\newcommand{\shenyang}[1]{\textcolor{red}{[Shenyang:\ #1]}}
\newcommand{\tianyu}[1]{{\color{blue}TIANYU: #1}}
\newcommand{\addressedyaoqing}[1]{{\color{cyan}[(Addressed) Yaoqing: #1]}}
\newcommand{\addressedshuhua}[1]{{\color{pink}[(Addressed) Shuhua: #1]}}
\newcommand{\addressedshenyang}[1]{\textcolor{pink}{[(Addressed) Shenyang:\ #1]}}
}{
\newcommand{\LEI}[1]{}
\newcommand{\yaoqing}[1]{}
\newcommand{\shuhua}[1]{}
\newcommand{\tianyu}[1]{}
\newcommand{\shenyang}[1]{}
\newcommand{\addressedyaoqing}[1]{}
\newcommand{\addressedshuhua}[1]{}
\newcommand{\addressedshenyang}[1]{}
}

\renewcommand{\cite}[1]{\citep{#1}}
\title{HTMuon: Improving Muon via Heavy-Tailed Spectral Correction}
\date{}

\author{
 Tianyu Pang$^*$\textsuperscript{1},
 Yujie Fang$^*$\textsuperscript{2},
 Zihang Liu\textsuperscript{3,4},
 Shenyang Deng\textsuperscript{1},
 Lei Hsiung\textsuperscript{1},
\\
 Shuhua Yu\textsuperscript{5},
 Yaoqing Yang\textsuperscript{1}
\\
 \textsuperscript{1}Dartmouth College \\
 \textsuperscript{2}Microsoft \\
 \textsuperscript{3}International Computer Science Institute \\
 \textsuperscript{4}University of California, Berkeley \\
 \textsuperscript{5}Meta \\
}

\begin{document}
\maketitle
\def\thefootnote{*}\footnotetext{Equal contribution.}
\begin{abstract}
\texttt{Muon} has recently shown promising results in LLM training. In this work, we study how to further improve \texttt{Muon}. We argue that \texttt{Muon}'s orthogonalized update rule suppresses the emergence of heavy-tailed weight spectra and over-emphasizes the training along noise-dominated directions. Motivated by the Heavy-Tailed Self-Regularization (HT-SR) theory, we propose \texttt{HTMuon}. \texttt{HTMuon} preserves \texttt{Muon}'s ability to capture parameter interdependencies while producing heavier-tailed updates and inducing heavier-tailed weight spectra. Experiments on LLM pretraining and image classification show that \texttt{HTMuon} consistently improves performance over state-of-the-art baselines and can also serve as a plug-in on top of existing \texttt{Muon} variants. For example, on LLaMA pretraining on the C4 dataset, \texttt{HTMuon} reduces perplexity by up to $0.98$ compared to \texttt{Muon}. We further theoretically show that \texttt{HTMuon} corresponds to steepest descent under the Schatten-$q$ norm constraint and provide convergence analysis in smooth non-convex settings. The implementation of \texttt{HTMuon} is available at  \url{https://github.com/TDCSZ327/HTmuon}.
\end{abstract}
\section{Introduction}
Optimizers play a central role in training Large language models (LLMs). A well-designed optimizer can help LLMs learn more effectively from large-scale datasets \citep{semenov2025benchmarking}, leading not only to lower training loss but also to improved generalization \citep{foret2020sharpness,kaddour2022flat}.
Over the decades, optimizers have often been developed from a \emph{vector-based} view, where gradients are treated as flattened vectors during updates. Within this framework, \texttt{Adam} \citep{kingma2014Adam} and \texttt{AdamW} \citep{loshchilov2017decoupled} have become the default choices for training LLMs \citep{groeneveld2024olmo, olmo20242}. A key reason is that they incorporate first- and second-order moment information in an element-wise manner, allowing them to adapt to the different statistical patterns (e.g., local geometry of the loss landscape) across individual coordinates. However, an overly element-wise update scheme often ignores interdependencies among coordinates (e.g., geometric correlations among parameters), which can limit the effectiveness of the optimizer. Although recent work \cite{zhou2023temperaturebalancinglayerwiseweight,liu2024model, zhang2024Adam, wang2025sharpness} has gone beyond \texttt{Adam} and its variants by using layer-wise or module-wise learning rates to capture parameters' relationships, whether these adjustments are sufficient to capture the complex coupling across these parameters remains to be seen.

Recently, \texttt{Muon} \citep{jordan2024muon} has been proposed as a representative \emph{matrix-based} optimizer\footnote{\emph{Matrix-based} optimizers use matrix-valued preconditioners, which can better capture geometry-induced dependencies among different parameters.}. Building on earlier matrix-based optimizers such as \texttt{AdaGrad} \citep{duchi2011adaptive} and \texttt{Shampoo} \citep{gupta2018shampoo}, \texttt{Muon} performs preconditioning on the momentum matrix, and its update can be interpreted as an orthogonalization step, which effectively captures geometry interdependencies among parameters. Moreover, the orthogonalization update can be viewed as the steepest descent under the Schatten$-\infty$ norm constraint \citep{bernstein2024modular,pethick2025training}, which leads to a competitive convergence rate and improved stability \citep{shen2025convergence, chen2025muon,  ma2026preconditioningbenefitsspectralorthogonalization}.
\texttt{Muon} has shown promising results \citep{liu2025muon, shah2025practical, wen2025fantastic, wang2025muon} and has been adopted in large-scale LLM training, including Moonshot's Kimi K2 \citep{team2025kimi} and GLM-4.5 \citep{zeng2025glm}.

However, some studies have found that the improved performance of \texttt{Muon} is inversely proportional to the model scale and the number of training steps \citep{wen2025fantastic, semenov2025benchmarking}. The orthogonalization step in \texttt{Muon} update sets all singular values of the momentum matrix to one, which means it assigns the same weight to each singular-vector direction of the updates. Although this may help \texttt{Muon} decrease the loss faster in the early phase of training \citep{shen2025convergence}, it is well known that directions associated with small singular values tend to be more noise-dominated \citep{sharma2023truth, chen2024distributional, liu2025lift, defilippis2025scaling}. In Section~\ref{sec:Unit Singular Values},  we conduct experiments to show that using uniform weights on all singular vector directions may be suboptimal.

More importantly, the orthogonalization step makes the spectrum of the momentum update matrix light-tailed. This, in turn, makes the spectrum of learned weight matrices light-tailed.
Here, the spectrum refers to the \emph{empirical spectral density} (ESD) of the weight matrices.
However, a line of work by \citet{martin2021implicit, martin2021predicting} shows that \emph{well-trained} neural networks---namely, networks that have extracted strong correlations in their weights through learning from data---tend to have heavy-tailed ESDs in their weight matrices \citep{kothapalli2025spikes, hodgkinson2025models}.
Within a reasonable range, the degree of heavy-tailedness is strongly correlated with model quality \citep{martin2021predicting}.
Motivated by these observations, they proposed the Heavy-Tailed Self-Regularization theory (HT-SR), which has been supported by extensive empirical studies and further theoretical analyses \citep{yang2023test,martin2025setol,defilippis2025scaling}. In Section~\ref{sec:Muon_makes_less_HT}, we empirically show that the \texttt{Muon} optimizer leads to a less heavy-tailed weight ESD, and it performs worse than variants of \texttt{Muon} that induce more heavy-tailed spectra.
Based on this, we claim that the \texttt{Muon} update rule, which sets all singular values of the momentum matrix to one, limits the final quality that the model can achieve.

Motivated by this observation, we propose a matrix-based optimizer \texttt{HTMuon}. \texttt{HTMuon} aims to make \texttt{Muon}'s momentum  update more heavy-tailed, which can lead to more heavy-tailed weight ESDs, while preserving \texttt{Muon}'s advantage in capturing interdependencies among parameters.
The design of \texttt{HTMuon} is simple: we raise the singular values of the momentum matrix to the power of $p$, where $p \in (0,1)$. See Algorithm~\ref{algo:HTMUON}. Setting $p=1$ reduces the method to \texttt{SGDM} \citep{sutskever2013importance}, whose updates are independent across parameters.
In contrast, $p=0$  recovers \texttt{Muon}, which, as discussed earlier, leads to less heavy-tailed updates. Therefore, we choose $p \in (0,1)$ in \texttt{HTMuon} to maintain the \emph{matrix-based} ability to model parameter coupling, while producing updates that are more heavy-tailed than those of \texttt{Muon}. Unless otherwise specified, we use $p=0.125$ for \texttt{HTMuon} throughout. We discuss the specific choice of 
$p$ in Section~\ref{sec:ablation study}.
To evaluate the empirical performance of \texttt{HTMuon}, we study a range of models and tasks, including training LLaMA models \citep{touvron2023llama} on the C4 dataset, GPT-2 models on the OpenWebText dataset and training image classification models using the ResNet \citep{he2016deep} and ViT \citep{dosovitskiy2020image} architectures. All of these tasks are standard and have been widely used in prior literature \citep{he2025alphadecay,zhao2024galore, yuan2024mars}.
We compare \texttt{HTMuon} against commonly used baseline optimizers such as \texttt{Adam}, \texttt{Muon}, \texttt{Cautious} \citep{liang2024cautious}, \texttt{GaLore}~\citep{zhao2024galore}, \texttt{Sophia}~\citep{liu2024sophiascalablestochasticsecondorder}, \texttt{Mars} \citep{yuan2024mars}, \texttt{SOAP} \citep{vyas2024soap}, \texttt{COSMOS} \citep{liu2025cosmos}, and show that \texttt{HTMuon} not only achieves superior performance but can also serve as a plug-in module on top of existing \texttt{Muon} variants like \texttt{NorMuon} \citep{li2025normuon}, \texttt{AdaMuon} \citep{si2025adamuon} to further improve test performance.
Moreover, our analysis reveals that \texttt{HTMuon} produces update and weight matrices that are more heavy-tailed than those of \texttt{Muon} during training, consistent with prior work on HT-SR theory. Finally, we present theoretical results showing that \texttt{HTMuon} is equivalent to the steepest-descent method under the Schatten$-q$ norm constraint, generalizing \texttt{Muon}’s equivalence to steepest descent under a Schatten$-\infty$ norm constraint, this  provides a matrix-version of \texttt{pbSGD} \citep{zhou2020pbsgd} and also places \texttt{HTMuon} within the linear minimization oracle framework \citep{pethick2025training, bernstein2024modular}.
We also provide a convergence analysis, showing that \texttt{HTMuon} matches the sample complexity upper bound of \texttt{Muon} and \texttt{SGDM} in smooth non-convex settings. 

To summarize, the main contributions of our work are the following:

\begin{itemize}
    \item We argue that \texttt{Muon}'s orthogonalized update rule biases updates toward noise-dominated directions and suppresses the emergence of heavy-tailed ESDs in the model's weight matrices, thereby limiting generalization performance according to HT-SR theory.
    
    \item We propose \texttt{HTMuon}, an optimizer that makes \texttt{Muon} updates more heavy-tailed while preserving \texttt{Muon}'s strength in modeling parameter interdependence. We demonstrate \texttt{HTMuon}'s effectiveness in improving performance on both LLM pretraining and image classification tasks relative to state-of-the-art optimizers including \texttt{Cautious} \citep{liang2024cautious},  \texttt{Mars} \citep{yuan2024mars}, \texttt{SOAP} \citep{vyas2024soap}, \texttt{NorMuon} \citep{li2025normuon}, \texttt{AdaMuon} \citep{si2025adamuon}, \texttt{COSMOS} \citep{liu2025cosmos}, etc.
    For example, \texttt{HTMuon} reduces perplexity by $0.92$ compared to \texttt{Muon} when training LLaMA-60M on the C4 dataset, and by $0.98$ when training LLaMA-135M on C4. Moreover, \texttt{HTMuon} can be used as an add-on method in combination with existing \texttt{Muon} variants to achieve further improvements. We also design two accelerated implementations of \texttt{HTMuon}. Using these implementations, we show that \texttt{HTMuon} outperforms \texttt{Muon} on LLaMA-1B, highlighting its potential for training large-scale models.

    \item We present theoretical results analyzing \texttt{HTMuon}. Specifically, we show that it is equivalent to steepest descent under a Schatten$-q$ norm constraint, and we provide a convergence analysis demonstrating that \texttt{HTMuon} matches the sample-complexity upper bounds of \texttt{Muon} and \texttt{SGDM} in nonconvex smooth settings. These results support \texttt{HTMuon}'s competitive convergence rate and improved training stability. 
\end{itemize}

\section{Related Work}\label{sec:related_work}
\paragraph{Pretraining Optimizers.} A strong optimizer is critical for successful LLM training \citep{zhang2024working,zhang2025pretrained}. For a long time, \texttt{Adam} \citep{kingma2014Adam} and \texttt{AdamW} \citep{loshchilov2017decoupled} have been the dominant choices. In recent years, a number of methods have emerged that build on the \texttt{Adam} family: some focus on reducing gradient variance (e.g., \texttt{MARS} \citep{yuan2024mars} and \texttt{Cautious} \citep{liang2024cautious}), while others target memory efficiency (e.g., \texttt{Lion} \cite{chen2023symbolic}, \texttt{Adam-mini} \citep{zhang2024Adam} and \texttt{GaLore} \citep{zhao2024galore}). 
However, these \texttt{Adam} variants remain \emph{vector-based} optimizers and have difficulty capturing interdependencies among parameters. Some works \citep{zhou2023temperaturebalancinglayerwiseweight,wang2025sharpness} extend \texttt{Adam} by using layer- or module-wise learning rates to capture relationships among parameters at a coarse level. Recently, \texttt{Muon} \citep{jordan2024muon} has been proposed as a representative \emph{matrix-based} optimizer. It uses the momentum matrix as a preconditioner, which better captures interdependencies among parameters and improves training stability. \texttt{Muon} has shown promising results in LLM training \citep{shah2025practical, liu2025muon, wen2025fantastic}.  Following this line, several \texttt{Muon} variants have emerged, including \texttt{NorMuon} \citep{li2025normuon} and \texttt{AdaMuon} \citep{si2025adamuon}. Among these variants, \citet{qi2026delving} is a concurrent work that also investigates a spectral family of transformations, which is closely related to \texttt{HTMuon}.

\paragraph{Heavy-tailed Phenomenon.} Heavy tailed weight spectrum in machine learning have been observed and studied in various forms. Among the most prominent empirical findings are a series of works by \citet{martin2021implicit, martin2021predicting}, which propose the HT-SR theory for deep neural networks. They observe that well-trained networks often exhibit heavy-tailed spectra in their weight matrices, and the degree of heavy-tailedness is strongly correlated with model quality. Furthermore, they quantify this behavior via power-law fits to the ESDs of weight matrices. From a theoretical perspective, a number of studies \citep{simsekli2020hausdorff,hodgkinson2022generalization,simon2023eigenlearning,dandi2024random,kothapalli2025spikes,defilippis2025scaling, deng2026suspicious} have used tools such as stochastic differential equations, random matrix theory, eigenlearning framework, and approximate message passing to characterize power-law behavior in weight ESDs and its strong connections to generalization bounds and test loss. Empirically, treating HT-SR theory as a diagnostic tool, subsequent work has explored layer-wise hyperparameter scheduling \citep{liu2024model, he2025alphadecay, zhou2023temperaturebalancinglayerwiseweight} and layer-wise pruning \citep{lu2024alphapruning,hueigenspectrum} based on the power-law fit to improve training performance and efficiency. In this work, motivated by HT-SR theory, we use \texttt{HTMuon} to make \texttt{Muon} updates more heavy-tailed to improve training.

\section{Motivation for \texttt{HTMuon}}
In this section, we first briefly introduce HT-SR theory and then use two examples to motivate the importance of heavy-tailed spectra for the design of HTMuon.

\subsection{Background on HT-SR Theory}
The HT-SR theory \citep{martin2021implicit,martin2021predicting} offers a principled lens for analyzing the ESD of neural network weight matrices. Empirically, well-trained models tend to exhibit more heavy-tailed ESDs, and the degree of heavy-tailedness correlates with training quality. Based on prior work \citep{martin2021implicit,zhou2023temperature}, heavy-tailedness is quantified by fitting a power law (PL) to the ESD and using the resulting PL exponent $\alpha$ as the metric.

Given a network with $L$ layers and weight matrices $\{\mW_l\}_{l=1}^L$of shape $n\times m$, we compute the ESD by extracting the eigenvalues of the correlation matrix $\mX_l=\mW_l^{\top}\mW_l$ for each module. We fit a power law (PL) to the ESD in the form
$
p(\lambda)\propto \lambda^{-\alpha},\quad \lambda_{\min}<\lambda<\lambda_{\max},
$
where $p(\lambda)$ denotes the eigenvalue density within the specified range. 
The PL exponent $\alpha$ serves as a proxy for the degree of heavy-tailedness, 
smaller $\alpha$ indicates a more heavy-tailed weight ESD, and  more heavy-tailed spectra are often associated with better model quality.

\subsection{Unit Singular Values May Be Suboptimal}
\label{sec:Unit Singular Values}
In this subsection, we introduce an intriguing empirical finding: for \texttt{Muon} implementation, the numerically accelerated version by Newton Schulz, which we call \texttt{Muon\_NS} (see Algorithm~\ref{algo:MUON}) can outperform the theoretically exact implementation \texttt{Muon\_SVD} (see Algorithm~\ref{algo:MUON_SVD}). 

\begin{figure*}[!htb]
    \centering
    \begin{subfigure}[b]{0.315\linewidth}
        \centering
        \includegraphics[width=\linewidth]{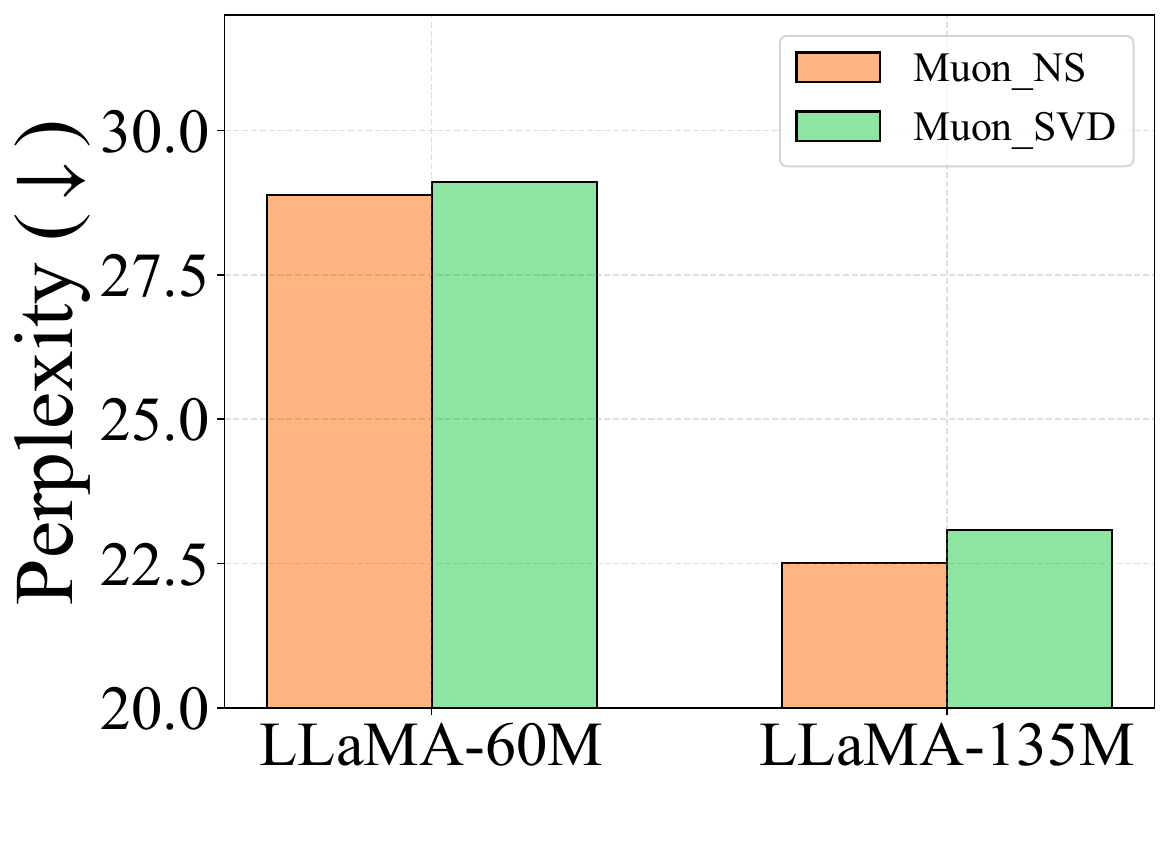}
        \caption{\shortstack{\texttt{Muon\_NS} vs \texttt{Muon\_SVD}}}
        \label{fig: Muon_ns_vs_Muon_SVD}
    \end{subfigure}
    \begin{subfigure}[b]{0.32\linewidth}
        \centering
        \includegraphics[width=\linewidth]{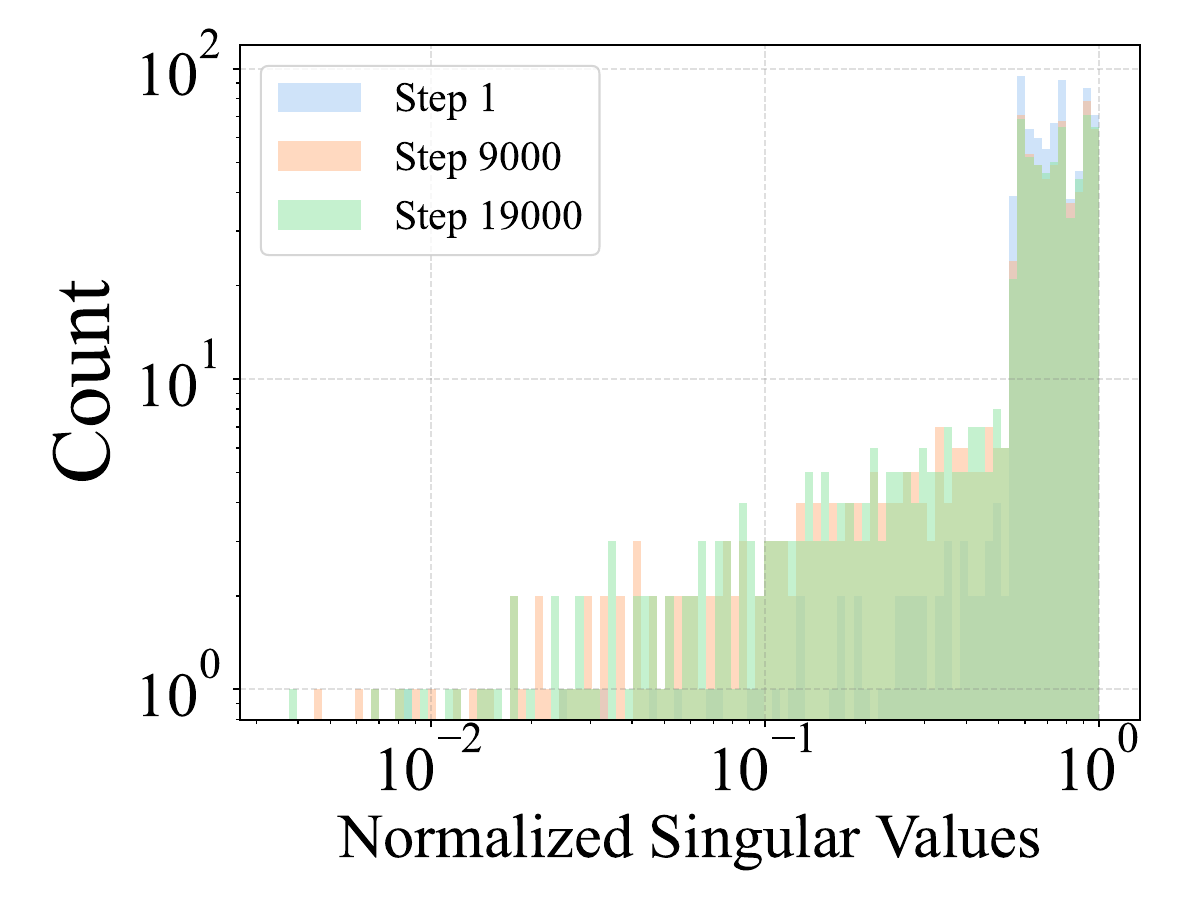}
        \caption{\shortstack{\texttt{Muon\_NS} Update Spectrum}}
        \label{fig:update_spectrum}
    \end{subfigure}
    \begin{subfigure}[b]{0.281\linewidth}
        \centering
        \includegraphics[width=\linewidth]{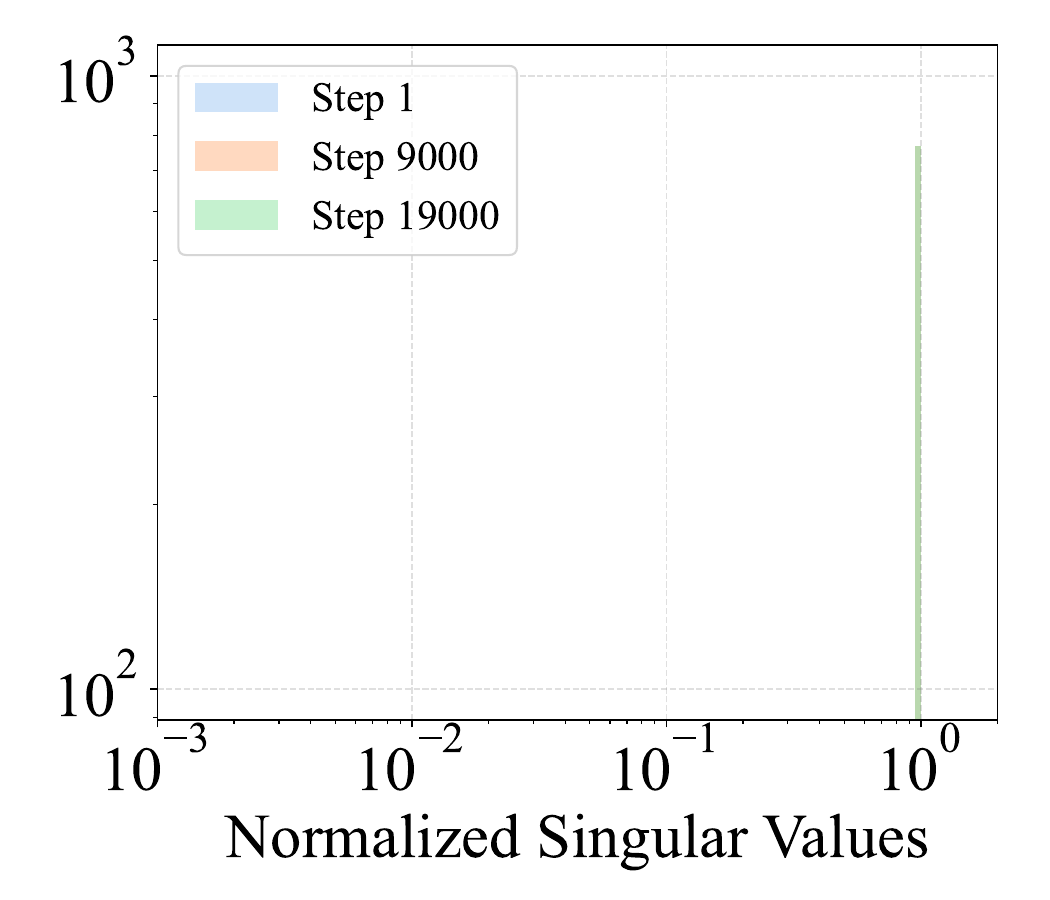}
        \caption{\shortstack{\texttt{Muon\_SVD} Update Spectrum}}
        \label{fig:update_spectrum}
    \end{subfigure}
\caption{\textbf{\texttt{Muon\_NS} vs.\ \texttt{Muon\_SVD} on C4 dataset.}
(a) Validation perplexity for LLaMA-60M/135M trained: \texttt{Muon\_NS} consistently achieves lower perplexity than \texttt{Muon\_SVD}. Both Learning rates for 60M is 0.03 and for 135M is 0.02.
(b)(c) Spectra of update matrices at steps $1/9000/19000$ shown in different colors: \texttt{Muon\_SVD} enforces an exactly "all-ones" spectrum in the update matrices; \texttt{Muon\_NS} stays close to one but retains noticeable deviations, implicitly down-weighting noise-dominated singular-vector directions and correlating with improved performance.}  
 \label{fig:teaser}
\end{figure*}

The theoretical \texttt{Muon} update in Algorithm~\ref{algo:MUON_SVD} has a similar form to polar decomposition: $
\mO_t=\mU_t\mV_t^{\top}=(\mM_t\mM_t^{\top})^{-\frac{1}{2}}\mM_t$.
This is equivalent to setting all singular values of the momentum matrix $\mM_t$ to one. In practice, however, computing $\mU_t$ and $\mV_t$ exactly via \texttt{SVD} is expensive, so \texttt{Muon} uses a \texttt{Newton\_Schulz} iteration \citep{higham1990fast} to approximate $\mO_t$.  In Figure~\ref{fig:teaser}, we train LLaMA-60M and LLaMA-135M on C4 dataset using \texttt{Muon\_SVD} and \texttt{Muon\_NS}, with identical hyperparameters for each model. Figure~\ref{fig:update_spectrum} visualizes the singular-value distributions of the \texttt{Muon} update matrix at different training steps. Although the \texttt{Newton-Schulz} iteration closely approximates the "all-ones" singular spectrum, many singular values still deviate noticeably from one. Meanwhile, as shown in Figure~\ref{fig: Muon_ns_vs_Muon_SVD}, \texttt{Muon\_NS} outperforms \texttt{Muon\_SVD} on both LLaMA-60M and LLaMA-135M. This suggests that keeping the same weight for all singular-vector directions throughout training may not be the most helpful. In particular, directions associated with smaller singular values are often more noise-dominated \citep{sharma2023truth, wang2023spectral,chen2024distributional, liu2025lift, defilippis2025scaling}; enforcing exactly equal weights across all directions yields a light-tailed update, which can make late-stage training more sensitive to noise and limit model capacity. In contrast, due to the polynomial used in the \texttt{Newton-Schulz} steps, \texttt{Muon\_NS} implicitly assigns smaller weights to noise-dominated singular-vector directions, which improves performance. This observation also supports the motivation for making \texttt{Muon} updates more heavy-tailed. 

\begin{algorithm*}[ht]
\caption{\texttt{Muon}}
\label{algo:MUON}
\begin{algorithmic}[1]
\State \textbf{Input:} Initial weights $\mathbf{W}_0 \in \mathbb{R}^{m\times n}$, loss function $L$, learning rate $\eta$, momentum parameter $\beta$,  weight decay $\lambda$.\
\State Initialize $\mathbf{M}_0 \in\mathbb{R}^{m\times n} \leftarrow \mathbf{0}$
\For{$t=1,2,\ldots$}
\State $\mathbf{G}_t \leftarrow \nabla_{\mathbf{W}}L(\mathbf{W}_t)$
\State $\mathbf{M}_t \leftarrow \beta\mathbf{M}_{t-1} + (1-\beta)\mathbf{G}_t$
\State $\mO_t \leftarrow \texttt{ NewtonSchulz5}(\mathbf{M}_t)$
\State $s \leftarrow \sqrt{\max \left( 1, \frac{m}{n}\right)}$
\State $\mathbf{W}_{t+1} \leftarrow \mathbf{W}_t - \eta\lambda\mathbf{W}_t - \eta s\mO_t$
\EndFor
\end{algorithmic}
\end{algorithm*}

\begin{algorithm*}[ht]
\caption{\texttt{Muon\_SVD}}
\label{algo:MUON_SVD}
\begin{algorithmic}[1]
\State \textbf{Input:} Initial weights $\mathbf{W}_0 \in \mathbb{R}^{m\times n}$, loss function $L$, learning rate $\eta$, momentum parameter $\beta$,  weight decay $\lambda$.\
\State Initialize $\mathbf{M}_0 \in\mathbb{R}^{m\times n} \leftarrow \mathbf{0}$
\For{$t=1,2,\ldots$}
\State $\mathbf{G}_t \leftarrow \nabla_{\mathbf{W}}L(\mathbf{W}_t)$
\State $\mathbf{M}_t \leftarrow \beta\mathbf{M}_{t-1} + (1-\beta)\mathbf{G}_t$
\State $\mU_t,\mSigma_t, \mV_t^{\top} \leftarrow \texttt{SVD}(\mathbf{M}_t)$
\State $\mO_t \leftarrow  \mU_t\mV_t^{\top}$
\State $s \leftarrow \sqrt{\max \left( 1, \frac{m}{n}\right)}$
\State $\mathbf{W}_{t+1} \leftarrow \mathbf{W}_t - \eta\lambda\mathbf{W}_t - \eta s\mO_t$
\EndFor
\end{algorithmic}
\end{algorithm*}

\subsection{From Updates to Weights: \texttt{Muon} Produces Lighter-Tailed Weight Spectra}
\label{sec:Muon_makes_less_HT}
In this section, we  show that \texttt{Muon}'s orthogonalization update makes weight matrices ESDs less heavy-tailed, which, based on HT-SR theory,  may limit the model performance.

\begin{figure}[!htb]
    \centering
    
    \begin{subfigure}[t]{0.35\linewidth}
        \includegraphics[width=\textwidth]{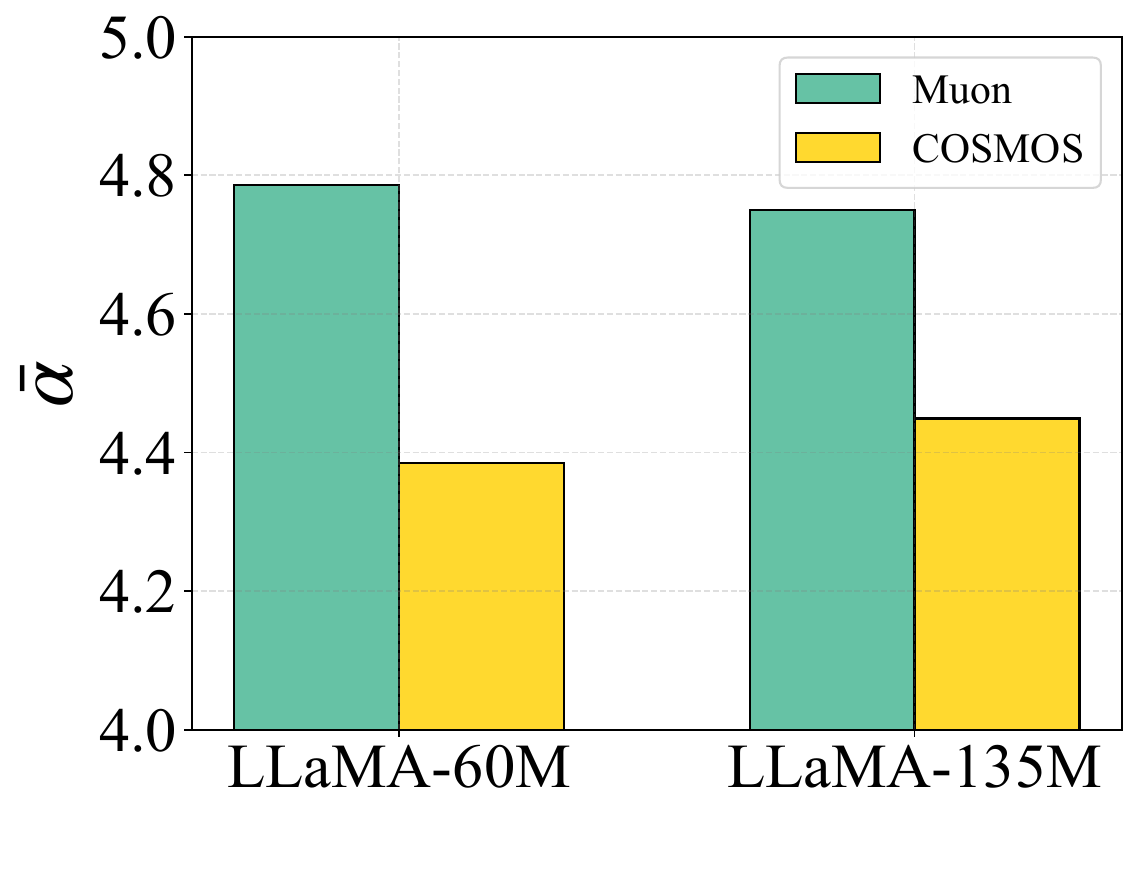}
        \caption{Average PL $\bar{\alpha}$}
        \label{fig:LLaMA-60M-alpha-C4-teaser}
    \end{subfigure}
    \begin{subfigure}[t]{0.35\linewidth}
        \includegraphics[width=\textwidth]{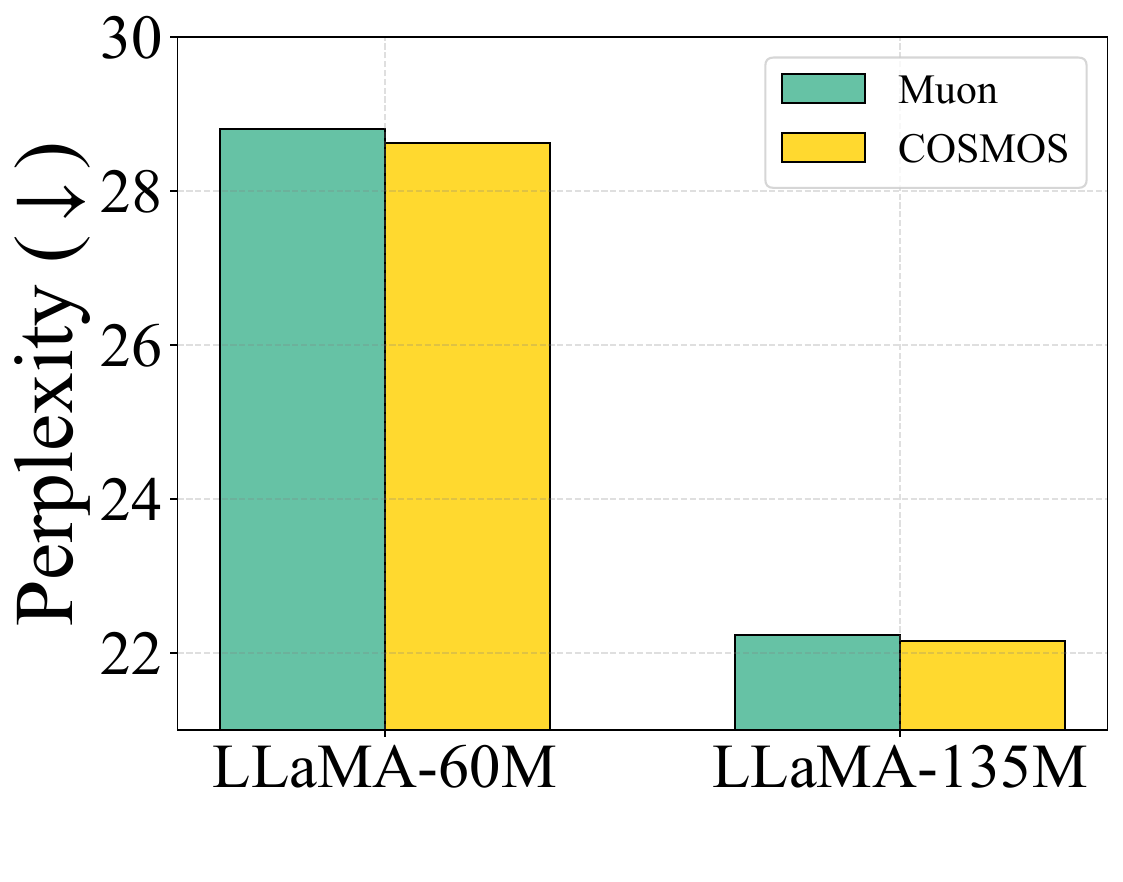}
        \caption{Perplexity}
        \label{fig:LLaMA-135M-alpha-C4-teaser}
    \end{subfigure}
    \caption{(a)Average PL $\bar{\alpha}$ of weight ESDs for  LLaMA-60M and  LLaMA-135M trained on C4 with \texttt{Muon} and \texttt{COSMOS}. \texttt{Muon} yields a higher mean $\bar{\alpha}$, indicating less heavy-tailed spectra than \texttt{COSMOS}; (b) \texttt{COSMOS} outperforms \texttt{Muon} for LLaMA-60M and LLaMA-135M models on C4 datatset.}
\label{fig:LLaMA-alpha-C4-teaser}
\end{figure}

We train LLaMA-60M and LLaMA-135M on the C4 dataset using \texttt{Muon\_NS} and \texttt{COSMOS}, a variant of \texttt{Muon} \citep{liu2025cosmos}. In Figure~\ref{fig:LLaMA-alpha-C4-teaser}, we visualize the average fitted PL exponent $\bar{\alpha}$ for the trained LLaMA models across layers under the two optimizers. We find that  $\bar{\alpha}$ across layers is  higher for \texttt{Muon}, indicating that \texttt{Muon} makes the weight ESD less heavy-tailed. We further observe that \texttt{Muon} yields higher perplexity (PPL) than \texttt{COSMOS} on both models.
Based on HT-SR theory, this indicates that \texttt{Muon}'s orthogonalized update rule may limit the model's final quality, which motivates us to make \texttt{Muon} update more heavy-tailed, leading to more heavy-tailed weight matrices.

\section{Methodology}

In this section, we introduce our method \texttt{HTMuon}. Our design goal is to preserve \texttt{Muon}'s ability to capture parameter interdependencies while making its updates more heavy-tailed, which in turn leads to more heavy-tailed weight ESDs. We put \texttt{HTMuon} in Algorithm~\ref{alg:HTMUON}. \texttt{HTMuon}  is similar to \texttt{Muon}, and the only difference is that we consider a different power transform of the momentum matrix's singular $p$ in line 7. It is easy to know that when  $p=1$, \texttt{HTMuon} reduces to \texttt{SGDM}. \texttt{SGDM} can be viewed as a vector-based optimizer and has limited ability to capture interdependencies among parameters. When $p=0$, \texttt{HTMuon} reduces to \texttt{Muon}; as discussed in Section~\ref{sec:Muon_makes_less_HT},  this update makes weight-matrix ESDs less heavy-tailed.

We therefore propose \texttt{HTMuon} with $p\in(0,1)$. We believe \texttt{HTMuon} helps mitigate the design limitations of \texttt{Muon} and \texttt{SGDM}: On the one hand, in this regime, \texttt{HTMuon} remains a \emph{matrix-based} optimizer and thus retains the ability to capture parameter interdependencies. On the other hand, \texttt{HTMuon} updates are more heavy-tailed than \texttt{Muon} updates, and we illustrate this with a simple example: suppose that during training, the singular values of the momentum matrix $\mM_t$ follow $s_k = s_1 k^{-s}$, where $s_1$ is the largest singular value of $\mM_t$. By Lemma~\ref{lemma:alpha_vs_index}, the PL exponent $\alpha$ of the \texttt{HTMuon} update matrix $\mO_t$ is $1+\frac{1}{2sp}$, it is easy to find that when $p\rightarrow 0$, the $\alpha$ will increase, this means \texttt{HTMuon} update is more heavy-tailed than \texttt{Muon}'s, which will in turn lead to more heavy-tailed weight ESDs.

\begin{lemma}[Proof in Appendix~\ref{app:proof_PL_alpha}]\label{lemma:alpha_vs_index}
    Suppose  the singular values of  matrix $\mW \in \mathbb{R}^{n\times m}$ follow $s_k = s_1k^{-s}, 1\leq k\leq m$, we have PL exponent $\alpha$ of  $\mW$   satisfies $\alpha=1+\frac{1}{2s}.$
\end{lemma}

\begin{algorithm}[ht]
\caption{\texttt{HTMuon}}
\label{algo:HTMUON}
\begin{algorithmic}[1]
\State \textbf{Input:} Initial weights $\mathbf{W}_0 \in \mathbb{R}^{m\times n}$, loss function $L$, learning rate $\eta$, momentum parameter $\beta$,  weight decay $\lambda$, power $p\in (0,1)$.\
\State Initialize $\mathbf{M}_0 \in\mathbb{R}^{m\times n} \leftarrow \mathbf{0}$
\For{$t=1,2,\ldots$}

\State $\mathbf{G}_t \leftarrow \nabla_{\mathbf{W}}L(\mathbf{W}_t)$
\State $\mathbf{M}_t \leftarrow \beta\mathbf{M}_{t-1} + (1-\beta)\mathbf{G}_t$
\State $\mU_t,\mSigma_t, \mV_t^{\top} \leftarrow \texttt{SVD}(\mathbf{M}_t)$
\State \textcolor{blue}{$\mO_t \leftarrow \mU_t\mSigma_t^{p} \mV_t^{\top} $}
\State $s \leftarrow \sqrt{\max \left( 1, \frac{m}{n}\right)}$

\State $\mathbf{W}_{t+1} \leftarrow \mathbf{W}_t - \eta\lambda\mathbf{W}_t - \eta s\mO_t$
\EndFor
\end{algorithmic}
\label{alg:HTMUON}
\end{algorithm}
\section{Experiments}

In this section, we evaluate \texttt{HTMuon} on various pretraining
tasks, and compare it with state-of-the-art pretraining optimizers, including \texttt{Adam} \citep{kingma2014Adam}, \texttt{AdamW} \citep{loshchilov2017decoupled}, \texttt{SGDM} \citep{sutskever2013importance}, \texttt{NorMuon} \citep{li2025normuon}, \texttt{AdaMuon} \citep{si2025adamuon}, \texttt{Cautious} \citep{liang2024cautious}, \texttt{GaLore}~\citep{zhao2024galore}, \texttt{Sophia}~\citep{liu2024sophiascalablestochasticsecondorder}, \texttt{Mars} \citep{yuan2024mars}, \texttt{SOAP} \citep{vyas2024soap}, \texttt{COSMOS} \citep{liu2025cosmos}. We provide detailed definitions of these optimizers in Appendix~\ref{app:optimizerdesign}. Unless otherwise specified, all \texttt{Muon} results in our experiments use the \texttt{Muon\_NS} implementation. We demonstrate that \texttt{HTMuon} achieves superior performance than these optimizers across all datasets. In addition, we show that \texttt{HTMuon} can be used as an add-on method to combine with existing \texttt{Muon} variant optimizers to achieve further improvements. Furthermore, to reduce the overhead of \texttt{SVD} in \texttt{HTMuon}, we design two acceleration schemes that significantly cut the computational cost while still outperforming \texttt{Muon}. To ensure a fair comparison, we reproduce all results of the baseline methods with the codebases provided in previous papers.

\subsection{Experimental Setup}
\textbf{Tasks.} We conduct experiments on various pretraining
tasks, including: 1) \textbf{LLM Pretraining}, in which we  pretrain LLaMA-family models on the C4 dataset \citep{raffel2020exploring}, GPT-2 family models on the OpenWebText dataset \citep{Gokaslan2019OpenWeb}. 2) \textbf{Image Classification}, where we train ResNet models on the CIFAR-100 and CIFAR-10 datasets \citep{krizhevsky2009learning} and train ViT-tiny on the ImageNet-1K dataset \citep{deng2009imagenet}.

\noindent\textbf{Models.} For LLM pretraining, we evaluate \texttt{HTMuon} on LLaMA-60M, LLaMA-135M, LLaMA-350M, LLaMA-1B \citep{touvron2023llama} and GPT-2 small (125M) \citep{radford2019language}. For image classification, we use ResNet18, ResNet50 \citep{he2016deep} and ViT-tiny \citep{dosovitskiy2020image}. 

Unless otherwise specified, we use $p=0.125$ for \texttt{HTMuon} throughout. We discuss the specific choice of 
$p$ in Section~\ref{sec:ablation study}. 
All baseline methods and \texttt{HTMuon} are carefully tuned for a fair comparison.  For detailed hyperparameter settings, please refer to Appendix~\ref{app:hyperparameter}.

\subsection{LLM Pretraining}
In this section, we evaluate \texttt{HTMuon}  against several baseline optimizers on C4  and OpenWebText datasets. 

\paragraph{\texttt{HTMuon} consistently outperforms \texttt{Muon}.} In Table~\ref{table:main_result}, We find that \texttt{HTMuon} consistently outperforms \texttt{Muon} across three model scales (60M, 135M, and 350M) on C4 dataset. In particular, on LLaMA-60M, \texttt{HTMuon} achieves a PPL that is 0.92 lower than \texttt{Muon} and 4.33 lower than \texttt{Adam}; on LLaMA-135M, \texttt{HTMuon} is 0.98 lower than \texttt{Muon} and 2.08 lower than \texttt{AdamW}. In Table~\ref{table:OpenWebText_results}, we show that on GPT-2 small, \texttt{HTMuon} achieves a PPL that is 0.26 lower than \texttt{Muon} and 2.99 lower than \texttt{AdamW}.  In Figure~\ref{fig:loss} in Appendix~\ref{app:more_results}, we provide the training loss curves of \texttt{HTMuon} and \texttt{Muon}. These results suggest that \texttt{HTMuon}'s update rule effectively mitigates the limitations of \texttt{Muon} updates.

\begin{table}[!t]
\centering
\caption{Comparison with dominant Pretraining Optimizers on LLaMA Models of Varying Sizes on the C4 dataset. Lower perplexity indicates better performance. Detailed hyperparameters are provides in Table~\ref{Table:Hyperparameters of LLaMA}, ~\ref{table:hyperparam_C4} and~\ref{table:hyperparam_C4_2} in Appendix~\ref{app:hyperparameter}.}
\scalebox{0.95}{
\begin{tabular}{@{}lccc@{}}
\toprule
 & \textbf{LLaMa-60M} & \textbf{LLaMa-135M} & \textbf{LLaMa-350M} \\
\midrule
\texttt{Adam} & 32.21 &23.01  & 17.11 \\
\texttt{AdamW} & 31.85& 23.33 & 16.96 \\
\texttt{Muon} &  28.80& 22.23 &  16.81\\
\rowcolor[HTML]{FFF9E5}
\texttt{HTMuon} & \textbf{27.88} & \textbf{21.25} & \textbf{16.79} \\
\bottomrule
\end{tabular}}
\label{table:main_result}
\vspace{-5pt}
\end{table}

\begin{figure}[!htb]
    \centering
    \begin{subfigure}{0.78\linewidth}
        \includegraphics[width=\textwidth]{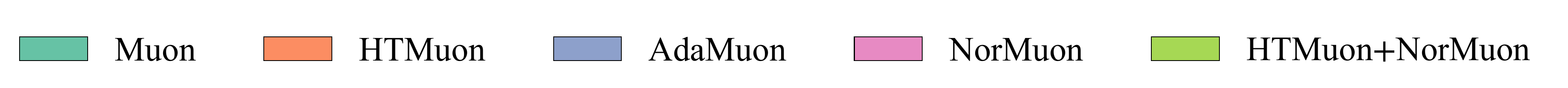}
    \end{subfigure}

    \begin{subfigure}[t]{0.35\linewidth}
        \includegraphics[width=\textwidth]{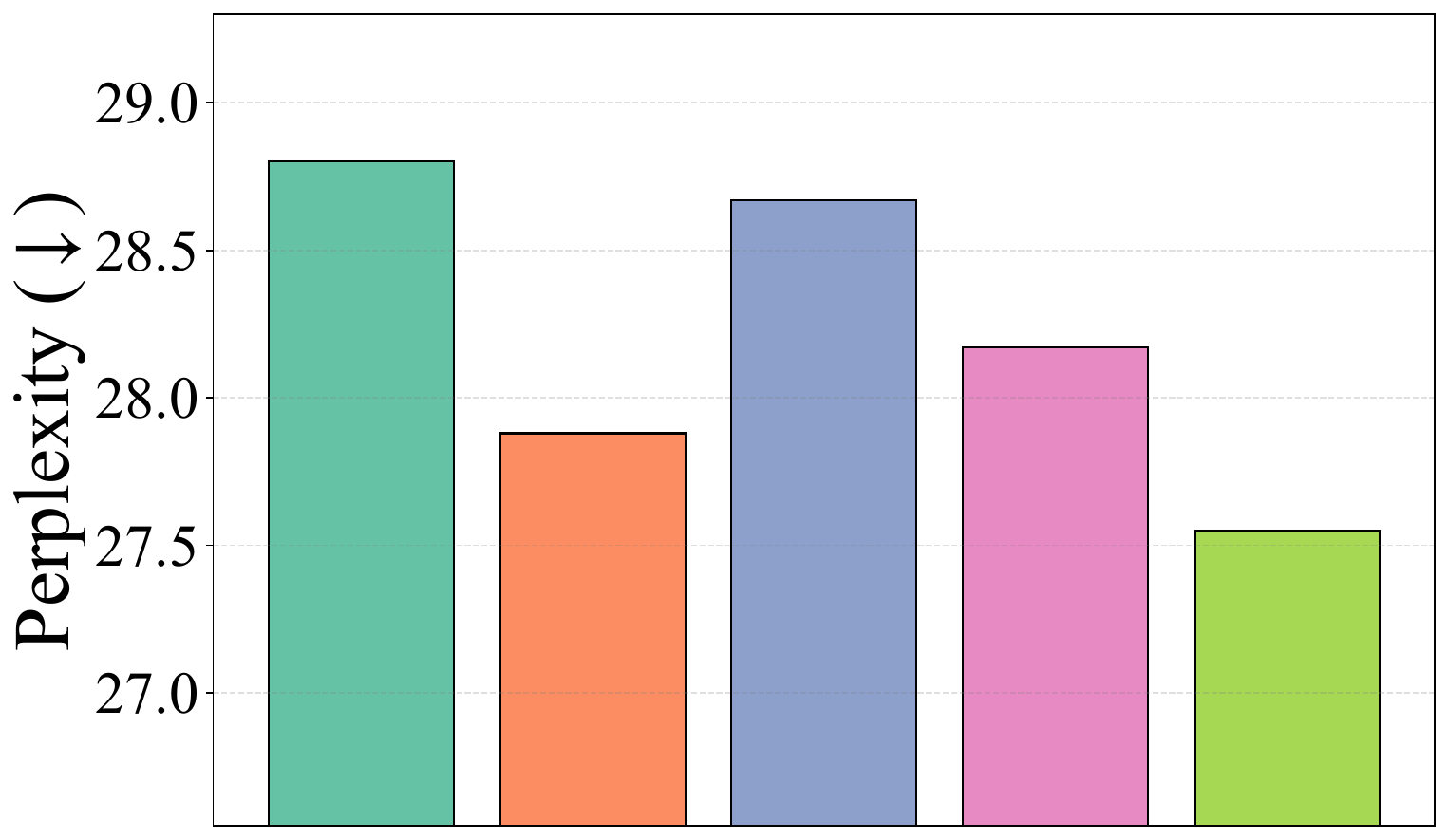}
        \caption{LLaMA-60M }
        \label{fig:LLaMA-60M_muon_variant}
    \end{subfigure}
    \begin{subfigure}[t]{0.35\linewidth}
        \includegraphics[width=\textwidth]{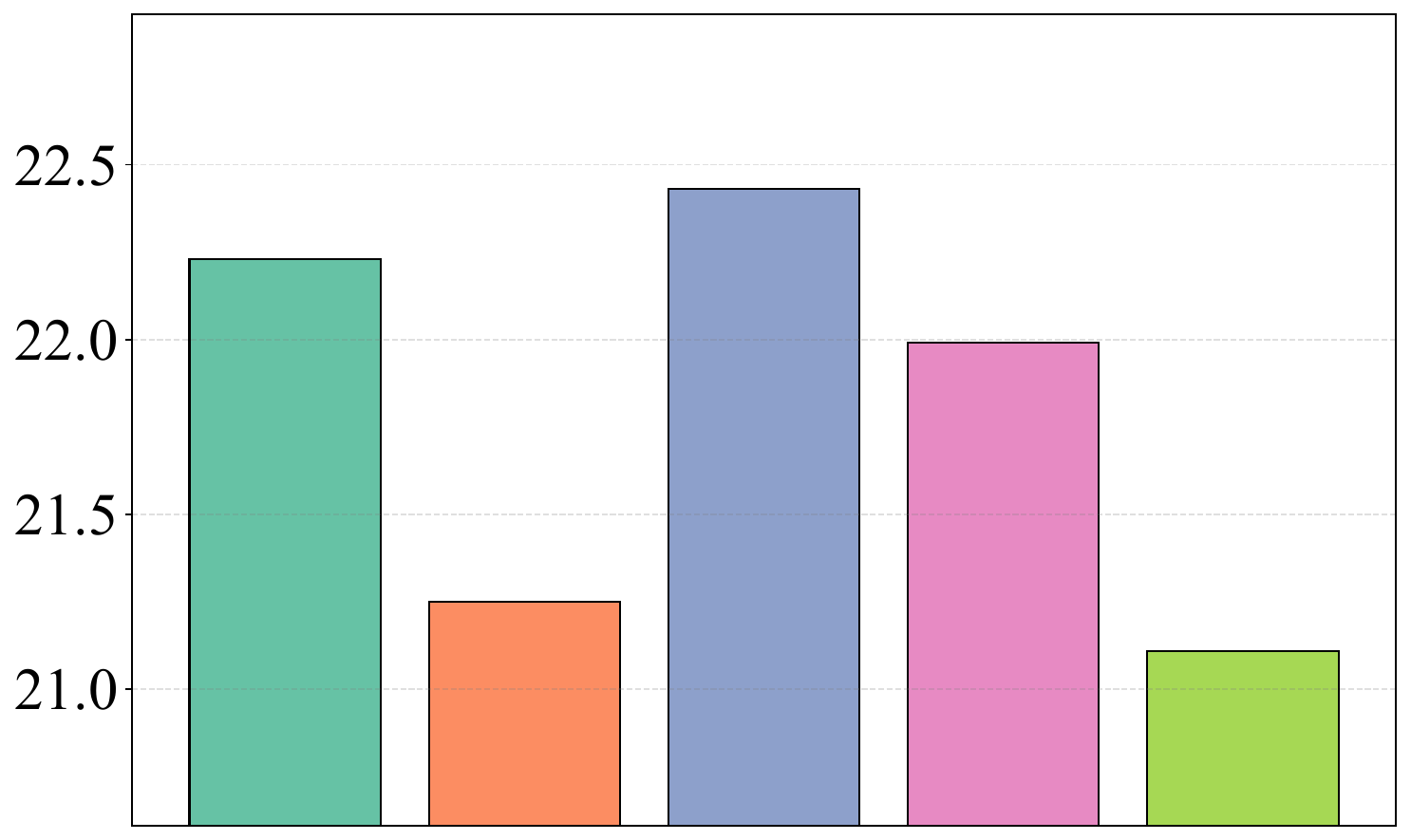}
        \caption{LLaMA-135M }
        \label{fig:LLaMA-130M_muon_variant}
    \end{subfigure}
    \caption{Comparison with \texttt{Muon} variant  optimizers on LLaMA-60M and 135M on C4 dataset. All optimizers are carefully tuned via grid search; detailed results and hyperparameter settings are provided in Table~\ref{table: muon_variants} and~\ref{table:hyperparam_C4} in Appendix~\ref{app:more_results} and~\ref{app:hyperparameter}.}
\label{fig:Muon_variants}
\end{figure}

\paragraph{\texttt{HTMuon} consistently outperforms other variants of \texttt{Muon}.} In Figure~\ref{fig:Muon_variants}, we find that \texttt{HTMuon} consistently outperforms \texttt{Muon} variant optimizers like \texttt{AdaMuon} and \texttt{NorMuon}. Specifically, on LLaMA-60M, our PPL is 0.29 lower than that of \texttt{NorMuon}, the strongest baseline after \texttt{HTMuon}; on LLaMA-135M, we outperform \texttt{NorMuon} by 0.74 PPL and \texttt{AdaMuon} by 1.18 PPL. Moreover, we observe that \textbf{combining \texttt{HTMuon} with \texttt{NorMuon} yields further gains}, reducing PPL by an additional 0.33 on LLaMA-60M and 0.14 on LLaMA-135M. 

\begin{table}[!t]
\centering
\caption{Comparison with dominant Pretraining Optimizers on GPT-2 small on the OpenWebText dataset. Lower perplexity indicates better performance.  Hyperparameter settings are provided in  Appendix~\ref{app:hyperparameter}.}
\setlength{\tabcolsep}{20pt}
\scalebox{0.88}{
\begin{tabular}{@{}lccc@{}}
\toprule
 & \texttt{AdamW}&  \texttt{Muon} &  \texttt{HTMuon} \\
\midrule
GPT-2 small & 25.19 &  22.46 &   \textbf{22.20} \\
\bottomrule
\end{tabular}}
\vspace{-5pt}
\label{table:OpenWebText_results}
\end{table}

\begin{figure}[!htb]
    \centering
    
    \begin{subfigure}{0.78\linewidth}
        \includegraphics[width=\textwidth]{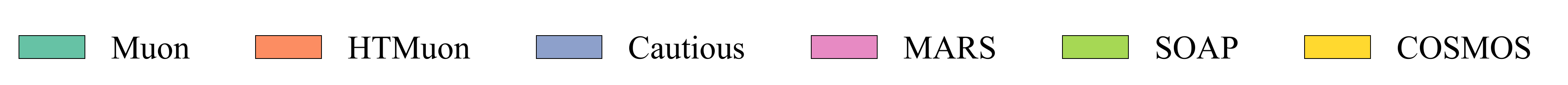}
    \end{subfigure}

    \begin{subfigure}[t]{0.35\linewidth}
        \includegraphics[width=\textwidth]{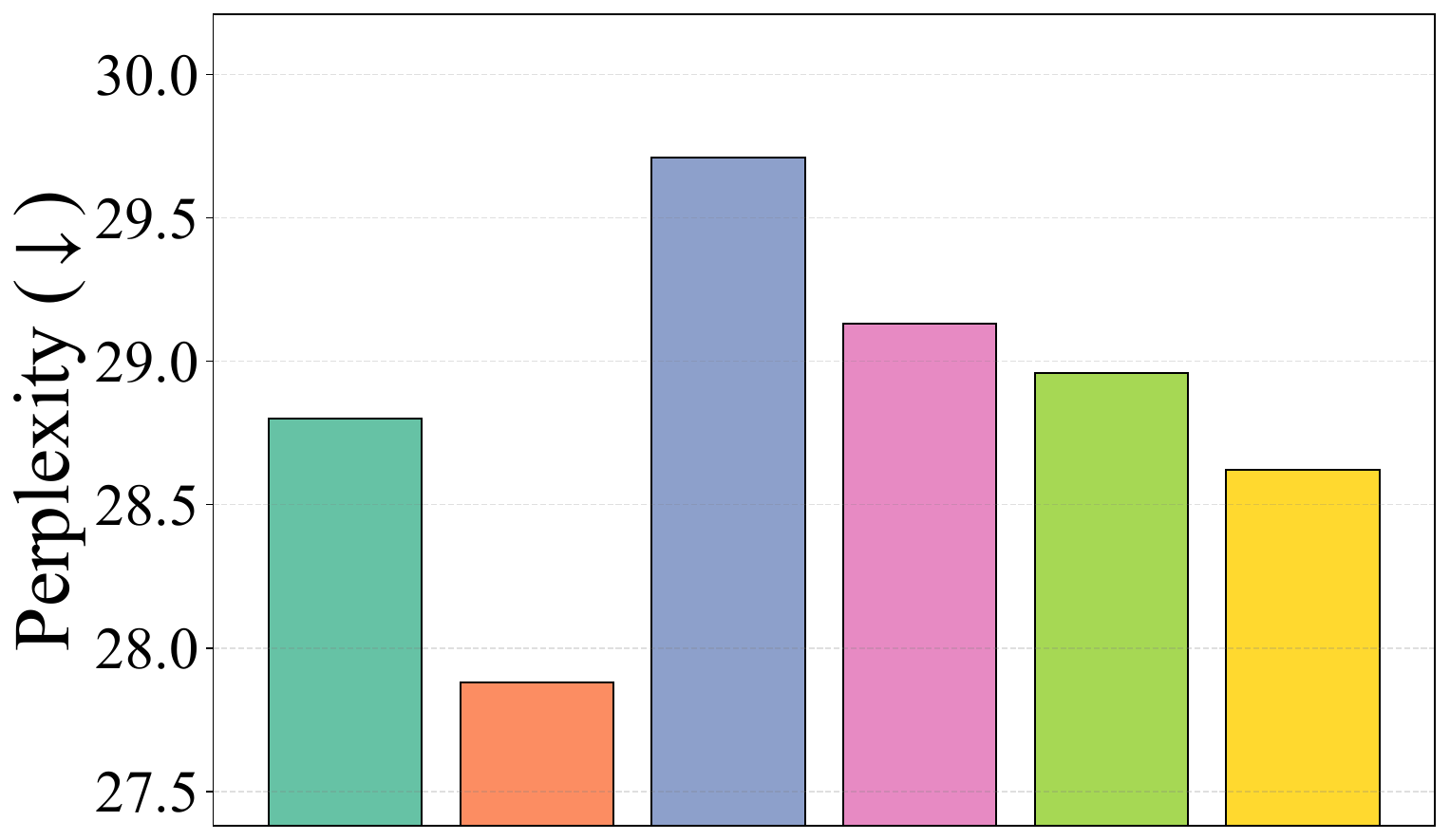}
        \caption{LLaMA-60M }
        \label{fig:LLaMA-60M_optimizer_variant}
    \end{subfigure}
    \begin{subfigure}[t]{0.35\linewidth}
        \includegraphics[width=\textwidth]{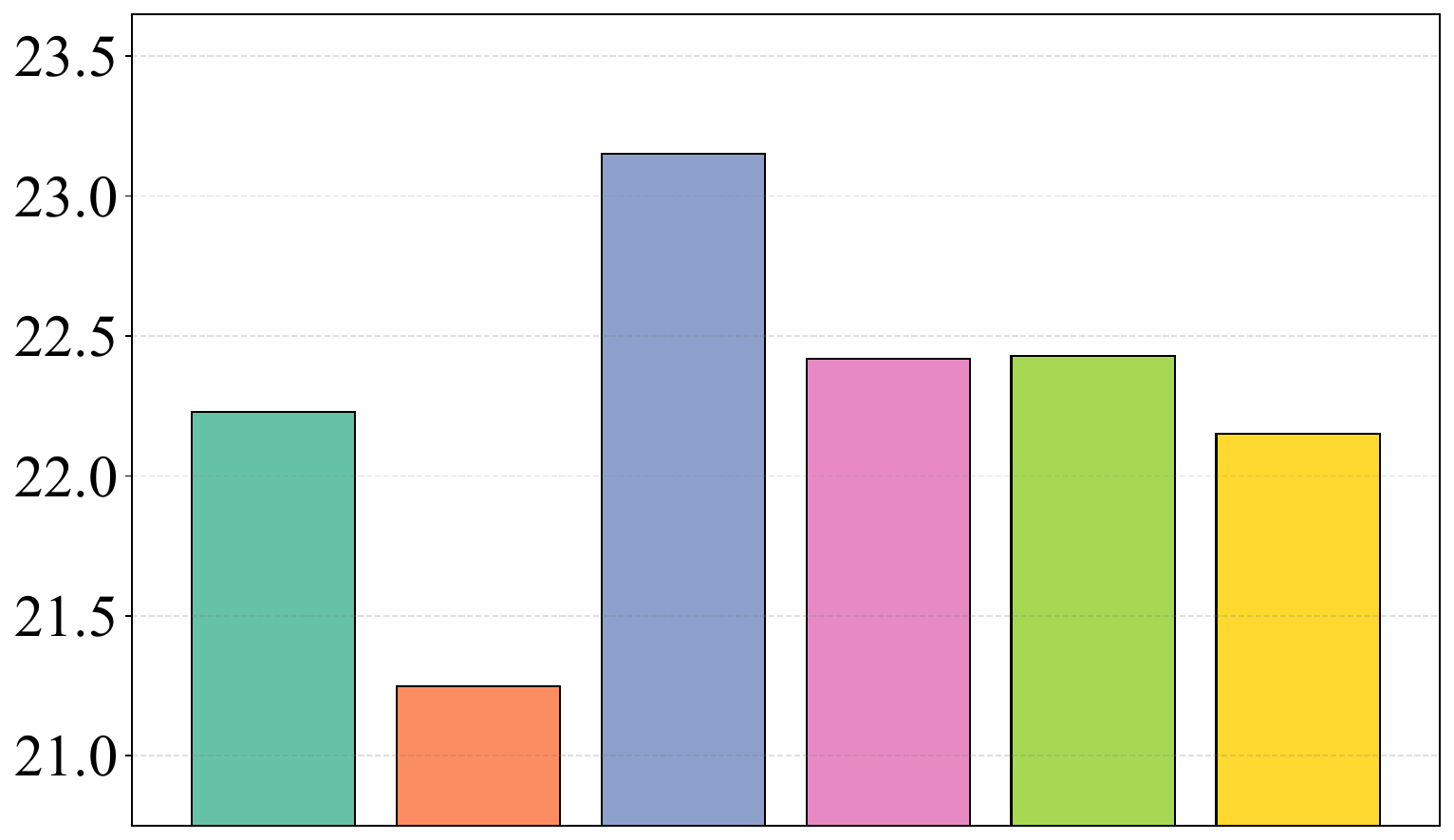}
        \caption{LLaMA-135M }
        \label{fig:LLaMA-130M_optimizer_variant}
    \end{subfigure}
\caption{Comparison with state-of-the-art pretraining optimizers on LLaMA-60M and 135M on C4 dataset. All optimizers are carefully tuned via grid search; detailed results and hyperparameter settings are provided in Table~\ref{table: muon_variants} and~\ref{table:hyperparam_C4} in Appendix ~\ref{app:more_results} and~\ref{app:hyperparameter}.}
\label{fig:LLaMA_optimizer_variant}
\end{figure}

\paragraph{\texttt{HTMuon} consistently outperforms state-of-the-art pretraining optimizers.} In Figure~\ref{fig:LLaMA_optimizer_variant}, we compare \texttt{HTMuon} with state-of-the-art optimizers on LLaMA-60M and LLaMA-135M. \texttt{HTMuon} consistently outperforms all competing methods at both scales. For example, compared to \texttt{COSMOS} (the second-best method), \texttt{HTMuon} effectively reduces PPL by 1.07 on LLaMA-60M and by 1.04 on LLaMA-135M. In Appendix~\ref{app:more_results}, we include more recent optimizers, including \texttt{GaLore} and \texttt{Sophia}.

\subsection{Image Classification}

In this section, we compare \texttt{HTMuon}  against several baseline optimizers on CIFAR and ImageNet-1K datasets. 

In Table~\ref{tab:cifar}, we find that, when training ResNet18 and ResNet50 on CIFAR-10 and CIFAR-100, \texttt{HTMuon} consistently achieves higher accuracy than \texttt{SGDM}, \texttt{Muon}, and \texttt{NorMuon}. For example, relative to \texttt{Muon}, \texttt{HTMuon} improves accuracy by up to 0.31 \%  on CIFAR-100 and up to 0.24 \% on CIFAR-10. Moreover, on CIFAR-100, combining \texttt{HTMuon}  with \texttt{NorMuon} further improves the accuracy of both ResNet18 and ResNet50. In Table~\ref{table:Imagenet_1k_results}, we show that \texttt{HTMuon} also outperforms \texttt{Muon} and \texttt{Adam} when training ViT-tiny on ImageNet-1K.

\begin{table}[t]
\centering
\caption{Results on CIFAR-100 and CIFAR-10 dataset. Higher accuracy indicates better performance.}
\label{tab:cifar}
\setlength{\tabcolsep}{3pt}
\renewcommand{\arraystretch}{0.9} %
\resizebox{0.6\linewidth}{!}{
\begin{tabular}{l|cc|cc}
\toprule
\multicolumn{1}{c}{} & \multicolumn{2}{c}{CIFAR-100} & \multicolumn{2}{c}{CIFAR-10} \\
\cmidrule(lr){2-3} \cmidrule(lr){4-5}
\multicolumn{1}{c}{Optimizer} & ~ResNet18~ & \multicolumn{1}{c}{~ResNet50~} & \multicolumn{1}{c}{~ResNet18~} & ~ResNet50~ \\
\midrule
\texttt{SGDM}        & 77.82 & 78.57 & 95.15 & 95.16 \\
\texttt{Muon}        & 77.95 & 79.85 & 95.39 & 95.97\\
\texttt{NorMuon}     & 77.82 & 79.78 & 95.57 & 96.03 \\
\rowcolor[HTML]{FFF9E5}
\texttt{HTMuon}      & 78.24  & 80.16 & \textbf{95.63} & \textbf{96.13} \\
\rowcolor[HTML]{FFF9E5}
\texttt{HTMuon} + \texttt{NorMuon}~ & \textbf{78.51} &  \textbf{80.22}& 95.40 & 96.04 \\
\bottomrule
\end{tabular}}
\end{table}

\begin{table}[!t]
\centering
\caption{Results on ImageNet-1K. Hyperparameter settings are provided in  Appendix~\ref{app:hyperparameter}.}
\setlength{\tabcolsep}{20pt}
\scalebox{0.88}{
\begin{tabular}{@{}lccc@{}}
\toprule
 & \texttt{Adam}& \texttt{Muon} & \texttt{HTMuon} \\
\midrule
ViT-Tiny & 67.67 & 71.02 &   \textbf{71.16} \\
\bottomrule
\end{tabular}}
\vspace{-5pt}
\label{table:Imagenet_1k_results}
\end{table}

\subsection{More Efficient Implementations of \texttt{HTMuon}}\label{sec:More_Efficient_Implementations}
\begin{figure*}[!htb]
    \centering

    \begin{subfigure}{0.85\linewidth}
        \centering
        \includegraphics[width=\linewidth]{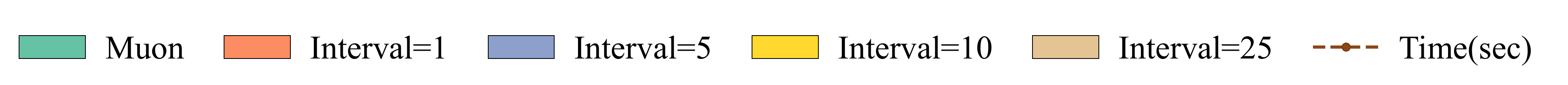}
    \end{subfigure}

    \vspace{2mm} %

    \begin{subfigure}[b]{0.24\linewidth}
        \centering
        \includegraphics[width=\linewidth]{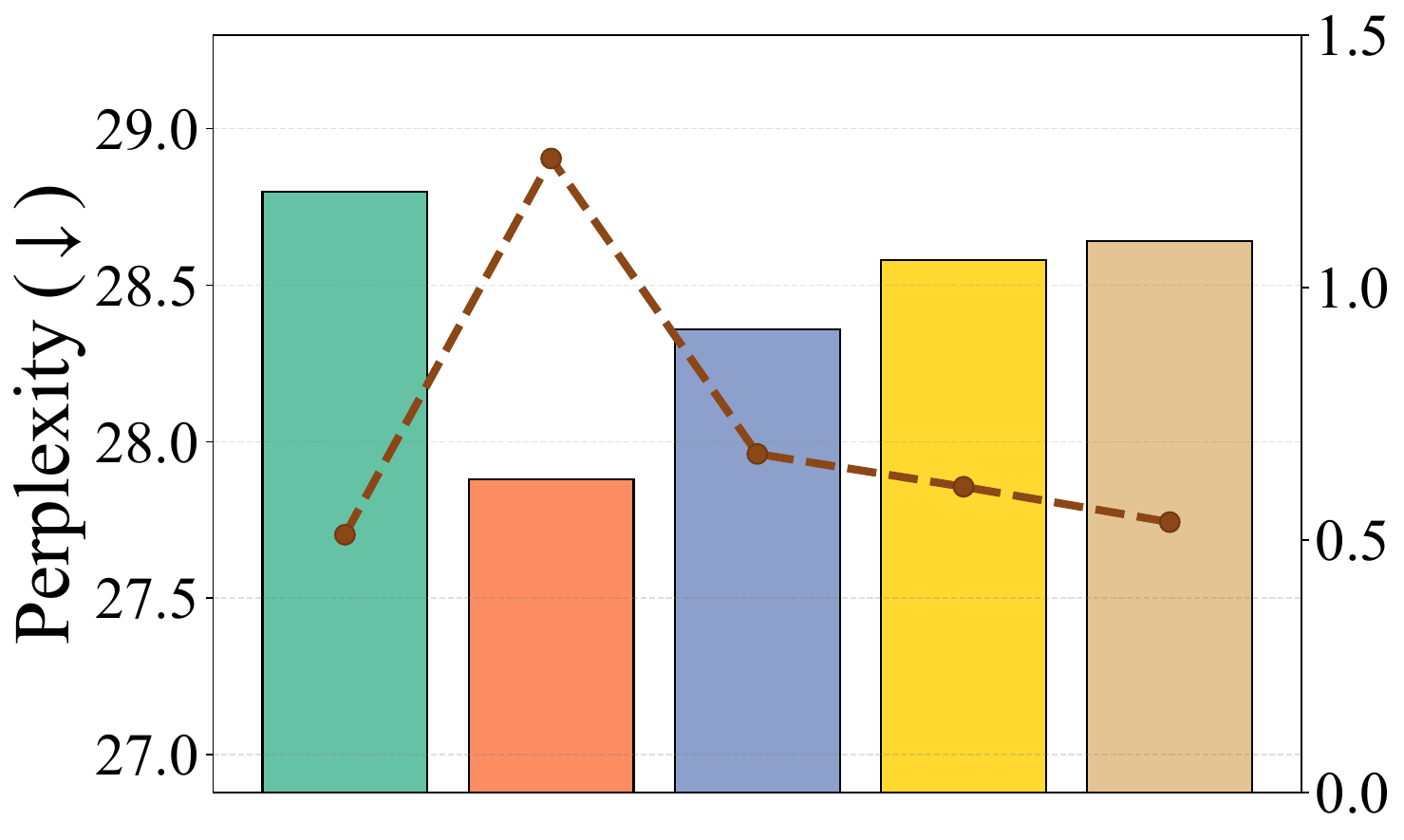}
        \caption{\shortstack{HTMuon\\LLaMA-60M}}
        \label{fig:LLaMA-60M_htmuon_interval}
    \end{subfigure}\hfill
    \begin{subfigure}[b]{0.24\linewidth}
        \centering
        \includegraphics[width=\linewidth]{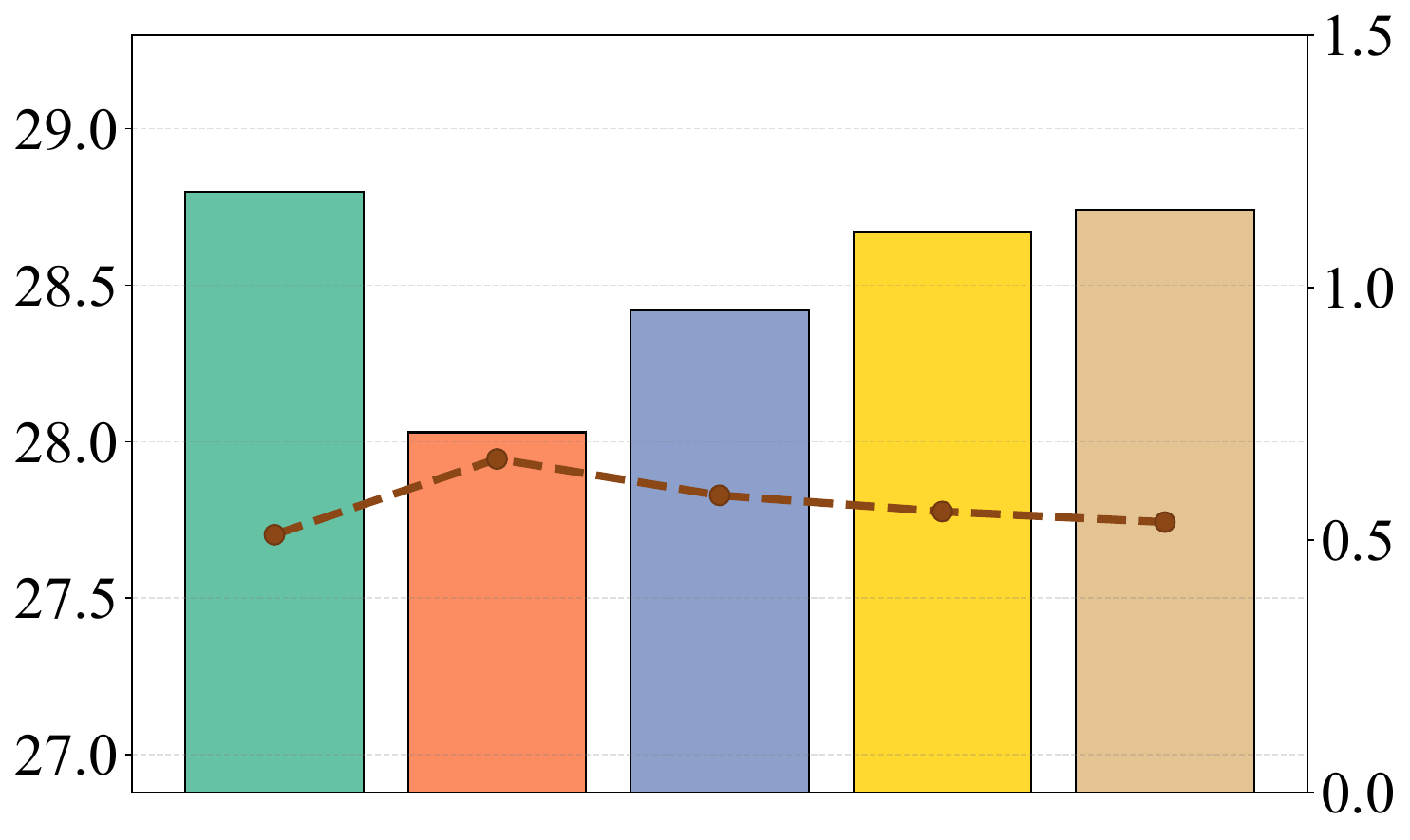}
        \caption{\shortstack{HTMuon\_NS\\LLaMA-60M}}
        \label{fig:LLaMA-60M_htmuon_interval_acc}
    \end{subfigure}\hfill
    \begin{subfigure}[b]{0.24\linewidth}
        \centering
        \includegraphics[width=\linewidth]{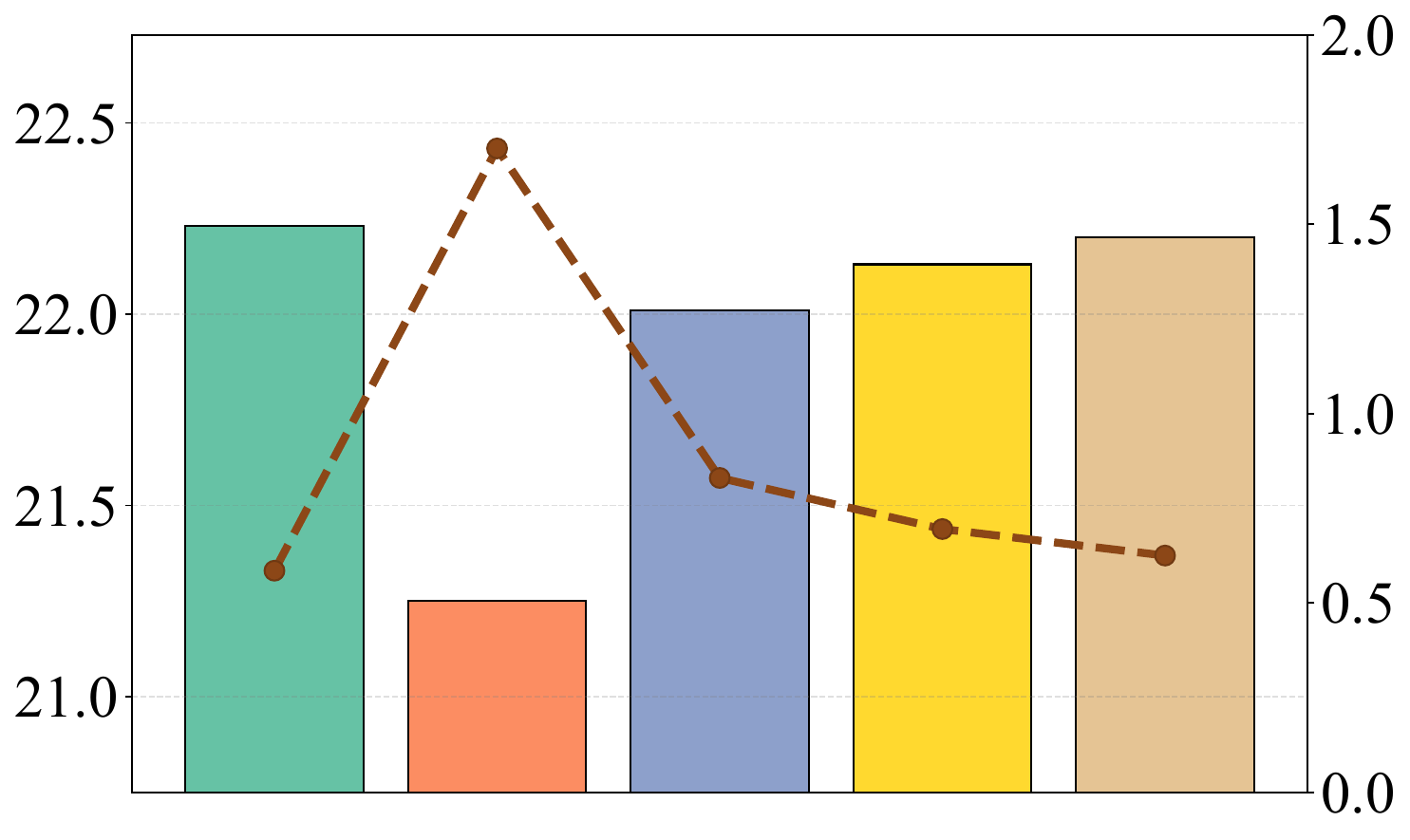}
        \caption{\shortstack{HTMuon\\LLaMA-135M}}
        \label{fig:LLaMA-135M_htmuon_interval}
    \end{subfigure}\hfill
    \begin{subfigure}[b]{0.24\linewidth}
        \centering
        \includegraphics[width=\linewidth]{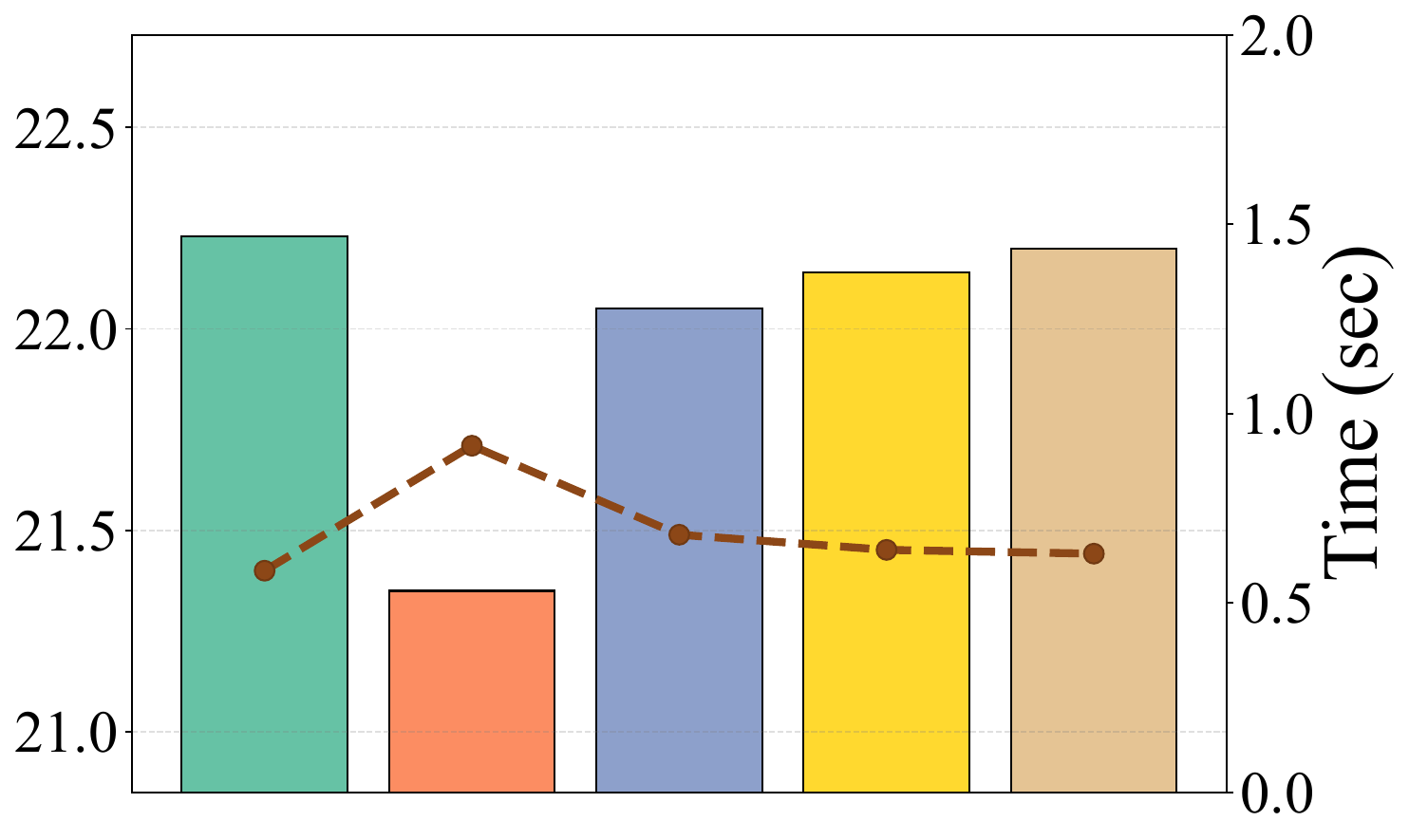}
        \caption{\shortstack{HTMuon\_NS\\LLaMA-135M}}
        \label{fig:LLaMA-135M_htmuon_interval_acc}
    \end{subfigure}

    \caption{we evaluate applying \texttt{HTMuon} and \texttt{HTMuon\_NS}  on LLaMA-60M and LLaMA-135M every 1, 5, 10, and 25 steps. We report the average per-step runtime overhead for all methods. Detailed results and hyperparameter settings are provided in Table~\ref{table:ppl_intervals} and~\ref{table:time_intervals} in Appendix~\ref{app:more_results}.}
    \label{fig:interval}
\end{figure*}

In this section, to reduce the overhead introduced by \texttt{SVD} in \texttt{HTMuon}, we consider two acceleration strategies:

\paragraph{(i) Apply \texttt{HTMuon} every fixed number of training steps and use Muon otherwise.}  In Figure~\ref{fig:LLaMA-60M_htmuon_interval} and~\ref{fig:LLaMA-135M_htmuon_interval},  we evaluate applying \texttt{HTMuon}  on LLaMA-60M and LLaMA-135M every 5, 10, and 25 steps. Compared to using \texttt{HTMuon} at every step, this strategy greatly reduces runtime overhead, making it competitive with \texttt{Muon}, while still outperforming \texttt{Muon}. We also observe that smaller intervals lead to better performance, further supporting the effectiveness of our optimizer.  

Based on these findings, we train LLaMA-1B on C4 using \texttt{HTMuon} with interval $=5$. As shown in Table~\ref{table:LLaMA_1B}, \texttt{HTMuon} still outperforms \texttt{Muon}, highlighting its potential for large-scale training.

\paragraph{(ii) Replace the \texttt{SVD} with a numerical iterative method for faster computation.} In Algorithm~\ref{alg:HTMUON}, We note that the \texttt{HTMuon} update $\mO_t=\mU_t \mSigma_t^{p}\mV_t^{\top}$ admits the factorization $\mO_t = (\mU_t\mV_t^{\top})\,(\mV_t\mSigma_t^{p}\mV_t^{\top}).$
For the $\mU_t\mV_t^{\top}$ part, this is \texttt{Muon}'s update; we can consider  \texttt{Newton-Schulz} with 5 steps. To compute  $\mV_t\mSigma_t^{p}\mV_t^{\top}$ efficiently, we apply the \texttt{NS\_root} routine to $\mM_t^{\top}\mM_t$ in Algorithm~\ref{algo:NS_2p_root}, since $\mM_t^{\top}\mM_t=\mV_t\mSigma_t^{2}\mV_t^{\top},$
we have $\mV_t\mSigma_t^{p}\mV_t^{\top}=(\mM_t^{\top}\mM_t)^{\frac{p}{2}}$. Accordingly, \texttt{NS\_root} uses \texttt{Newton-Schulz} iterations to approximate matrix square roots and applies successive square-root operations for $[1-\log_2 p]$ rounds. Combining these two, we obtain \texttt{HTMuon\_NS} in Algorithm~\ref{algo:HTMUON_NS}. In Figure~\ref{fig:LLaMA-60M_htmuon_interval_acc} and~\ref{fig:LLaMA-135M_htmuon_interval_acc}, comparing \texttt{HTMuon\_NS} with  \texttt{HTMuon} with interval$=1$,  we find \texttt{HTMuon\_NS} substantially reduces runtime overhead, while only slightly increases PPL. We also apply \texttt{HTMuon\_NS} on LLaMA-60M and LLaMA-135M every 5, 10, and 25 steps. We find that this further reduces runtime overhead while still maintaining better performance than \texttt{Muon}, for example, \texttt{HTMuon\_NS} (interval $= 5$) achieves 0.59 s/step compared to \texttt{Muon}’s 0.51 s/step in Table~\ref{table:time_intervals}(which only introduces a minor increase in overhead), Table~\ref{table:ppl_intervals} shows that \texttt{HTMuon\_NS} (interval $= 5$) still outperforms Muon by 0.38 PPL.

\begin{algorithm*}[ht]
\caption{\texttt{HTMuon\_NS}}
\label{algo:HTMUON_NS}
\begin{algorithmic}[1]
\State \textbf{Input:} Initial weights $\mathbf{W}_0 \in \mathbb{R}^{m\times n}$, loss function $L$,  learning rate $\eta$, momentum parameter $\beta$,  weight decay $\lambda$, power $p$.\
\State Initialize $\mathbf{M}_0 \in\mathbb{R}^{m\times n} \leftarrow \mathbf{0}$
\For{$t=1,2,\ldots$}
\State $\mathbf{G}_t \leftarrow \nabla_{\mathbf{W}}L(\mathbf{W}_t)$
\State $\mathbf{M}_t \leftarrow \beta\mathbf{M}_{t-1} + (1-\beta)\mathbf{G}_t$
\State $\mO^{\prime}_t \leftarrow \texttt{ NS5}(\mathbf{M}_t)$
\State $\mO^{\prime \prime}_t \leftarrow \texttt{ NS\_2p\_root}(\mathbf{M}_t^{\top}\mathbf{M}_t)$
\State $\mO_t \leftarrow \mO^{\prime}_t \mO^{\prime \prime}_t $
\State $s \leftarrow \sqrt{\max \left( 1, \frac{m}{n}\right)}$
\State $\mathbf{W}_{t+1} \leftarrow \mathbf{W}_t - \eta\lambda\mathbf{W}_t - \eta s\mO_t$
\EndFor
\end{algorithmic}
\end{algorithm*}

\begin{algorithm*}[ht]
\caption{\texttt{ NS\_root}}
\label{algo:NS_2p_root}
\begin{algorithmic}[1]
\State \textbf{Input:}  Positive definite matrix $\mX^{n\times n}$,  small constant $\varepsilon$, power $p$, ns\_steps $T$.\
\For{$t=1,2,\ldots, [\log_2{\frac{2}{p}}]$}
\State $\alpha \leftarrow \|X\|_\F$
\State $\mX \leftarrow \frac{\mX}{\|X\|_\F+\varepsilon}$
\State $\mY \leftarrow \mX, \mZ \leftarrow \mI$
\For {$i=1,2,\ldots, T$}
\State $\mQ \leftarrow 3.0\mI- \mZ\mY$
\State $\mY \leftarrow 0.5\mY\mQ$
\State $\mZ \leftarrow 0.5\mQ\mZ $
\EndFor
\State $\mX \leftarrow \sqrt{\alpha}\mY$ , $\mX \leftarrow \frac{\mX+\mX^{\top}}{2} + \epsilon \mI$
\EndFor
\end{algorithmic}
\end{algorithm*}

We also evaluate \texttt{HTMuon} $+$ \texttt{NorMuon} with different intervals. As observed earlier, using larger intervals reduces runtime overhead while still outperforming \texttt{NorMuon}. Please refer to Figure~\ref{fig:Normuon_interval} in Appendix~\ref{app:more_results}.

Besides reporting per-step runtime overhead, we also report the wall-clock time  for \texttt{Muon}, \texttt{HTMuon} and \texttt{HTMuon\_NS} with intervals. To better report wall-clock time, we rerun training on NVIDIA RTX PRO 6000  GPUs: LLaMA-60M on C4 using 2 GPUs, and LLaMA-135M on C4 using 4 GPUs. We use the optimal hyperparameter settings reported in the  Table~\ref{table:hyperparam_C4}. We summarize the PPL and wall-clock time of \texttt{Muon}, \texttt{HTMuon}, \texttt{HTMuon\_NS}, \texttt{HTMuon}  (Interval $= 5$), and \texttt{HTMuon\_NS} (Interval $= 5$) in the Table~\ref{table:wall_time_clock} in Appendix~\ref{app:more_results}. 

We observe that for LLaMA-60M, \texttt{HTMuon\_NS} (Interval $= 5$) achieves only a $\sim$6\% additional overhead yet outperforms \texttt{Muon} by 0.34 PPL. Moreover, the time for \texttt{HTMuon\_NS} (Interval $=5$) to reach 28.84 PPL is 0.64 hour, compared to 0.67 hour for \texttt{Muon}. For LLaMA-135M, \texttt{HTMuon\_NS} (Interval = 5) incurs only $\sim$11\% additional overhead while outperforming \texttt{Muon} by 0.25 PPL; it reaches 22.27 PPL in 1.32 hour, compared to 1.40 hour for Muon. We also give the FLOPs analysis for \texttt{Muon} and \texttt{HTMuon\_NS} in Appendix~\ref{app:more_results}.

We find that although \texttt{HTMuon} could take longer than \texttt{Muon}, the substantial gains achieved through heavy-tailed spectral correction suggest that \texttt{HTMuon} captures the correct inductive bias for learning. Furthermore, we believe that efficient variants such as \texttt{HTMuon\_NS} with interval-based updates provide a strong balance between performance and efficiency.

\begin{table}[!t]
\centering
\caption{We train LLaMA-1B with interval$=5$ on C4. Hyperparameter settings are provided in Table~\ref{table:hyperparam_C4_2} in Appendix~\ref{app:hyperparameter}.}
\setlength{\tabcolsep}{18pt}
\scalebox{0.88}{
\begin{tabular}{@{}lcccc@{}}
\toprule
 & \texttt{Adam} & \texttt{AdamW} & \texttt{Muon} & \texttt{HTMuon} \\
\midrule
\textbf{LLaMA-1B} & 15.22& 15.11 & 14.33 &  \textbf{14.17}  \\
\bottomrule
\end{tabular}}
\vspace{-5pt}
\label{table:LLaMA_1B}
\end{table}

\subsection{ Weight Spectrum Analysis}
\begin{figure*}[!htb]
    \centering

        \begin{subfigure}[b]{0.24\linewidth}
        \centering
        \includegraphics[width=\linewidth]{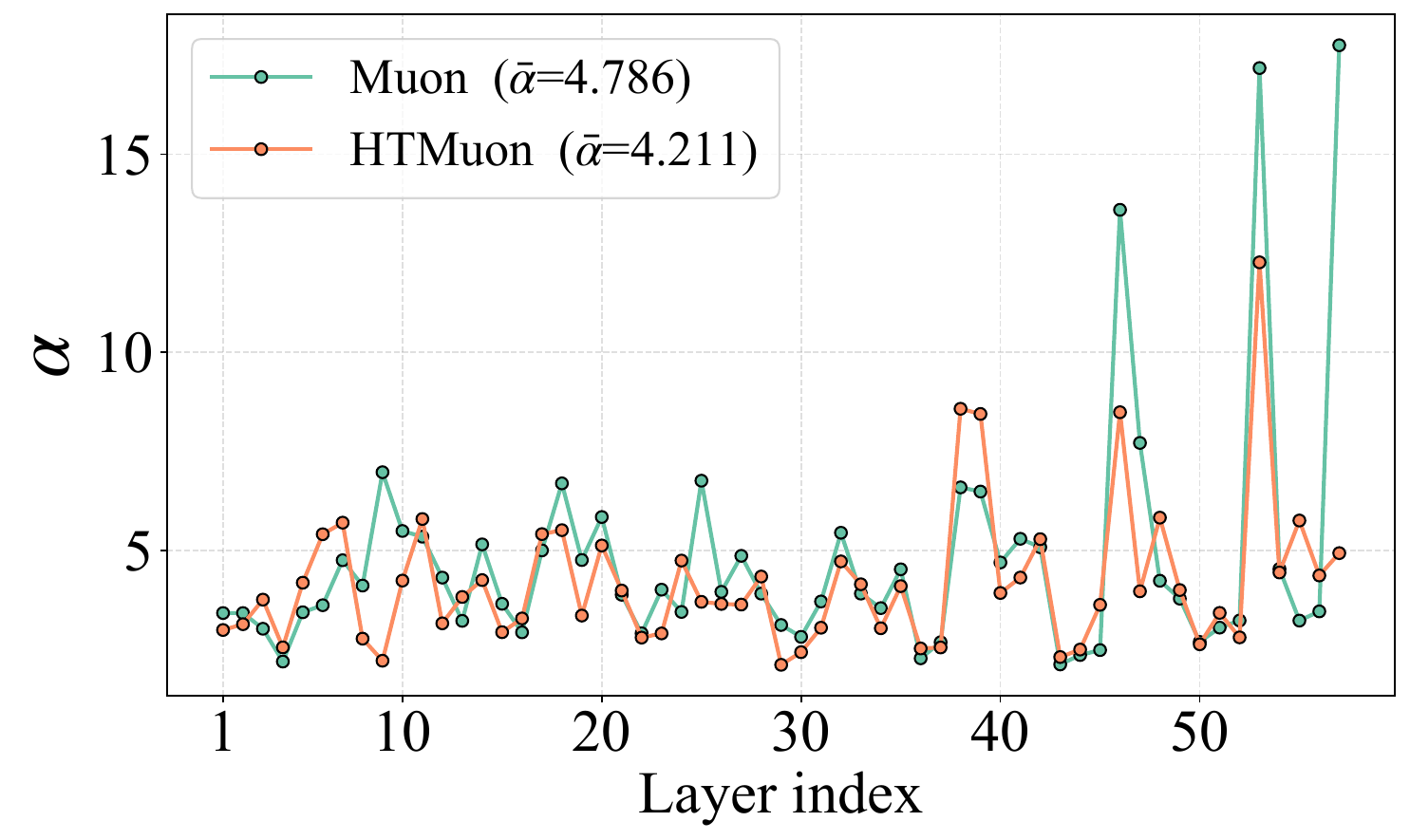}
        \caption{LLaMA-60M on C4}
        \label{fig:LLaMA-60M-C4-alpha}
    \end{subfigure}\hfill
    \begin{subfigure}[b]{0.24\linewidth}
        \centering
        \includegraphics[width=\linewidth]{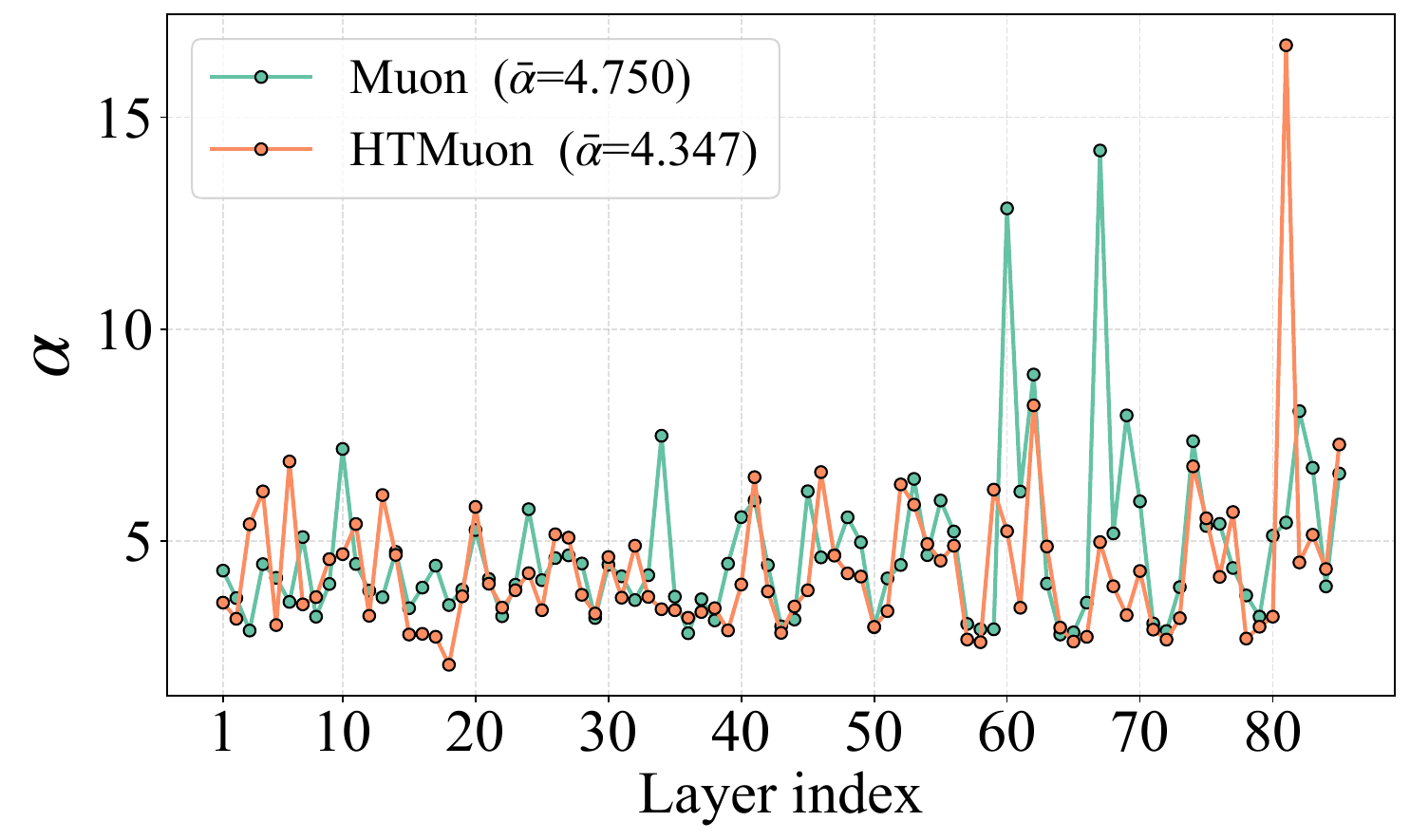}
        \caption{LLaMA-135M on C4}
        \label{fig:LLaMA-135M-alpha-C4}
    \end{subfigure} \hfill
    \begin{subfigure}[b]{0.24\linewidth}
        \centering
        \includegraphics[width=\linewidth]{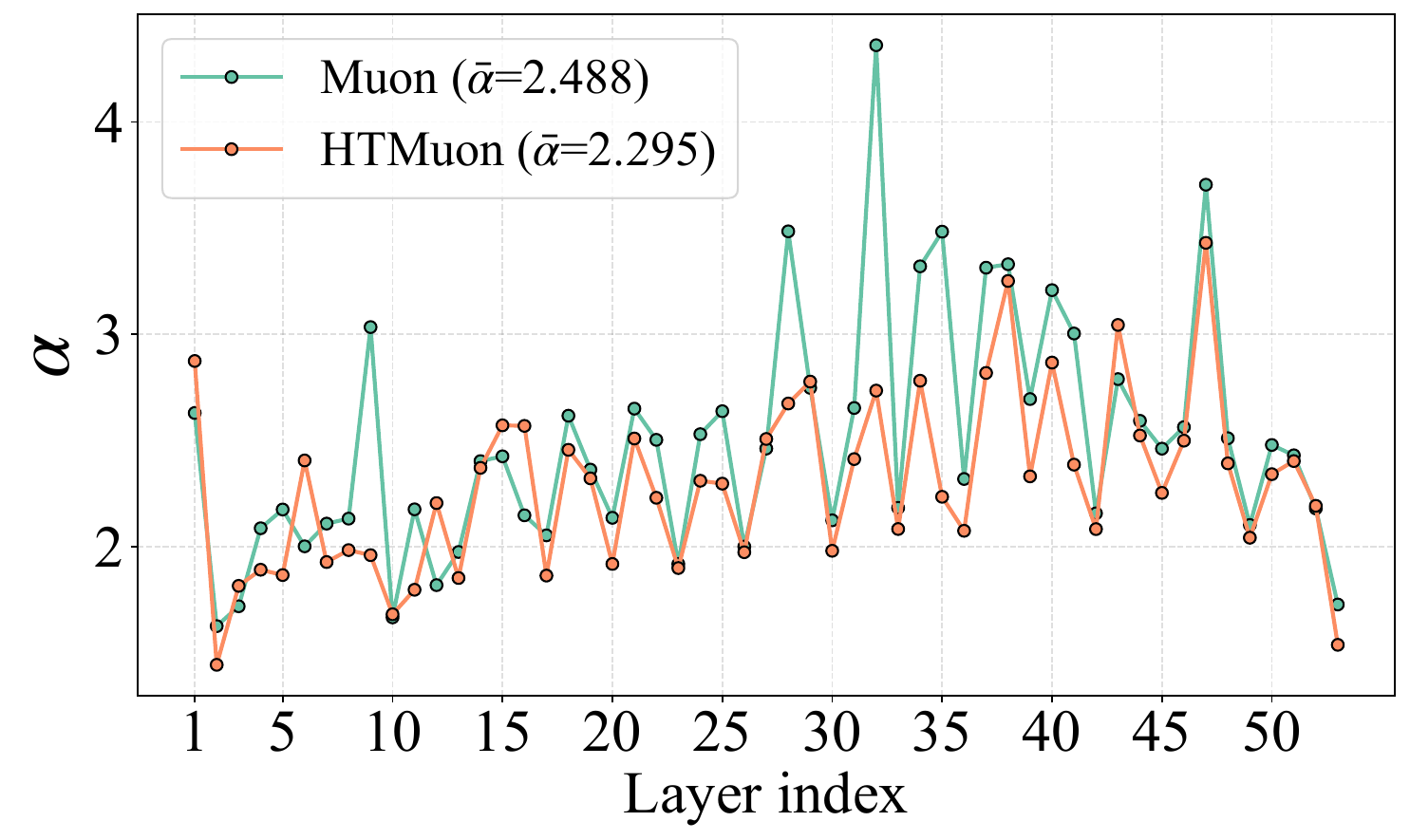}
        \caption{ResNet50 on CIFAR-100}
        \label{fig:Resnet50-alpha-cifar100-alpha}
    \end{subfigure}\hfill
    \begin{subfigure}[b]{0.24\linewidth}
        \centering
        \includegraphics[width=\linewidth]{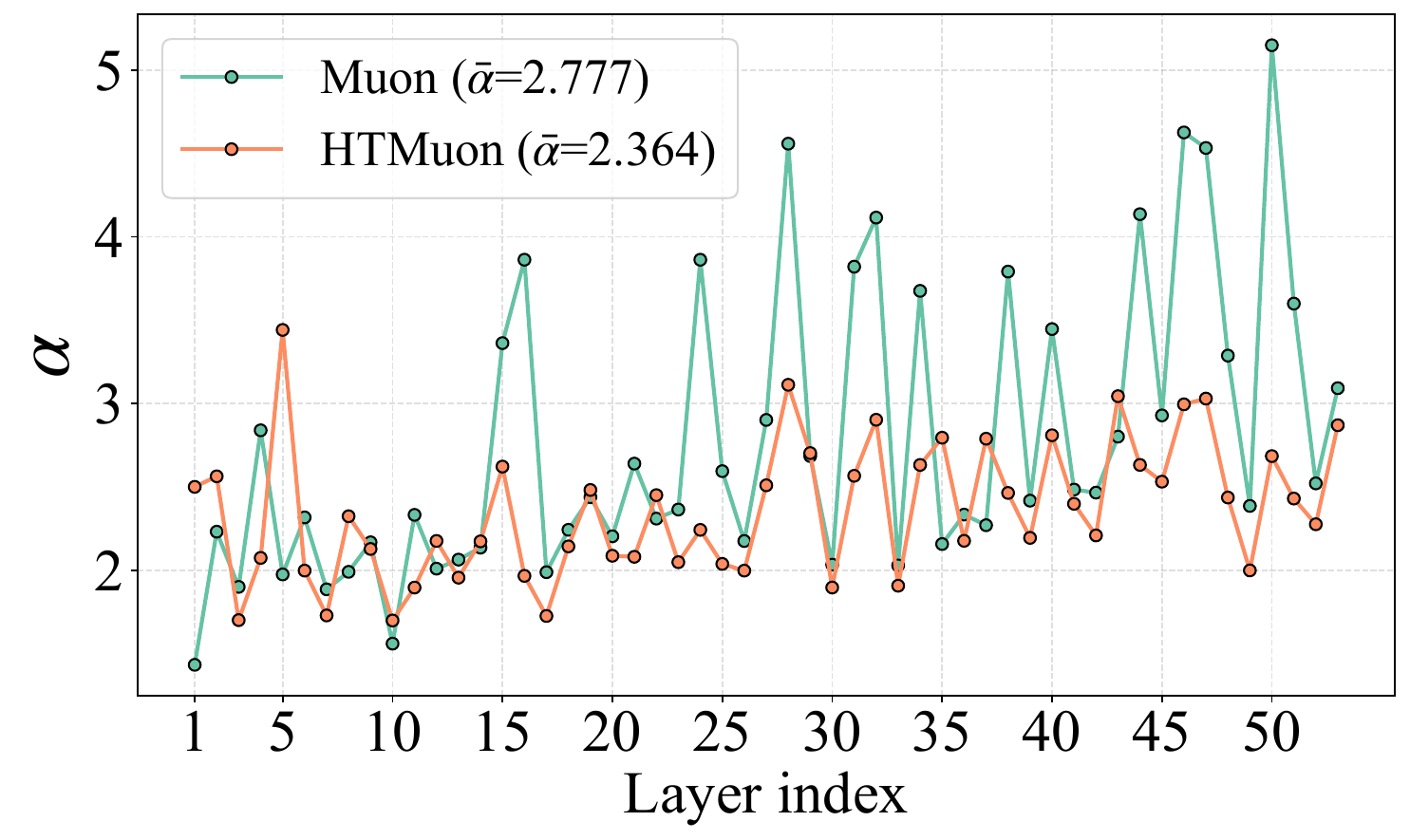}
        \caption{ResNet50 on CIFAR-10}
        \label{fig:Resnet50-alpha-cifar10-alpha}
    \end{subfigure}
    \caption{Layer-wise PL $\alpha$   for LLaMA and ResNet model weights trained with \texttt{Muon} and \texttt{HTMuon}. All models used for visualization are trained using each optimizer’s best-performing hyperparameter configuration. For hyperparameter configurations, please refer to Appendix~\ref{app:hyperparameter}. }
\label{fig:alpha}
\end{figure*}

\paragraph{\texttt{HTMuon} successfully produces more heavy-tailed weight matrices.} In  Figure~\ref{fig:alpha}, we  fit the layer-wise PL $\alpha$ for models trained with \texttt{Muon} and \texttt{HTMuon}. Across both LLaMA and ResNet models, we find that \texttt{HTMuon} yields smaller $\alpha$ than \texttt{Muon} for most layers, leading to a lower average $\bar{\alpha}$. Meanwhile, \texttt{HTMuon} achieves better performance, consistent with the HT-SR observation that better-trained models tend to have lower $\alpha$.  For more PL $\alpha$ plots, please refer to Figure~\ref{fig:Resnet18-alpha-cifar100-alpha} and~\ref{fig:Resnet18-alpha-cifar10-alpha} in Appendix~\ref{app:more_results}.

\begin{figure*}[!htb]
    \centering

    \begin{subfigure}[b]{0.24\linewidth}
        \centering
        \includegraphics[width=\linewidth]{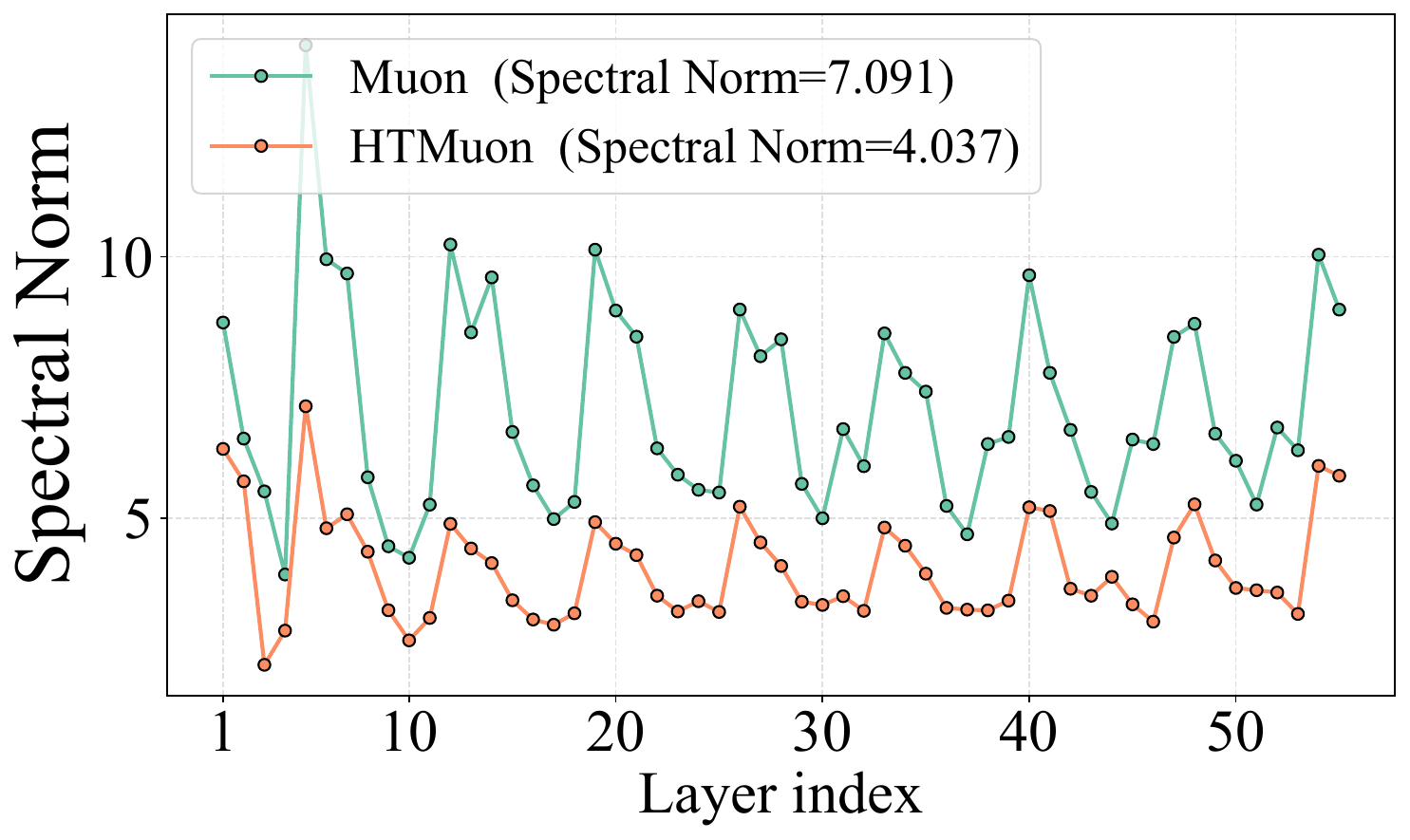}
        \caption{LLaMA-60M on C4}
        \label{fig:LLaMA-60M-spectral_norm}
    \end{subfigure}\hfill
    \begin{subfigure}[b]{0.24\linewidth}
        \centering
        \includegraphics[width=\linewidth]{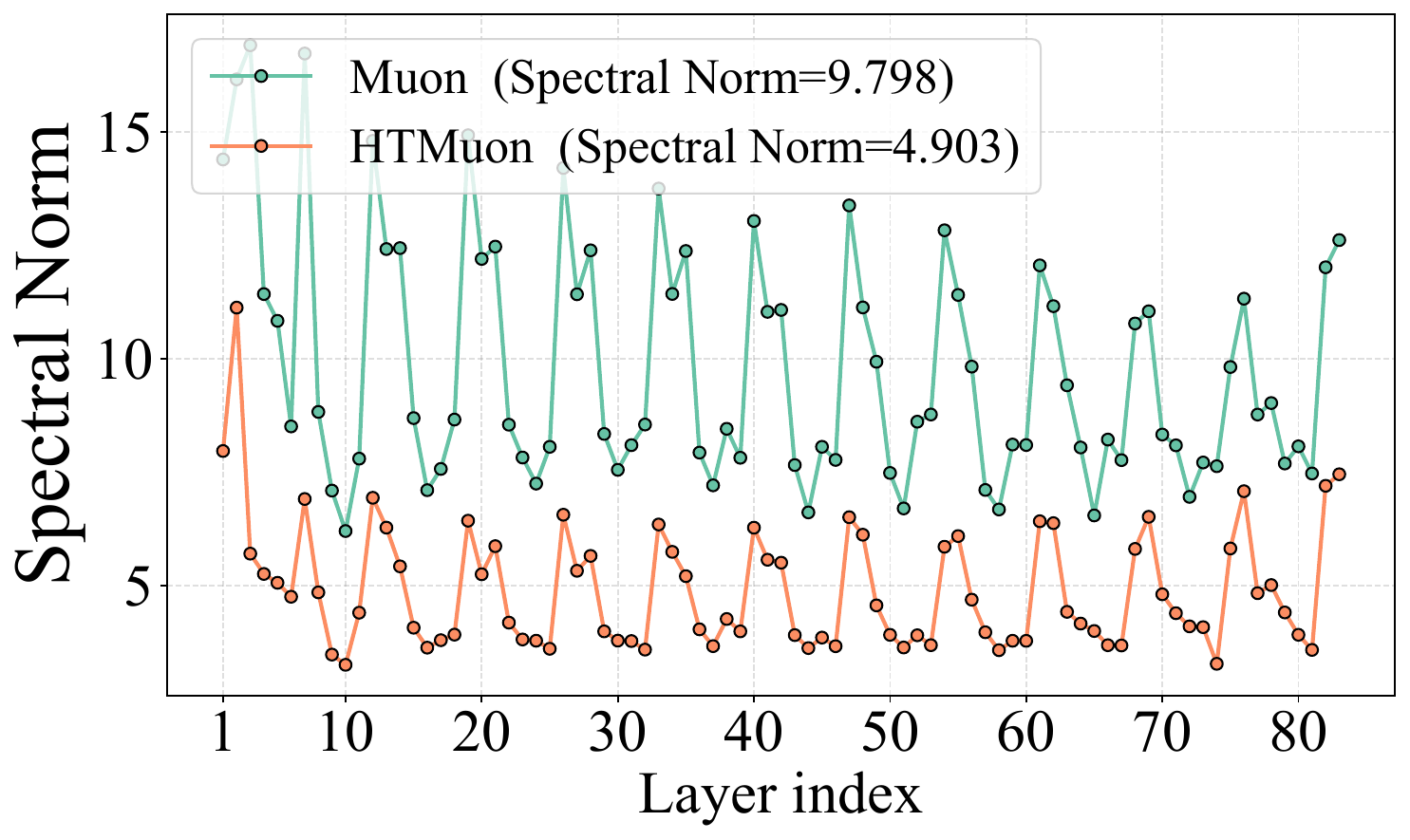}
        \caption{LLaMA-135M on C4}
        \label{fig:LLaMA-135M-spectral_norm}
    \end{subfigure} \hfill
  \begin{subfigure}[b]{0.24\linewidth}
        \centering
        \includegraphics[width=\linewidth]{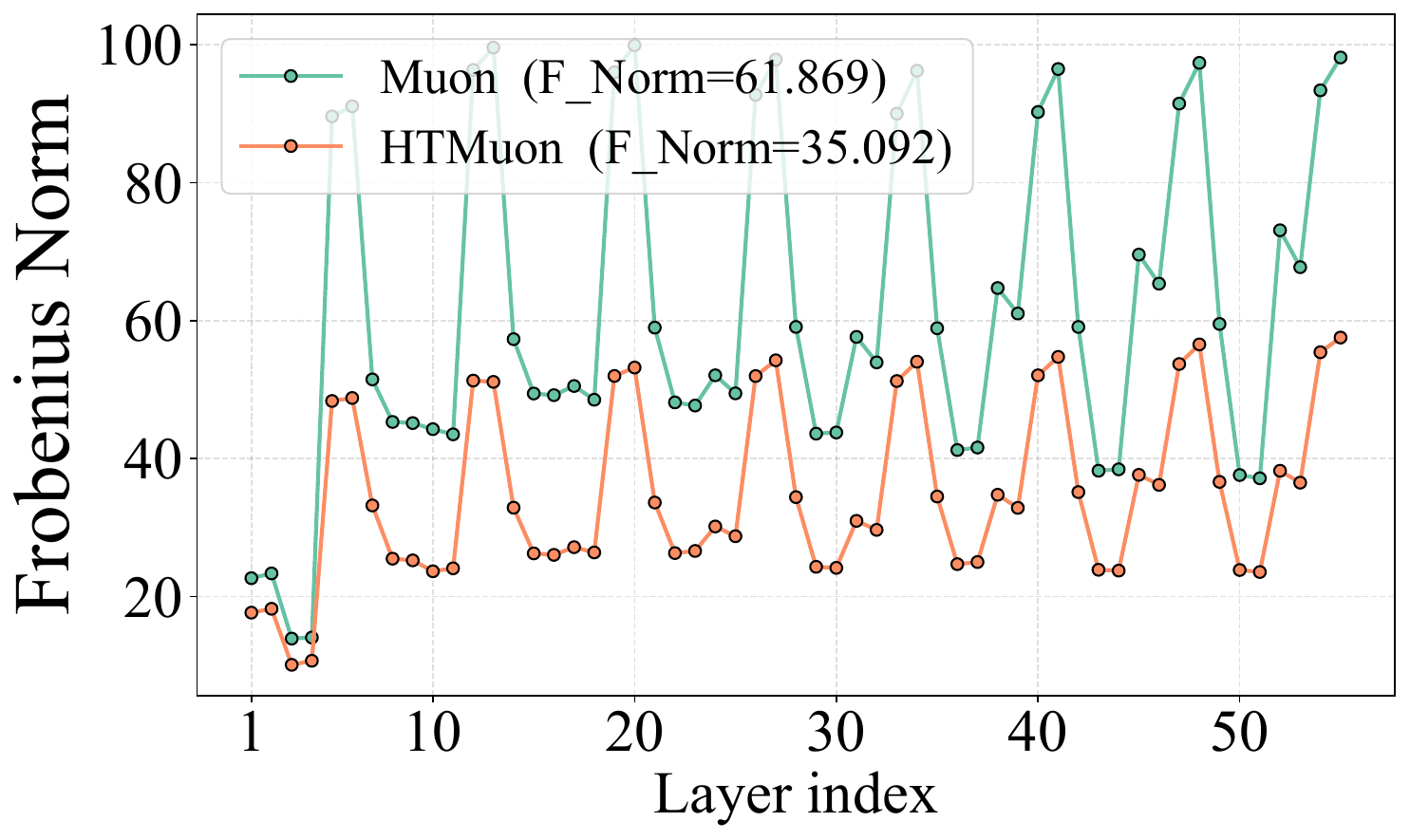}
        \caption{LLaMA-60M on C4}
        \label{fig:LLaMA-60M-f_norm}
    \end{subfigure}\hfill
    \begin{subfigure}[b]{0.24\linewidth}
        \centering
        \includegraphics[width=\linewidth]{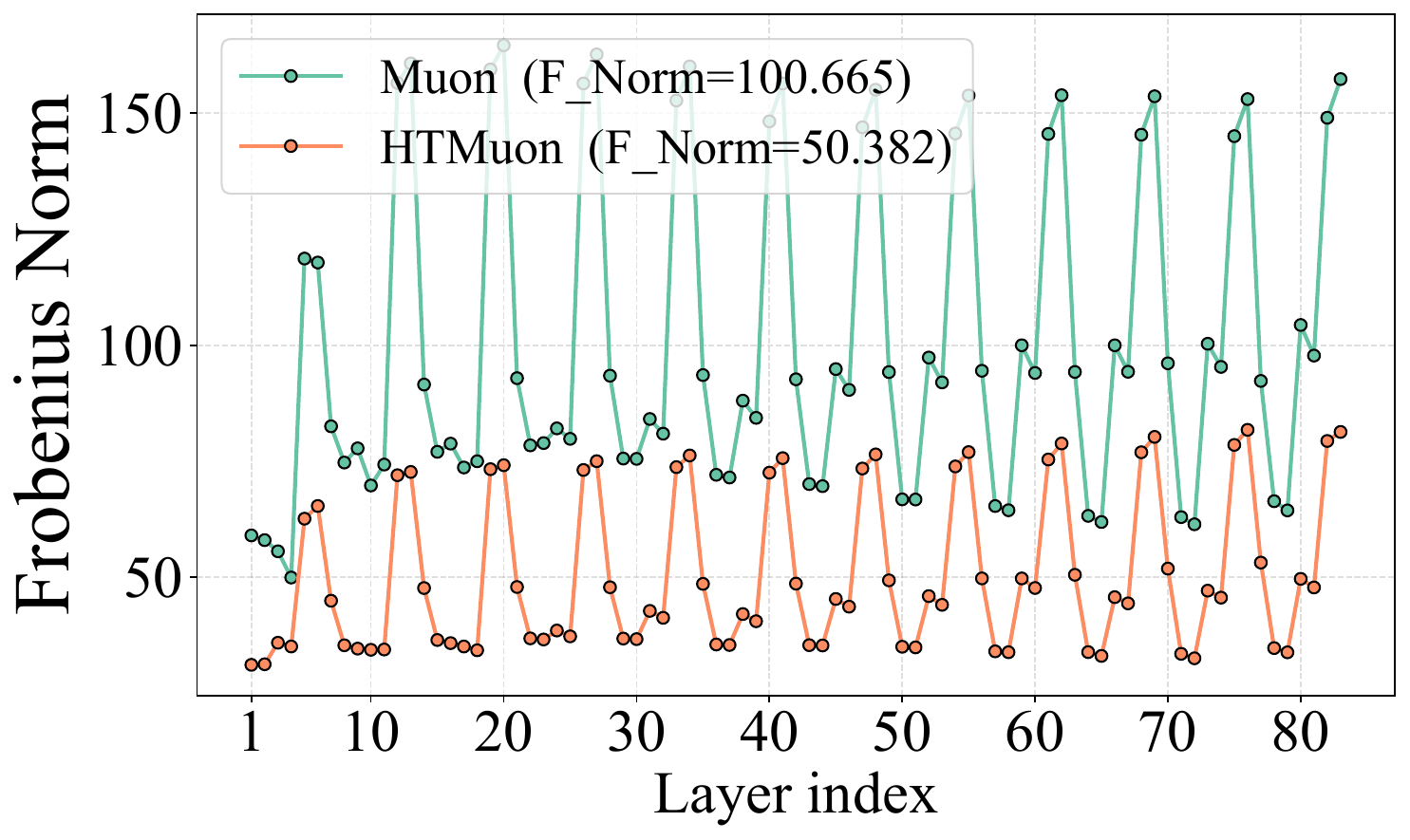}
        \caption{LLaMA-135M on C4}
        \label{fig:LLaMA-135M-f_norm}
    \end{subfigure}
    \caption{ Layer-wise spectral norm and frobenius norm   for LLaMA model weights   trained with \texttt{Muon} and \texttt{HTMuon}.  All models used for visualization are trained using each optimizer’s best-performing hyperparameter configuration. For hyperparameter configurations, please refer to Appendix~\ref{app:hyperparameter}.}
\label{fig:norm}
\end{figure*}

\paragraph{Additional generalization metrics.} In Figure~\ref{fig:norm}, we visualize the layer-wise spectral norm and Frobenius norm for LLaMA models trained with \texttt{Muon} and \texttt{HTMuon}. We find that, for these models, \texttt{HTMuon} consistently yields smaller norms than \texttt{Muon}. This is consistent with prior findings \citep{miyato2018spectral} that smaller spectral and Frobenius norms are often associated with better generalization.

\subsection{Downstream Tasks}
To further evaluate \texttt{HTMuon}'s applicability to broader NLP scenarios, in this section, we evaluate LLaMA-1B model trained with \texttt{HTMuon} on 7 commonsense tasks using the lm\_eval\_harness framework. For all tasks we use the default prompt and perform zero-shot evaluation. 

As shown in the Table~\ref{table:commonsense_task}, \texttt{HTMuon} achieves the best test accuracy for 4 out of 7 commonsense tasks, and achieves the best average score across 7 tasks, significantly outperforming the second best optimizer (\texttt{Muon}) by 1.05. This demonstrates the effectiveness of \texttt{HTMuon} on wider NLP scenarios.

\begin{table}[!t]
\centering
\caption{Zero-shot evaluation results ($\uparrow$) on seven commonsense reasoning benchmarks for the LLaMA-1B model pretrained with different methods.}
\scalebox{0.95}{
\begin{tabular}{@{}lcccccccc@{}}
\toprule
\textbf{Optimizer} & \textbf{ARC-c} & \textbf{ARC-e} & \textbf{PIQA} & \textbf{Hellaswag} & \textbf{OBQA} & \textbf{Winogrande} & \textbf{BOOLQ} & \textbf{Avg.} \\
\midrule
\texttt{Adam}   & 21.93 & 31.27 & 65.18 & 27.49 & 17.40 & \textbf{53.83} & 62.17 & 39.90 \\
\texttt{AdamW}  & 21.16 & 32.37 & 64.25 & 27.29 & \textbf{17.80} & 52.49 & \textbf{62.23} & 39.66 \\
\texttt{Muon}   & 20.56 & 32.66 & 65.13 & 31.53 & 17.00 & 51.22 & 62.14 & 40.03 \\
\texttt{HTMuon} & \textbf{22.18} & \textbf{33.16} & \textbf{66.59} & \textbf{33.76} & 17.20 & 52.72 & 61.93 & \textbf{41.08} \\
\bottomrule
\end{tabular}}
\label{table:commonsense_task}
\vspace{-5pt}
\end{table}
\subsection{Ablation Study}\label{sec:ablation study}

\paragraph{Varying different $p$.} Since $p$ is an additional hyperparameter introduced by \texttt{HTMuon}, we perform a grid search over $p$ on LLaMA models in Figure~\ref{fig:p}. We find that $p=0.125$ is a good choice, which is why we use $p=0.125$ in most experiments. In Appendix~\ref{More_abalation_study_p}, we  discuss more about the hyperparameter sensitivity for $p$, we show that although introducing 
$p$ adds a tuning dimension and add hyperparameter search burden, $p=0.125$ generalizes well across tasks and architectures and often can be optimal or near-optimal across tasks and architectures. We  suggest that $p=0.125$ can serve as a recommended default for other tasks.

\begin{figure}[!htb]
    \centering
    \begin{subfigure}{0.5\linewidth}
        \includegraphics[width=\textwidth]{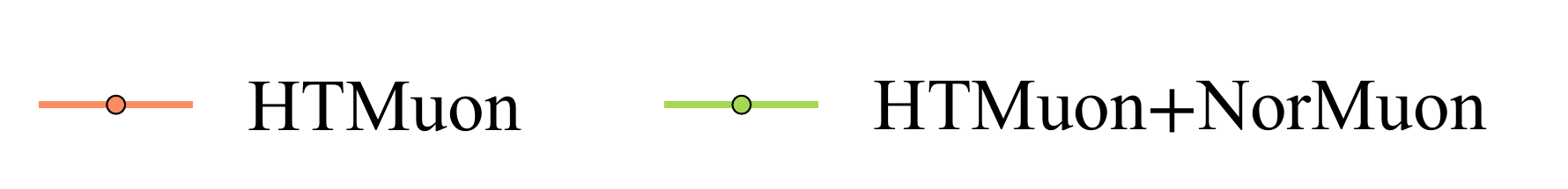}
    \end{subfigure}

    \begin{subfigure}[t]{0.3\linewidth}
        \includegraphics[width=\textwidth]{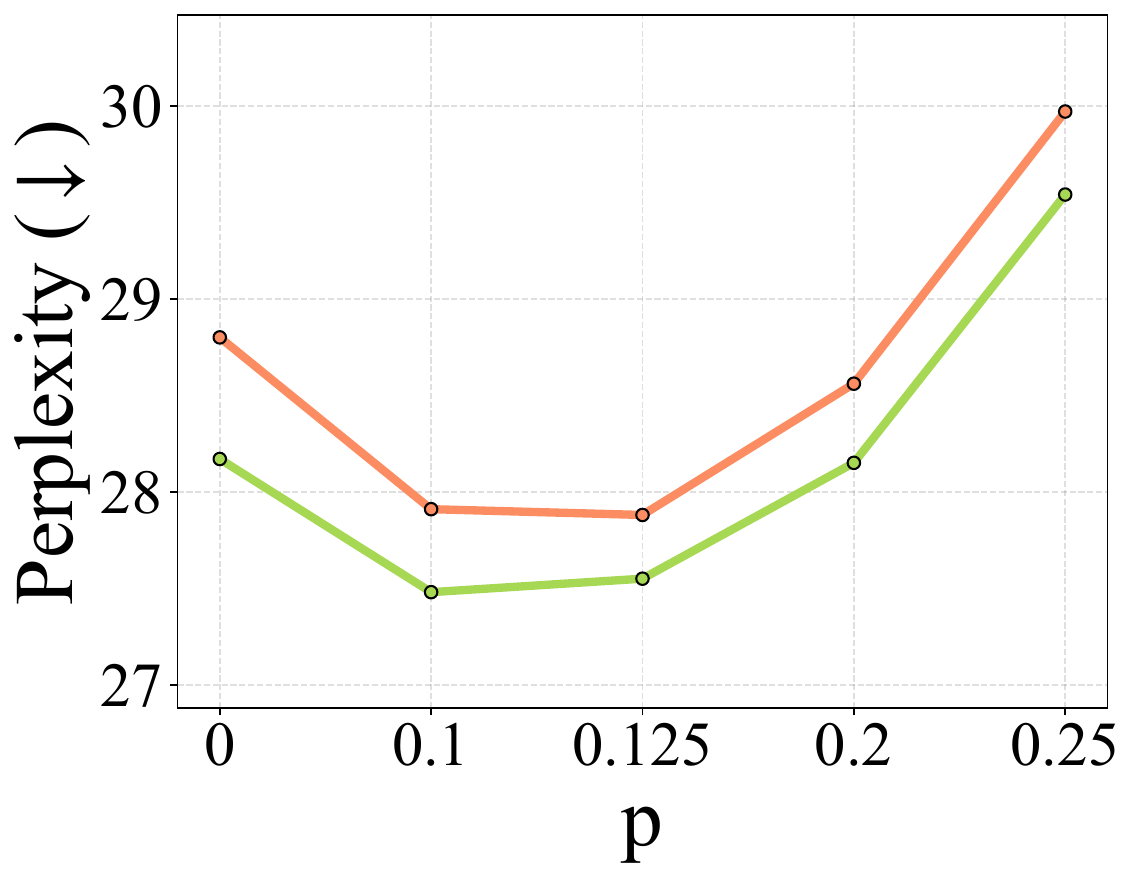}
        \caption{LLaMA-60M }
        \label{fig:LLaMA-60M_optimizer_p}
    \end{subfigure}
    \begin{subfigure}[t]{0.29\linewidth}
        \includegraphics[width=\textwidth]{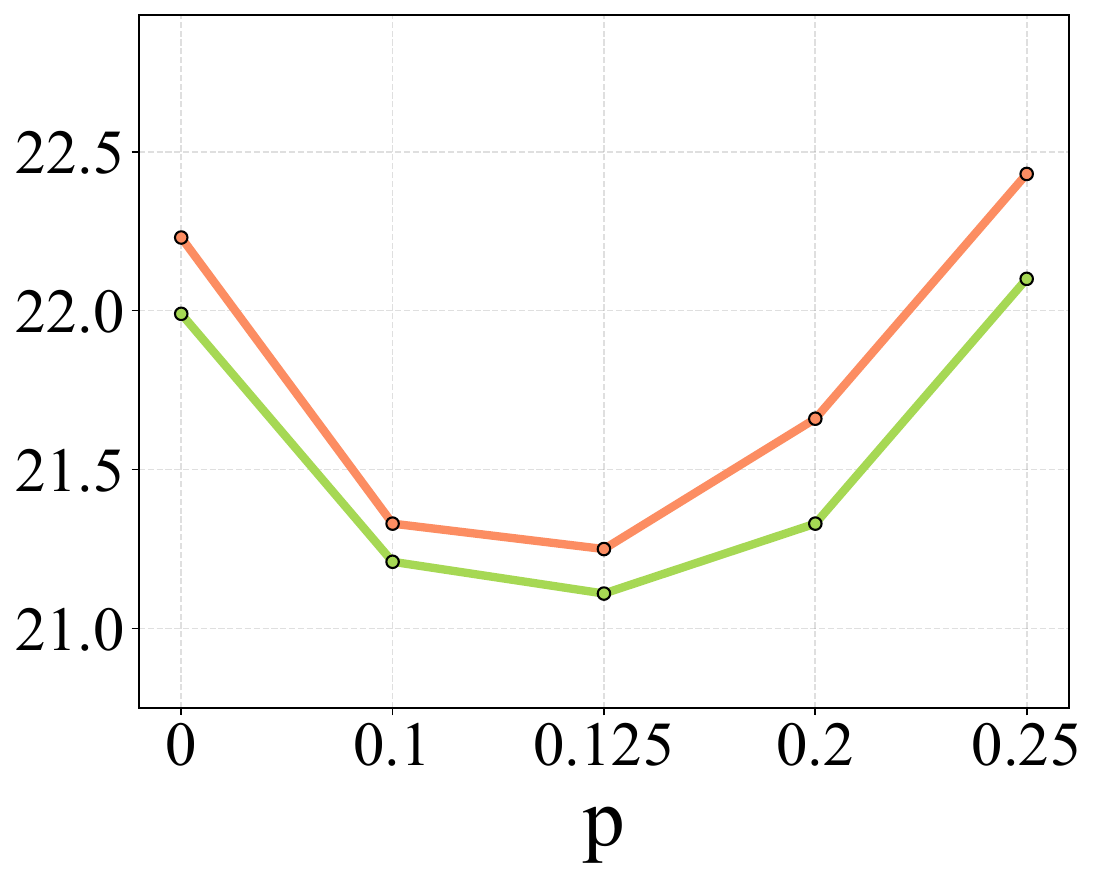}
        \caption{LLaMA-135M }
        \label{fig:LLaMA-130M_optimizer_p}
    \end{subfigure}
    \caption{ We conduct grid search on $p$. We find $p=0.125$ is a strong choice. Note that $p=0$ reduces to \texttt{Muon}. Detailed results provided in Table~\ref{Table:p} in Appendix~\ref{app:hyperparameter}.}
\label{fig:p}
\end{figure}

\paragraph{Varying learning rates.} In Figure~\ref{fig:lr}, we visualize performance across learning-rate grids for LLaMA models on C4 and ResNet models on CIFAR datasets. We find that, over most learning rates, \texttt{HTMuon} consistently outperforms \texttt{Muon} and \texttt{NorMuon}, demonstrating the stability of \texttt{HTMuon}. For more details, please refer to Table~\ref{Table:Lr grid_C4},~\ref{Table:Lr grid_Cifar100} and~\ref{Table:Lr grid_Cifar10} in Appendix~\ref{app:more_results}.

\begin{algorithm*}[ht]
\caption{\texttt{HTMuon\_HT}}
\label{algo:HTMUON_HT}
\begin{algorithmic}[1]
\State \textbf{Input:} Initial weights $\mathbf{W}_0 \in \mathbb{R}^{m\times n}$, loss function $L$,  learning rate $\eta$, momentum parameter $\beta$,  weight decay $\lambda$, power $\alpha$.\
\State Initialize $\mathbf{M}_0 \in\mathbb{R}^{m\times n} \leftarrow \mathbf{0}$
\For{$t=1,2,\ldots$}
\State $\mathbf{G}_t \leftarrow \nabla_{\mathbf{W}}L(\mathbf{W}_t)$
\State $\mathbf{M}_t \leftarrow \beta\mathbf{M}_{t-1} + (1-\beta)\mathbf{G}_t$
\State $\mU_t,\mSigma_t, {\mV_t}^{\top} \leftarrow \texttt{SVD}(\mathbf{M}_t)$
\State ${\mSigma_t}_{ii} \leftarrow i^{-\alpha}$
\State $\mO_t \leftarrow \mU_t\mSigma_t {\mV_t}^{\top} $
\State $s \leftarrow \sqrt{\max \left( 1, \frac{m}{n}\right)}$
\State $\mathbf{W}_{t+1} \leftarrow \mathbf{W}_t - \eta\lambda\mathbf{W}_t - \eta s\mO_t$
\EndFor
\end{algorithmic}
\end{algorithm*}

\paragraph{Varying different ways to make \texttt{Muon} update more heavy-tailed.} To better prove making \texttt{Muon} update more heavy-tailed can improve performance, we design another optimizer \texttt{HTMuon\_HT} in Algorithm~\ref{algo:HTMUON_HT}. Specifically, at each step, \texttt{HTMuon\_HT} directly replaces the singular values of the momentum matrix with a heavy-tailed sequence $\{i^{-\alpha}\}_{i=1}^{m}$. As shown in Table~\ref{Table:HTMuon_HT}, although \texttt{HTMuon\_HT} is not as strong as \texttt{HTMuon}, it still outperforms \texttt{Muon} on LLaMA models. This shows that merely inducing a heavy-tailed spectrum already yields non-trival gains, and our algorithm  \texttt{HTMuon} further show that heavy tail spectral correction on top of preserved spectral information is more effective, we put more discussion about the performance improvement attribution in Appendix~\ref{app:more_ablation_study_on_HTMuon_HT}.  In Figure~\ref{fig:HTMUON_HT_60M-alpha} and~\ref{fig:HTMUON_HT_135M-alpha}, we visualize the layer-wise PL $\alpha$ for \texttt{HTMuon\_HT}, and as intended, it yields lower $\alpha$ than \texttt{Muon} across layers.

\section{Theoretical Analysis}
In this section, we provide theoretical understandings of \texttt{HTMuon}.

\paragraph{Notation.} For a vector $\vv$, we denote its $l_2$ norm as $\|\vv\|_2$. For a matrix $\mA,\mB\in \R^{m\times n}$, we denote its  spectral norm as $\|\mA\|$, Frobenius norm as $\|\mA\|_\F$,  nuclear norm as $\|\mA\|_*$, where $\|\mA\|_*=\sum_{i=1}^r \sigma_i$. Here, $\{\sigma_i\}_{i=1}^r$ are  singular values. For $p>1$, we denote Schatten$-p$ norm as $\|\mA\|_p$, where $\|\mA\|_p=\left(\sum_{i=1}^r \sigma_i^p\right)^{\frac{1}{p}}$.  We use $\langle \mA, \mB\rangle$ to denote the inner product between $\mA$ and $\mB$, i.e.,
$\langle \mA, \mB\rangle = \operatorname{Tr }\left(\mA^{\top}\mB\right)$.

\subsection{Update Under the Schatten$-q$ Norm Constraint}
We are here to provide a view of Schatten norm constrained optimization for \texttt{HTMuon}. To determine the descent direction at step $t$, we formulate the update as the following constrained minimization problem:
\begin{align}\label{eq:Schatten_Norm_problem}
    &\min_{\Delta\mW\in \mathbb{R}^{m\times n}} \quad f(\mW_t+\Delta \mW) \\
    &\text{subject to} \quad \|\Delta \mW\|_q \leq \delta \notag
\end{align}
where $\|\cdot\|_q$ denotes the Schatten$-q$ norm. 
When we employ the first-order approximation for our optimization objective 
\begin{equation}
f(\mW_t + \Delta \mW) \approx f(\mW_t) + \operatorname{Tr}(\nabla f(\mW_t)^\top \Delta \mW) 
\end{equation}
It is straightforward to derive that optimizing this approximation is equivalent to the following objective:
\begin{align}\label{eq:equivalent_problem}
&\max_{\Delta \mW}    \quad  \operatorname{Tr}(\widetilde{\mG_t}^\top \Delta \mW) \\
&\text{subject to}       \quad      \|\Delta \mW\|_q \leq \delta, \notag
\end{align}
where $\widetilde{\mG_t}  = -\nabla f(\mW_t)$ denotes the negative gradient.

\begin{theorem}\label{Theorem:Schatten$-q$ Norm Constraint}
Let $\widetilde{\mG_t} =\mU\mSigma\mV^\top$ be the \texttt{SVD} of the negative gradient $\widetilde{\mG_t}  = -\nabla f(\mW_t)$, where $\mSigma = \operatorname{diag}(\sigma_1, \dots, \sigma_r)$. Let $p$ be the conjugate exponent of $q$, satisfying $1/p^{\prime} + 1/q = 1$. The explicit solution $\Delta \mW^*$ to the optimization problem~\ref{eq:equivalent_problem} is given by:
\begin{equation}
    \Delta \mW^* = \delta \frac{1}{\|\widetilde{\mG_t} \|_{p^{\prime}}^{p^{\prime}/q}} \mU \mSigma^{p^{\prime}-1} \mV^\top,
\end{equation}
where $\mSigma^{p^{\prime}-1} = \operatorname{diag}(\sigma_1^{p^{\prime}-1}, \dots, \sigma_r^{p^{\prime}-1})$.
\end{theorem}

We put proof in Appendix~\ref{proof:Schatten$-q$ Norm Constraint}. Based on Theorem~\ref{Theorem:Schatten$-q$ Norm Constraint}, when $q \in (2, \infty), p^{\prime} \in (1,2)$, we can assume $p=p^{\prime}-1$ and find the $\Delta \mW^*$ is exactly the update by $\texttt{HTMuon}$, which means \texttt{HTMuon} is equivalent to steepest descent under a Schatten$-q$ norm constraint. It is well known  that \texttt{Muon}  can be viewed as the steepest descent under the Schatten$-\infty$ norm constraint \citep{bernstein2024modular,pethick2025training}, so  \texttt{HTMuon} generalizes \texttt{Muon}’s equivalence to steepest descent under a Schatten$-\infty$ norm constraint and this will also benefit training \citep{davis2025spectral}. 

\subsection{Convergence Analysis}
Here we give the convergence analysis under a smooth non-convex setting for finding $\epsilon$-stationary points, i.e. $ \|\nabla f(\mW)\|_{*} \leq \epsilon$. We consider the following stochastic optimization problem: 
\begin{equation}
    \min_{\mW\in \R^{m\times n}} f(\mW)=\E_{\vxi}[f(\mW;\vxi)],
\end{equation}
where $\vxi$ is stochastic noise. We denote $f^*=\inf_\mW f(\mW)$, $r=\min\{m,n\}$. We assume $f^*>-\infty$. 

For ease of theoretical analysis, we rewrite \texttt{HTMuon} in a stochastic setting and present it in Algorithm~\ref{alg:HTMUON_Stochastic}.

We assume $f$ is an $L$ Frobenius norm Lipchitz smooth function. We assume that $\nabla f(\mW;\xi)$ is an unbiased stochastic estimator of the true gradient $\nabla f(\mW)$ and has a bounded variance. We also assume that $\mM_t$'s singular values are bounded by $l$.

We then give assumptions below, some assumptions and definitions are adapted from \citet{shen2025convergence}.
\begin{assumption}[F-norm Lipschitz smooth ]\label{ass: smooth}
Define $f:\R^{m\times n}\to \R$ is $L$ Frobenius norm Lipschitz smooth if for any $\mW, \mW'\in \R^{m\times n}$, we have
\begin{align*}
    \|\nabla f(\mW)-\nabla f(\mW')\|_\F\leq L\|\mW-\mW'\|_\F.
\end{align*}
\end{assumption}

\begin{assumption}[Bounded variance]\label{ass: variance}
    We assume $\nabla f(\mW;\xi)$ is an unbiased stochastic estimator of the true gradient $\nabla f(\mW)$ and has a bounded variance, i.e.
    \begin{align*}
        &\E[\nabla f(\mW;\xi)]=\nabla f(\mW), \\
        &\E\|\nabla f(\mW;\xi)-\nabla f(\mW)\|_\F^2]\leq \sigma^2.
    \end{align*}
\end{assumption}

\begin{assumption}[Bounded singular value]\label{ass: bounded_sv}
    We assume the largest singular value $\sigma_{\max}(\mM_t) \le l$. 
\end{assumption}

\begin{definition}\label{Def:nuclear norm stationary point}
  We say $\mW$ is an $\epsilon$-nuclear norm stationary point of $f$ if $\|\nabla f( \mW)\|_*\leq\epsilon$.
\end{definition}

\begin{theorem}[\texttt{HTMuon}]\label{thm: htmuon_nonconvex}
Under Assumptions ~\ref{ass: smooth}, ~\ref{ass: variance} and~\ref{ass: bounded_sv}, if we apply \texttt{HTMuon} in Algorithm~\ref{alg:HTMUON_Stochastic} with adaptive learning rate  $\eta_t=\frac{\langle  \mM_t,\rho(\mM_t)\rangle  }{L\|\rho(\mM_t)\|^2_\F}$, $\rho(\mM_t)=\mU_t\mSigma_t^{p}\mV_t^{\top}$, $\eta=\max\{\eta_t\}_{t=1}^T$, batch size $B=T$ and $\beta=\frac{1}{\sqrt{T}}$, $\Delta=f(\mW_0)-f^*$,  we have 
\begin{align*}
       &\frac{1}{T}\sum_{t=1}^T\E \bigl[ \|\nabla f(\mW_t)\|^2_{1+p} \bigr] \leq \frac{2Lr^{\frac{1}{s}}}{1-\varepsilon^{\prime}}\left( \frac{1-\varepsilon^{\prime}}{L}+\frac{1}{L\varepsilon^{\prime}}\right)\bigl[\frac{2\sigma}{T}+\frac{4\sigma^2}{T^3} +\frac{4rl^{\frac{2}{q}}L^2\eta^2}{T} \bigr]+ \frac{2Lr^{\frac{1}{s}}}{1-\varepsilon^{\prime}}\frac{\Delta }{T},
\end{align*}
where $s=\frac{1+p}{1-p}$ and $0<\varepsilon^{\prime}<1$. 
\end{theorem}

We put proof in Appendix~\ref{proof: Convergence Analysis}. Based on  Theorem~\ref{thm: htmuon_nonconvex} and Definition~\ref{Def:nuclear norm stationary point}, when $B=T$, the sample complexity upper bound  is  $O(\epsilon^{-4})$, which matches \texttt{Muon}'s sample complexity  in Appendix Theorem~\ref{thm: muon_nonconvex} and \texttt{SGDM}'s sample complexity in \citet{garrigos2023handbook,arjevani2023lower}. 

\section{Conclusion}

In this work, motivated by HT-SR theory, we introduce \texttt{HTMuon} optimizer. \texttt{HTMuon} maintains \texttt{Muon}’s capacity to model parameter interdependencies, while yielding more heavy-tailed updates and promoting more heavy-tailed weight spectra. It delivers consistent gains in accuracy and training stability on LLM pretraining and image classification, and can be seamlessly integrated into existing \texttt{Muon} variants. To reduce the \texttt{SVD} overhead in \texttt{HTMuon}, we introduce two acceleration implementations that further lower runtime cost while still outperforming \texttt{Muon}. Finally, we establish a theoretical connection to steepest descent under a Schatten-$q$ norm constraint and provide convergence analysis for smooth non-convex settings.

\section*{Limitations}
Despite achieving improvements in LLM pretraining and image classification, \texttt{HTMuon} has some potential limitations. Due to limited computational resources, we have not evaluated \texttt{HTMuon} on larger models than 1B or larger-scale datasets; we leave this to future work. In addition, \texttt{HTMuon} and \texttt{HTMuon\_NS} incur slightly higher per-step runtime overhead than \texttt{Muon}. While using \texttt{HTMuon} with larger update intervals can reduce this cost, achieving stronger performance typically requires more frequent \texttt{HTMuon} updates, which increases the overall runtime. Exploring better numerical methods to accelerate \texttt{HTMuon} is  also part of future work.

\section*{Acknowledgments}

We thank our colleagues and funding agencies. This work is supported by the DARPA AIQ and DARPA DIAL programs, the U.S. Department of Energy under Award Number DE-SC0025584, Dartmouth College, and Lambda AI.

\bibliographystyle{plainnat}
\bibliography{custom}

@article{arjevani2023lower,
  title={Lower bounds for non-convex stochastic optimization},
  author={Arjevani, Yossi and Carmon, Yair and Duchi, John C and Foster, Dylan J and Srebro, Nathan and Woodworth, Blake},
  journal={Mathematical Programming},
  volume={199},
  number={1},
  pages={165--214},
  year={2023},
  publisher={Springer}
}

@article{shen2025convergence,
  title={On the convergence analysis of muon},
  author={Shen, Wei and Huang, Ruichuan and Huang, Minhui and Shen, Cong and Zhang, Jiawei},
  journal={arXiv preprint arXiv:2505.23737},
  year={2025}
}

@article{kaddour2022flat,
  title={When do flat minima optimizers work?},
  author={Kaddour, Jean and Liu, Linqing and Silva, Ricardo and Kusner, Matt J},
  journal={Advances in Neural Information Processing Systems},
  volume={35},
  pages={16577--16595},
  year={2022}
}

@article{foret2020sharpness,
  title={Sharpness-aware minimization for efficiently improving generalization},
  author={Foret, Pierre and Kleiner, Ariel and Mobahi, Hossein and Neyshabur, Behnam},
  journal={arXiv preprint arXiv:2010.01412},
  year={2020}
}

@article{kingma2014adam,
  title={Adam: A method for stochastic optimization},
  author={Kingma, Diederik P and Ba, Jimmy},
  journal={arXiv preprint arXiv:1412.6980},
  year={2014}
}

@article{loshchilov2017decoupled,
  title={Decoupled weight decay regularization},
  author={Loshchilov, Ilya and Hutter, Frank},
  journal={arXiv preprint arXiv:1711.05101},
  year={2017}
}

@article{wang2025sharpness,
  title={The sharpness disparity principle in transformers for accelerating language model pre-training},
  author={Wang, Jinbo and Wang, Mingze and Zhou, Zhanpeng and Yan, Junchi and Wu, Lei},
  journal={arXiv preprint arXiv:2502.19002},
  year={2025}
}

@misc{zhou2023temperaturebalancinglayerwiseweight,
      title={Temperature Balancing, Layer-wise Weight Analysis, and Neural Network Training}, 
      author={Yefan Zhou and Tianyu Pang and Keqin Liu and Charles H. Martin and Michael W. Mahoney and Yaoqing Yang},
      year={2023},
      eprint={2312.00359},
      archivePrefix={arXiv},
      primaryClass={cs.LG},
      url={https://arxiv.org/abs/2312.00359}, 
}

@article{liu2024model,
  title={Model balancing helps low-data training and fine-tuning},
  author={Liu, Zihang and Hu, Yuanzhe and Pang, Tianyu and Zhou, Yefan and Ren, Pu and Yang, Yaoqing},
  journal={arXiv preprint arXiv:2410.12178},
  year={2024}
}

@article{li2025normuon,
  title={NorMuon: Making Muon more efficient and scalable},
  author={Li, Zichong and Liu, Liming and Liang, Chen and Chen, Weizhu and Zhao, Tuo},
  journal={arXiv preprint arXiv:2510.05491},
  year={2025}
}

@article{liu2025cosmos,
  title={Cosmos: A hybrid adaptive optimizer for memory-efficient training of llms},
  author={Liu, Liming and Xu, Zhenghao and Zhang, Zixuan and Kang, Hao and Li, Zichong and Liang, Chen and Chen, Weizhu and Zhao, Tuo},
  journal={arXiv preprint arXiv:2502.17410},
  year={2025}
}

@article{vyas2024soap,
  title={Soap: Improving and stabilizing shampoo using adam},
  author={Vyas, Nikhil and Morwani, Depen and Zhao, Rosie and Kwun, Mujin and Shapira, Itai and Brandfonbrener, David and Janson, Lucas and Kakade, Sham},
  journal={arXiv preprint arXiv:2409.11321},
  year={2024}
}

@article{yuan2024mars,
  title={Mars: Unleashing the power of variance reduction for training large models},
  author={Yuan, Huizhuo and Liu, Yifeng and Wu, Shuang and Zhou, Xun and Gu, Quanquan},
  journal={arXiv preprint arXiv:2411.10438},
  year={2024}
}

@article{si2025adamuon,
  title={Adamuon: Adaptive muon optimizer},
  author={Si, Chongjie and Zhang, Debing and Shen, Wei},
  journal={arXiv preprint arXiv:2507.11005},
  year={2025}
}

@article{liang2024cautious,
  title={Cautious optimizers: Improving training with one line of code},
  author={Liang, Kaizhao and Chen, Lizhang and Liu, Bo and Liu, Qiang},
  journal={arXiv preprint arXiv:2411.16085},
  year={2024}
}

@inproceedings{
chen2023symbolic,
title={Symbolic Discovery of Optimization Algorithms},
author={Xiangning Chen and Chen Liang and Da Huang and Esteban Real and Kaiyuan Wang and Hieu Pham and Xuanyi Dong and Thang Luong and Cho-Jui Hsieh and Yifeng Lu and Quoc V Le},
booktitle={Thirty-seventh Conference on Neural Information Processing Systems},
year={2023},
url={https://openreview.net/forum?id=ne6zeqLFCZ}
}

@misc{jordan2024muon,
  author       = {Keller Jordan and Yuchen Jin and Vlado Boza and Jiacheng You and
                  Franz Cesista and Laker Newhouse and Jeremy Bernstein},
  title        = {Muon: An optimizer for hidden layers in neural networks},
  year         = {2024},
  url          = {https://kellerjordan.github.io/posts/muon/}
}

@article{liu2025muon,
  title={Muon is scalable for LLM training},
  author={Jingyuan Liu and Jianlin Su and Xingcheng Yao and Zhejun Jiang and Guokun Lai and Yulun Du and Yidao Qin and Weixin Xu and Enzhe Lu and Junjie Yan and Yanru Chen and Huabin Zheng and Yibo Liu and Shaowei Liu and Bohong Yin and Weiran He and Han Zhu and Yuzhi Wang and Jianzhou Wang and Mengnan Dong and Zheng Zhang and Yongsheng Kang and Hao Zhang and Xinran Xu and Yutao Zhang and Yuxin Wu and Xinyu Zhou and Zhilin Yang},
  journal={arXiv preprint arXiv:2502.16982},
  year={2025}
}

@article{chen2025muon,
  title={Muon Optimizes Under Spectral Norm Constraints},
  author={Chen, Lizhang and Li, Jonathan and Liu, Qiang},
  journal={arXiv preprint arXiv:2506.15054},
  year={2025}
}

@article{shah2025practical,
  title={Practical efficiency of muon for pretraining},
  author={Ishaan Shah and Anthony M. Polloreno and Karl Stratos and Philip Monk and Adarsh Chaluvaraju and Andrew Hojel and Andrew Ma and Anil Thomas and Ashish Tanwer and Darsh J Shah and Khoi Nguyen and Kurt Smith and Michael Callahan and Michael Pust and Mohit Parmar and Peter Rushton and Platon Mazarakis and Ritvik Kapila and Saurabh Srivastava and Somanshu Singla and Tim Romanski and Yash Vanjani and Ashish Vaswani},
  journal={arXiv preprint arXiv:2505.02222},
  year={2025}
}

@article{wen2025fantastic,
  title={Fantastic pretraining optimizers and where to find them},
  author={Wen, Kaiyue and Hall, David and Ma, Tengyu and Liang, Percy},
  journal={arXiv preprint arXiv:2509.02046},
  year={2025}
}

@article{semenov2025benchmarking,
  title={Benchmarking optimizers for large language model pretraining},
  author={Semenov, Andrei and Pagliardini, Matteo and Jaggi, Martin},
  journal={arXiv preprint arXiv:2509.01440},
  year={2025}
}

@article{zeng2025glm,
  title={Glm-4.5: Agentic, reasoning, and coding (arc) foundation models},
  author={Zeng, Aohan and Lv, Xin and Zheng, Qinkai and Hou, Zhenyu and Chen, Bin and Xie, Chengxing and Wang, Cunxiang and Yin, Da and Zeng, Hao and Zhang, Jiajie and others},
  journal={arXiv preprint arXiv:2508.06471},
  year={2025}
}

@article{team2025kimi,
  title={Kimi k2: Open agentic intelligence},
  author={Team, Kimi and Bai, Yifan and Bao, Yiping and Chen, Guanduo and Chen, Jiahao and Chen, Ningxin and Chen, Ruijue and Chen, Yanru and Chen, Yuankun and Chen, Yutian and others},
  journal={arXiv preprint arXiv:2507.20534},
  year={2025}
}

@article{zhang2024adam,
  title={Adam-mini: Use fewer learning rates to gain more},
  author={Zhang, Yushun and Chen, Congliang and Li, Ziniu and Ding, Tian and Wu, Chenwei and Kingma, Diederik P and Ye, Yinyu and Luo, Zhi-Quan and Sun, Ruoyu},
  journal={arXiv preprint arXiv:2406.16793},
  year={2024}
}

@inproceedings{groeneveld2024olmo,
  title={OLMo: Accelerating the science of language models},
    author = "Groeneveld, Dirk  and
      Beltagy, Iz  and
      Walsh, Evan  and
      Bhagia, Akshita  and
      Kinney, Rodney  and
      Tafjord, Oyvind  and
      Jha, Ananya  and
      Ivison, Hamish  and
      Magnusson, Ian  and
      Wang, Yizhong  and
      Arora, Shane  and
      Atkinson, David  and
      Authur, Russell  and
      Chandu, Khyathi  and
      Cohan, Arman  and
      Dumas, Jennifer  and
      Elazar, Yanai  and
      Gu, Yuling  and
      Hessel, Jack  and
      Khot, Tushar  and
      Merrill, William  and
      Morrison, Jacob  and
      Muennighoff, Niklas  and
      Naik, Aakanksha  and
      Nam, Crystal  and
      Peters, Matthew  and
      Pyatkin, Valentina  and
      Ravichander, Abhilasha  and
      Schwenk, Dustin  and
      Shah, Saurabh  and
      Smith, William  and
      Strubell, Emma  and
      Subramani, Nishant  and
      Wortsman, Mitchell  and
      Dasigi, Pradeep  and
      Lambert, Nathan  and
      Richardson, Kyle  and
      Zettlemoyer, Luke  and
      Dodge, Jesse  and
      Lo, Kyle  and
      Soldaini, Luca  and
      Smith, Noah  and
      Hajishirzi, Hannaneh",
  booktitle={Proceedings of the 62nd annual meeting of the association for computational linguistics (volume 1: Long papers)},
  pages={15789--15809},
  year={2024}
}

@article{olmo20242,
  title={2 OLMo 2 Furious},
  author={OLMo, Team and Walsh, Pete and Soldaini, Luca and Groeneveld, Dirk and Lo, Kyle and Arora, Shane and Bhagia, Akshita and Gu, Yuling and Huang, Shengyi and Jordan, Matt and others},
  journal={arXiv preprint arXiv:2501.00656},
  year={2024}
}

@article{duchi2011adaptive,
  title={Adaptive subgradient methods for online learning and stochastic optimization.},
  author={Duchi, John and Hazan, Elad and Singer, Yoram},
  journal={Journal of machine learning research},
  volume={12},
  number={7},
  year={2011}
}

@inproceedings{gupta2018shampoo,
  title={Shampoo: Preconditioned stochastic tensor optimization},
  author={Gupta, Vineet and Koren, Tomer and Singer, Yoram},
  booktitle={International Conference on Machine Learning},
  pages={1842--1850},
  year={2018},
  organization={PMLR}
}

@article{pethick2025training,
  title={Training deep learning models with norm-constrained lmos},
  author={Pethick, Thomas and Xie, Wanyun and Antonakopoulos, Kimon and Zhu, Zhenyu and Silveti-Falls, Antonio and Cevher, Volkan},
  journal={arXiv preprint arXiv:2502.07529},
  year={2025}
}

@article{bernstein2024modular,
  title={Modular duality in deep learning},
  author={Bernstein, Jeremy and Newhouse, Laker},
  journal={arXiv preprint arXiv:2410.21265},
  year={2024}
}

@article{sharma2023truth,
  title={The truth is in there: Improving reasoning in language models with layer-selective rank reduction},
  author={Sharma, Pratyusha and Ash, Jordan T and Misra, Dipendra},
  journal={arXiv preprint arXiv:2312.13558},
  year={2023}
}

@article{chen2024distributional,
  title={Distributional associations vs in-context reasoning: A study of feed-forward and attention layers},
  author={Chen, Lei and Bruna, Joan and Bietti, Alberto},
  journal={arXiv preprint arXiv:2406.03068},
  year={2024}
}

@article{liu2025lift,
  title={LIFT the Veil for the Truth: Principal Weights Emerge after Rank Reduction for Reasoning-Focused Supervised Fine-Tuning},
  author={Liu, Zihang and Pang, Tianyu and Balabanov, Oleg and Yang, Chaoqun and Huang, Tianjin and Yin, Lu and Yang, Yaoqing and Liu, Shiwei},
  journal={arXiv preprint arXiv:2506.00772},
  year={2025}
}

@article{defilippis2025scaling,
  title={Scaling laws and spectra of shallow neural networks in the feature learning regime},
  author={Defilippis, Leonardo and Xu, Yizhou and Girardin, Julius and Troiani, Emanuele and Erba, Vittorio and Zdeborov{\'a}, Lenka and Loureiro, Bruno and Krzakala, Florent},
  journal={arXiv preprint arXiv:2509.24882},
  year={2025}
}

@article{martin2021implicit,
  title={Implicit self-regularization in deep neural networks: Evidence from random matrix theory and implications for learning},
  author={Martin, Charles H and Mahoney, Michael W},
  journal={Journal of Machine Learning Research},
  volume={22},
  number={165},
  pages={1--73},
  year={2021}
}

@article{martin2021predicting,
  title={Predicting trends in the quality of state-of-the-art neural networks without access to training or testing data},
  author={Martin, Charles H and Peng, Tongsu and Mahoney, Michael W},
  journal={Nature Communications},
  volume={12},
  number={1},
  pages={4122},
  year={2021},
  publisher={Nature Publishing Group UK London}
}

@article{hodgkinson2025models,
  title={Models of Heavy-Tailed Mechanistic Universality},
  author={Hodgkinson, Liam and Wang, Zhichao and Mahoney, Michael W},
  journal={arXiv preprint arXiv:2506.03470},
  year={2025}
}

@article{kothapalli2025spikes,
  title={From spikes to heavy tails: Unveiling the spectral evolution of neural networks},
  author={Kothapalli, Vignesh and Pang, Tianyu and Deng, Shenyang and Liu, Zongmin and Yang, Yaoqing},
  journal={Transactions on Machine Learning Research},
  year={2025}
}

@inproceedings{yang2023test,
  title={Test accuracy vs. generalization gap: Model selection in nlp without accessing training or testing data},
  author={Yang, Yaoqing and Theisen, Ryan and Hodgkinson, Liam and Gonzalez, Joseph E and Ramchandran, Kannan and Martin, Charles H and Mahoney, Michael W},
  booktitle={Proceedings of the 29th ACM SIGKDD Conference on Knowledge Discovery and Data Mining},
  pages={3011--3021},
  year={2023}
}

@article{dosovitskiy2020image,
  title={An image is worth 16x16 words: Transformers for image recognition at scale},
  author={Dosovitskiy, Alexey and Beyer, Lucas and Kolesnikov, Alexander and Weissenborn, Dirk and Zhai, Xiaohua and Unterthiner, Thomas and Dehghani, Mostafa and Minderer, Matthias and Heigold, Georg and Gelly, Sylvain and others},
  journal={arXiv preprint arXiv:2010.11929},
  year={2020}
}

@inproceedings{he2016deep,
  title={Deep residual learning for image recognition},
  author={He, Kaiming and Zhang, Xiangyu and Ren, Shaoqing and Sun, Jian},
  booktitle={Proceedings of the IEEE conference on computer vision and pattern recognition},
  pages={770--778},
  year={2016}
}

@article{touvron2023llama,
  title={Llama: Open and efficient foundation language models},
  author={Touvron, Hugo and Lavril, Thibaut and Izacard, Gautier and Martinet, Xavier and Lachaux, Marie-Anne and Lacroix, Timoth{\'e}e and Rozi{\`e}re, Baptiste and Goyal, Naman and Hambro, Eric and Azhar, Faisal and others},
  journal={arXiv preprint arXiv:2302.13971},
  year={2023}
}

@inproceedings{zhou2020pbsgd,
  title={pbSGD: Powered Stochastic Gradient Descent Methods for Accelerated Non-Convex Optimization.},
  author={Zhou, Beitong and Liu, Jun and Sun, Weigao and Chen, Ruijuan and Tomlin, Claire J and Yuan, Ye},
  booktitle={IJCAI},
  pages={3258--3266},
  year={2020}
}

@article{zhao2024galore,
  title={Galore: Memory-efficient llm training by gradient low-rank projection},
  author={Zhao, Jiawei and Zhang, Zhenyu and Chen, Beidi and Wang, Zhangyang and Anandkumar, Anima and Tian, Yuandong},
  journal={arXiv preprint arXiv:2403.03507},
  year={2024}
}

@article{lu2024alphapruning,
  title={Alphapruning: Using heavy-tailed self regularization theory for improved layer-wise pruning of large language models},
  author={Lu, Haiquan and Zhou, Yefan and Liu, Shiwei and Wang, Zhangyang and Mahoney, Michael W and Yang, Yaoqing},
  journal={Advances in neural information processing systems},
  volume={37},
  pages={9117--9152},
  year={2024}
}

@article{simsekli2020hausdorff,
  title={Hausdorff dimension, heavy tails, and generalization in neural networks},
  author={Simsekli, Umut and Sener, Ozan and Deligiannidis, George and Erdogdu, Murat A},
  journal={Advances in Neural Information Processing Systems},
  volume={33},
  pages={5138--5151},
  year={2020}
}

@inproceedings{hodgkinson2022generalization,
  title={Generalization bounds using lower tail exponents in stochastic optimizers},
  author={Hodgkinson, Liam and Simsekli, Umut and Khanna, Rajiv and Mahoney, Michael},
  booktitle={International Conference on Machine Learning},
  pages={8774--8795},
  year={2022},
  organization={PMLR}
}

@article{dandi2024random,
  title={A random matrix theory perspective on the spectrum of learned features and asymptotic generalization capabilities},
  author={Dandi, Yatin and Pesce, Luca and Cui, Hugo and Krzakala, Florent and Lu, Yue M and Loureiro, Bruno},
  journal={arXiv preprint arXiv:2410.18938},
  year={2024}
}

@article{he2025alphadecay,
  title={Alphadecay: Module-wise weight decay for heavy-tailed balancing in llms},
  author={He, Di and Tu, Songjun and Jaiswal, Ajay and Shen, Li and Yuan, Ganzhao and Liu, Shiwei and Yin, Lu},
  journal={arXiv preprint arXiv:2506.14562},
  year={2025}
}

@article{higham1990fast,
  title={Fast polar decomposition of an arbitrary matrix},
  author={Higham, Nicholas J and Schreiber, Robert S},
  journal={SIAM Journal on Scientific and Statistical Computing},
  volume={11},
  number={4},
  pages={648--655},
  year={1990},
  publisher={SIAM}
}

@article{wang2023spectral,
  title={Spectral evolution and invariance in linear-width neural networks},
  author={Wang, Zhichao and Engel, Andrew and Sarwate, Anand D and Dumitriu, Ioana and Chiang, Tony},
  journal={Advances in neural information processing systems},
  volume={36},
  pages={20695--20728},
  year={2023}
}

@article{simon2023eigenlearning,
  title={The eigenlearning framework: A conservation law perspective on kernel ridge regression and wide neural networks},
  author={Simon, James B and Dickens, Madeline and Karkada, Dhruva and DeWeese, Michael R},
  journal={Transactions on Machine Learning Research},
  year={2023}
}

@article{wang2025muon,
  title={Muon Outperforms Adam in Tail-End Associative Memory Learning},
  author={Wang, Shuche and Zhang, Fengzhuo and Li, Jiaxiang and Du, Cunxiao and Du, Chao and Pang, Tianyu and Yang, Zhuoran and Hong, Mingyi and Tan, Vincent YF},
  journal={arXiv preprint arXiv:2509.26030},
  year={2025}
}

@article{martin2025setol,
  title={SETOL: A Semi-Empirical Theory of (Deep) Learning},
  author={Martin, Charles H and Hinrichs, Christopher},
  journal={arXiv preprint arXiv:2507.17912},
  year={2025}
}

@inproceedings{hueigenspectrum,
  title={Eigenspectrum Analysis of Neural Networks without Aspect Ratio Bias},
  author={Hu, Yuanzhe and Goel, Kinshuk and Killiakov, Vlad and Yang, Yaoqing},
  booktitle={Forty-second International Conference on Machine Learning},
year={2025}
}

@inproceedings{sutskever2013importance,
  title={On the importance of initialization and momentum in deep learning},
  author={Sutskever, Ilya and Martens, James and Dahl, George and Hinton, Geoffrey},
  booktitle={International conference on machine learning},
  pages={1139--1147},
  year={2013},
  organization={pmlr}
}

@article{raffel2020exploring,
  title={Exploring the limits of transfer learning with a unified text-to-text transformer},
  author={Raffel, Colin and Shazeer, Noam and Roberts, Adam and Lee, Katherine and Narang, Sharan and Matena, Michael and Zhou, Yanqi and Li, Wei and Liu, Peter J},
  journal={Journal of machine learning research},
  volume={21},
  number={140},
  pages={1--67},
  year={2020}
}

@inproceedings{deng2009imagenet,
  title={Imagenet: A large-scale hierarchical image database},
  author={Deng, Jia and Dong, Wei and Socher, Richard and Li, Li-Jia and Li, Kai and Fei-Fei, Li},
  booktitle={2009 IEEE conference on computer vision and pattern recognition},
  pages={248--255},
  year={2009},
  organization={Ieee}
}

@article{krizhevsky2009learning,
  title={Learning multiple layers of features from tiny images},
  author={Krizhevsky, Alex and Hinton, Geoffrey},
  year={2009},
  publisher={Toronto, ON, Canada}
}

@article{garrigos2023handbook,
  title={Handbook of convergence theorems for (stochastic) gradient methods},
  author={Garrigos, Guillaume and Gower, Robert M},
  journal={arXiv preprint arXiv:2301.11235},
  year={2023}
}

@article{miyato2018spectral,
  title={Spectral normalization for generative adversarial networks},
  author={Miyato, Takeru and Kataoka, Toshiki and Koyama, Masanori and Yoshida, Yuichi},
  journal={arXiv preprint arXiv:1802.05957},
  year={2018}
}

@article{davis2025spectral,
  title={When do spectral gradient updates help in deep learning?},
  author={Davis, Damek and Drusvyatskiy, Dmitriy},
  journal={arXiv preprint arXiv:2512.04299},
  year={2025}
}

@article{chang2025convergence,
  title={On the Convergence of Muon and Beyond},
  author={Chang, Da and Liu, Yongxiang and Yuan, Ganzhao},
  journal={arXiv preprint arXiv:2509.15816},
  year={2025}
}

@misc{ma2026preconditioningbenefitsspectralorthogonalization,
      title={Preconditioning Benefits of Spectral Orthogonalization in Muon}, 
      author={Jianhao Ma and Yu Huang and Yuejie Chi and Yuxin Chen},
      year={2026},
      eprint={2601.13474},
      archivePrefix={arXiv},
      primaryClass={cs.LG},
      url={https://arxiv.org/abs/2601.13474}, 
}

@misc{liu2024sophiascalablestochasticsecondorder,
      title={Sophia: A Scalable Stochastic Second-order Optimizer for Language Model Pre-training}, 
      author={Hong Liu and Zhiyuan Li and David Hall and Percy Liang and Tengyu Ma},
      year={2024},
      eprint={2305.14342},
      archivePrefix={arXiv},
      primaryClass={cs.LG},
      url={https://arxiv.org/abs/2305.14342}, 
}

@misc{Gokaslan2019OpenWeb,  
	title={OpenWebText Corpus},
	author={Aaron Gokaslan and Vanya Cohen},
	howpublished={\url{http://Skylion007.github.io/OpenWebTextCorpus}}, 
	year={2019}
}

@article{radford2019language,
  title={Language models are unsupervised multitask learners},
  author={Radford, Alec and Wu, Jeffrey and Child, Rewon and Luan, David and Amodei, Dario and Sutskever, Ilya and others},
  journal={OpenAI blog},
  volume={1},
  number={8},
  pages={9},
  year={2019}
}

@article{deng2026suspicious,
  title={Suspicious Alignment of SGD: A Fine-Grained Step Size Condition Analysis},
  author={Deng, Shenyang and Liao, Boyao and Ouyang, Zhuoli and Pang, Tianyu and Song, Minhak and Yang, Yaoqing},
  journal={arXiv preprint arXiv:2601.11789},
  year={2026}
}

@inproceedings{zhang2025pretrained,
  title={Pretrained image-text models are secretly video captioners},
  author={Zhang, Chunhui and Jian, Yiren and Ouyang, Zhongyu and Vosoughi, Soroush},
  booktitle={Proceedings of the 2025 Conference of the Nations of the Americas Chapter of the Association for Computational Linguistics: Human Language Technologies},
  year={2025}
}

@inproceedings{zhang2024working,
  title={Working memory identifies reasoning limits in language models},
  author={Zhang, Chunhui and Jian, Yiren and Ouyang, Zhongyu and Vosoughi, Soroush},
  booktitle={Proceedings of the 2024 Conference on Empirical Methods in Natural Language Processing},
  pages={16896--16922},
  year={2024}
}

@article{qi2026delving,
  title={Delving into Muon and Beyond: Deep Analysis and Extensions},
  author={Qi, Xianbiao and Chen, Marco and Ye, Jiaquan and He, Yelin and Xiao, Rong},
  journal={arXiv preprint arXiv:2602.04669},
  year={2026}
}

\newpage
\appendix

\section{Proofs}

\paragraph{Notation.} For a vector $\vv$, we denote its $l_2$ norm as $\|\vv\|_2$. For a matrix $\mA,\mB\in \R^{m\times n}$, we denote its  spectral norm as $\|\mA\|$, Frobenius norm as $\|\mA\|_\F$,  nuclear norm as $\|\mA\|_*$, where $\|\mA\|_*=\sum_{i=1}^r \sigma_i$, here $\{\sigma_i\}_{i=1}^r$ are  singular values, for $p>1$, Schatten$-p$ norm as $\|\mA\|_p$, where $\|\mA\|_p=\left(\sum_{i=1}^r \sigma_i^p\right)^{\frac{1}{p}}$.  We denote $\langle \mA, \mB\rangle = \operatorname{Tr }\left(\mA^{\top}\mB\right)$.

\subsection{Update Under the Schatten$-q$ Norm Constraint} \label{proof:Schatten$-q$ Norm Constraint}
\begin{lemma} [Von Neumann trace inequality]
For  matrix $\mA \in \mathbb{R}^{m \times n}$, let $\sigma_i(\mA)$ denote the $i$-th largest singular value of $\mA$, and let $r=\min(m,n)$. Then for any $\mA,\mB \in \mathbb{R}^{m \times n}$, we have
\[
\langle \mA, \mB\rangle = \operatorname{Tr }\left(\mA^{\top}\mB\right) \leq \sum_{i=1}^{r} \sigma_i(\mA)\,\sigma_i(\mB).
\]
\end{lemma}

\begin{theorem}
Let $\widetilde{\mG_t} =\mU\mSigma\mV^\top$ be the \texttt{SVD} of the negative gradient $\widetilde{\mG_t}  = -\nabla f(\mW_t)$, where $\mSigma = \operatorname{diag}(\sigma_1, \dots, \sigma_r)$. Let $p$ be the conjugate exponent of $q$, satisfying $1/p^{\prime} + 1/q = 1$. The explicit solution $\Delta \mW^*$ to the optimization problem~\ref{eq:equivalent_problem} is given by:
\begin{equation}
    \Delta \mW^* = \delta \frac{1}{\|\widetilde{\mG_t} \|_{p^{\prime}}^{p^{\prime}/q}} \mU \mSigma^{p^{\prime}-1} \mV^\top,
\end{equation}
where $\mSigma^{p^{\prime}-1} = \operatorname{diag}(\sigma_1^{p^{\prime}-1}, \dots, \sigma_r^{p^{\prime}-1})$.
\end{theorem}

\paragraph{Proof.}
First, by invoking Von Neumann's trace inequality, we have the upper bound 
$$\operatorname{Tr}(\widetilde{\mG_t} ^\top \Delta \mW) \leq \sum_i \sigma_i(\widetilde{\mG_t} ) \sigma_i(\Delta \mW)$$
The equality holds if and only if $\Delta \mW$ shares the same left and right singular vectors as $\widetilde{\mG_t} $. Thus, the optimal $\Delta \mW$ must take the form:
\begin{equation}
    \Delta \mW = \mU \mLambda \mV^\top, \quad \text{where } \mLambda = \operatorname{diag}(\lambda_1, \dots, \lambda_r).
\end{equation}
Here, $\lambda_i \ge 0$ represents the singular values of $\Delta \mW$. The problem reduces to maximizing $\sum_i \sigma_i \lambda_i$ subject to $\|\lambda\|_q \leq \delta$. By Hölder's inequality, we have the upper bound:
\begin{equation}
    \sum_{i} \sigma_i \lambda_i \leq \|\sigma\|_{p^{\prime}} \|\lambda\|_q \leq \|\sigma\|_{p^{\prime}} \delta.
\end{equation}
Equality is achieved if and only if $\lambda_i^q \propto \sigma_i^{p^{\prime}}$, which implies:
\begin{equation}
    \lambda_i = c \sigma_i^{{p^{\prime}}/q},
\end{equation}
for some constant $c \ge 0$. Using the conjugate property $1/{p^{\prime}} + 1/q = 1$, the exponent simplifies to:
\begin{equation}
    \frac{{p^{\prime}}}{q} = {p^{\prime}} \left( 1 - \frac{1}{{p^{\prime}}} \right) = {p^{\prime}} - 1.
\end{equation}
Substituting this back yields the optimal singular values:
\begin{equation}
     \lambda_i = c \cdot \sigma_i^{{p^{\prime}}-1},
\end{equation}
where $c$ is a normalization constant. To determine $c$, we substitute these values into the active constraint $\|\Delta \mW\|_q = \delta$:
\begin{equation}
    \left( \sum_{i} (c \sigma_i^{{p^{\prime}}-1})^q \right)^{1/q} = c \left( \sum_{i} \sigma_i^{({p^{\prime}}-1)q} \right)^{1/q} = \delta.
\end{equation}
Using the algebraic relation $({p^{\prime}}-1)q = {p^{\prime}}$, the term involving singular values simplifies to the Schatten$-{p^{\prime}}$ norm of $\widetilde{\mG_t} $. Solving for $c$, we obtain:
\begin{equation}
    c \|\widetilde{\mG_t} \|_{p^{\prime}}^{{p^{\prime}}/q} = \delta \implies c = \frac{\delta}{\|\widetilde{\mG_t} \|_{p^{\prime}}^{{p^{\prime}}/q}}.
\end{equation}
Substituting $c$ back into the expression $\Delta \mW = U \Lambda V^\top$ yields the stated closed-form solution. \hfill $\square$

\subsection{Convergence Analysis}
\label{proof: Convergence Analysis}

We consider the following stochastic optimization problem: 
$$\min_{\mW\in \R^{m\times n}} f(\mW)=\E_{\vxi}[f(\mW;\vxi)],$$
where $\vxi$ is stochastic noise. We denote $f^*=\inf_\mW f(\mW)$, $r=\min\{m,n\}$. We assume $f^*>-\infty$. 

We give the convergence analysis under a smooth non-convex setting for finding $\epsilon$-stationary points, i.e. $ \|\nabla f(x)\|_{*} \leq \epsilon$.

To better understand the proof, we give a stochastic version of \texttt{HTMuon} in Algorithm~\ref{alg:HTMUON_Stochastic} for  convergence analysis.

\begin{algorithm}[ht]
\caption{\texttt{HTMuon\_Stochastic}}
\begin{algorithmic}[1]
\State \textbf{Input:} Initial weights $\mathbf{W}_0 \in \mathbb{R}^{m\times n}$, loss function $L$,  learning rate $\eta_t$, momentum parameter $\beta$, batch size B, power $p\in (0,1)$.\
\State Initialize $\mathbf{M}_0 \in\mathbb{R}^{m\times n} \leftarrow \mathbf{0}$
\For{$t=1,2,\ldots$}
\State Sample batch $\left\{\xi_{t, i}\right\}_{i=1}^B$ uniformly
\State $\mathbf{G}_t=\frac{1}{B} \sum_{i=1}^B  \nabla_{\mathbf{W}}L\left(W_t ; \xi_{t, i}\right)$ 
\State $\mathbf{M}_t \leftarrow \beta\mathbf{M}_{t-1} + (1-\beta)\mathbf{G}_t$
\State $\mU_t,\mSigma_t, \mV_t^{\top} \leftarrow \texttt{SVD}(\mathbf{M}_t)$
\State \textcolor{blue}{$\mO_t \leftarrow \mU_t\mSigma_t^{p} \mV_t^{\top} $}
\State $\mathbf{W}_{t+1} \leftarrow \mathbf{W}_t - \eta_t \mO_t$
\EndFor
\end{algorithmic}
\label{alg:HTMUON_Stochastic}
\end{algorithm}

\begin{lemma} For $t = 0, 1, \cdots, T, $ $\mM_t$ and $\mW_t$ are generated by Algorithm~\ref{alg:HTMUON_Stochastic}. Consider that $\mM_0=\mG_0, \mM_t=\beta\mM_{t-1}+(1-\beta)\mG_{t}, $ under Assumptions~\ref{ass: smooth} and~\ref{ass: variance}, if we assume $\eta=\max \{ \eta_t \}_{t=1}^T$, under Assumption~\ref{ass: bounded_sv}, we have
\begin{align*}
 \E \|\nabla f(\mW_t)-&\mM_t\|_\F \leq \sqrt{\frac{1-\beta}{1+\beta}}\frac{\sigma}{\sqrt{B}} +\frac{\beta^t\sigma}{\sqrt{B}}+\frac{\sqrt{r}l^{p}\beta L\eta}{1-\beta},
\end{align*}
where $\mG_t=\frac{1}{B}\sum_{i=1}^B\nabla f(\mW_t; \xi_{t,i})$ and B is batch size.

\label{lemma: bound_for_momentum}
\end{lemma}
\paragraph{Proof.} We define $\mC_0=\nabla f(\mW_0), \mC_t=\beta\mC_{t-1}+(1-\beta)\nabla f(\mW_t)=(1-\beta)\sum_{i=1}^t\beta^{t-i}\nabla f(\mW_i)+\beta^t\nabla f(\mW_0)$, under Assumptions~\ref{ass: smooth} and~\ref{ass: variance} we note that
\begin{align*}
    & \E[\|\nabla f(\mW_t-\mC_t)\|_{\F}]\\
    =&\E[\|\nabla f(\mW_t)-(\beta\mC_{t-1}+(1-\beta)\nabla f(\mW_t))\|_{\F}]\\
    =& \E[\beta\|\nabla f(\mW_t-\mC_{t-1})\|_{\F}]\\
    \leq & \E \bigr[\beta\|\nabla f(\mW_{t-1}-\mC_{t-1})\|_{\F} \bigr] \\
    + &  \E \bigl[\beta \| \nabla f(\mW_{t-1}))-\nabla f(\mW_t)) \|_{\F} \bigr] \\
 \leq & \E \bigr[\beta\|\nabla f(\mW_{t-1}-\mC_{t-1})\|_{\F} \bigr]\\
    + &  \E \bigl[\beta L \| \mW_{t-1}-\mW_t \|_{\F} \bigr] \\
    \leq & \E \bigr[\beta\|\nabla f(\mW_{t-1}-\mC_{t-1})\|_{\F} \bigr]\\
    +&  \E \bigl[\beta L \eta  \| \mU_{t-1}\mSigma^p_{t-1}\mV_{t-1}^{\top}\|_{\F} \bigr] \\ 
    \leq & \E \bigr[\beta\|\nabla f(\mW_{t-1}-\mC_{t-1})\|_{\F} + \beta L \eta \sqrt{r}l^p \bigr] \\
    \leq & \beta^t \| \nabla f(\mW_0) -\mC_0 \|_{\F} + \sum_{i=1}^t \beta^i L \eta \sqrt{r}l^p  \\
    \leq & \frac{\sqrt{r}l^p\beta L \eta}{1-\beta}.
\end{align*}

And follow the same proof in  Section B.1 in \citet{shen2025convergence}, we have
\begin{equation*}
    \E [\|\mC_t -\mM_t \|_{\F}] \leq \sqrt{\frac{1-\beta}{1+\beta}}\frac{\sigma}{\sqrt{B}}+\beta^t\frac{\sigma}{\sqrt{B}}.
\end{equation*}
So we have 
\begin{align*}
    &\E \|\nabla f(\mW_t)-\mM_t\|_\F \\ 
    \leq &   \E[\|\nabla f(\mW_t-\mC_t)\|_{\F}] +  \E [\|\mC_t -\mM_t \|_{\F}] \\
    \leq & \sqrt{\frac{1-\beta}{1+\beta}}\frac{\sigma}{\sqrt{B}} +\frac{\beta^t\sigma}{\sqrt{B}}+\frac{\sqrt{r}l^{p}\beta L\eta}{1-\beta}
\end{align*}\hfill $\square$

\begin{theorem}[Muon \citep{shen2025convergence}]\label{thm: muon_nonconvex}
    Under Assumptions \ref{ass: smooth} and \ref{ass: variance}, if we apply Muon with $\eta_t=\eta$,  then
\begin{align*}
\frac{1}{T}&\sum_{t=0}^{T-1}\E\bigl[\|\nabla f(\mW_t)\|_*\bigr]
\le \frac{\E\bigl[f(\mW_0)-f(\mW_T)\bigr]}{T\eta} + \frac{L r\eta}{2}+ \frac{2\sigma\sqrt{r(1-\beta)}}{\sqrt{B(1+\beta)}} + \frac{2\beta\sigma\sqrt{r}}{(1-\beta)T\sqrt{B}} + \frac{2r\eta \beta L}{1-\beta}.
\end{align*}
Where B is batch size.
Denote $\Delta = f(\mW_0)-f^*$. If we set $B=1$, 
$\eta = \sqrt{\frac{(1-\beta)\Delta}{rTL}}$, 
$1-\beta = \min\left\{\frac{\sqrt{L\Delta}}{\sigma\sqrt{T}},1\right\}$, then

\begin{align*}
&\frac{1}{T}\sum_{t=0}^{T-1}\E\bigl[\|\nabla f(\mW_t)\|_*\bigr]
\le  O\Biggl(
    \sqrt[4]{\frac{r^2L\Delta\sigma^2}{T}}
    + \sqrt{\frac{rL\Delta}{T}} 
    + \frac{\sqrt{r}\sigma^2}{\sqrt{L\Delta T}}
\Biggr).
\end{align*}
Thus, \texttt{Muon} can find an $\epsilon$-nuclear norm stationary point of $f$ with a complexity of $O(r^2L\sigma^2\Delta\epsilon^{-4})$. 
\end{theorem}

\begin{theorem}[HTMuon: Theorem~\ref{thm: htmuon_nonconvex}]
Under Assumptions ~\ref{ass: smooth}, ~\ref{ass: variance} and~\ref{ass: bounded_sv}, if we apply \texttt{HTMuon} in Algorithm~\ref{alg:HTMUON_Stochastic} with adaptive learning rate $\eta_t=\frac{\langle  \mM_t,\rho(\mM_t)\rangle  }{L\|\rho(\mM_t)\|^2_\F}$,  $\rho(\mM_t)=\mU_t\mSigma_t^{p}\mV_t^{\top}$ and batch size $B=T$, then 
\begin{align*}
    &\frac{1}{T}\sum_{t=1}^T\E \bigl[ \|\nabla f(\mW_t)\|^2_{1+p} \bigr] \leq \frac{2Lr^{\frac{1}{s}}}{1-\varepsilon^{\prime}}\left( \frac{1-\varepsilon^{\prime}}{L }+\frac{1}{L\varepsilon^{\prime}}\right)\bigl[\frac{2(1-\beta)\sigma}{B(1+\beta)} +\frac{2\beta^{2}\sigma^2}{(1-\beta^2)BT}+\frac{2rl^{2p}\beta^2L^2\eta^2}{(1-\beta)^2} \bigr]+ \frac{4Lr^{\frac{1}{s}}}{1-\varepsilon^{\prime}}\frac{\bigl[f(\mW_0)-f(\mW_{T})\bigr] }{T}.
\end{align*}
Where $s=\frac{1+p}{1-p}$ and $0<\varepsilon^{\prime}<1$. Furthermore, if we  define $\beta=\frac{1}{\sqrt{T}}$, $\Delta=f(\mW_0)-f^*$, thus we have 
\begin{align*}
       &\frac{1}{T}\sum_{t=1}^T\E \bigl[ \|\nabla f(\mW_t)\|^2_{1+p} \bigr] \leq \frac{2Lr^{\frac{1}{s}}}{1-\varepsilon^{\prime}}\left( \frac{1-\varepsilon^{\prime}}{L }+\frac{1}{L\varepsilon^{\prime}}\right)\bigl[\frac{2\sigma}{T}+\frac{4\sigma^2}{T} +\frac{4rl^{2p}L^2\eta^2}{T} \bigr]+ \frac{4Lr^{\frac{1}{s}}}{1-\varepsilon^{\prime}}\frac{\Delta }{T}.
\end{align*}
\end{theorem}

\paragraph{Proof.} Consider $\mM_{t}=\mU_t\mSigma_t\mV_t^{\top}$,
we first define $\rho(\mM_t)=\mU_t\mSigma_t^{p}\mV_t^{\top}$.
By Assumption ~\ref{ass: smooth}, we have
\begin{align*}
&\E \bigl[f(\mW_t)-f(\mW_{t+1})\bigr] \\
& \geq \E \bigl[ \eta_t \langle \nabla f(\mW_t),\rho(\mM_t)\rangle-\frac{L\eta_t^2}{2}\|\rho(\mM_t)\|^2_{\F}\bigr] \\
& \geq \E \bigl[ \eta_t \langle  \mM_t,\rho(\mM_t)\rangle-\frac{L\eta_t^2}{2}\|\rho(\mM_t)\|^2_{\F} -\eta_t \langle  \mM_t-\nabla f(\mW_t),\rho(\mM_t)\rangle\bigr]. 
\end{align*}
Here we consider adaptive learning rate $\eta_t=\frac{\langle  \mM_t,\rho(\mM_t)\rangle  }{L\|\rho(\mM_t)\|^2_\F}>0$, thus
\begin{align*}
&\E \bigl[f(\mW_t)-f(\mW_{t+1})\bigr] \\
& \geq \E \bigl[   \left(\frac{\langle  \mM_t,\rho(\mM_t)\rangle^2}{L\|\rho(\mM_t)\|^2_\F}-\frac{\langle  \mM_t,\rho(\mM_t)\rangle^2}{2L\|\rho(\mM_t)\|^2_\F}\right) -\eta_t \|\mM_t-\nabla f(\mW_t)\|_\F\|\rho(\mM_t)\|_F \bigr] \\ 
& \geq \E \bigl[   \left(\frac{\langle  \mM_t,\rho(\mM_t)\rangle^2}{L\|\rho(\mM_t)\|^2_\F}-\frac{\langle  \mM_t,\rho(\mM_t)\rangle^2}{2L\|\rho(\mM_t)\|^2_\F}\right) -\frac{ \langle  \mM_t,\rho(\mM_t)\rangle}{L\|\rho(\mM_t)\|_F}\|\mM_t-\nabla f(\mW_t)\|_\F\bigr]
\end{align*}
By $2ab\leq \varepsilon^{\prime} a^2+ \frac{b^2}{\varepsilon^{\prime} }$,  where $0< \varepsilon^{\prime} <1$, we have 
\begin{align*}
&\E \bigl[f(\mW_t)-f(\mW_{t+1})\bigr] \\
& \geq \E  \bigl[ \frac{\langle  \mM_t,\rho(\mM_t)\rangle^2}{L\|\rho(\mM_t)\|^2_\F}-\frac{\langle  \mM_t,\rho(\mM_t)\rangle^2}{2L\|\rho(\mM_t)\|^2_\F} -\frac{\varepsilon^{\prime}}{2L}\frac{\langle  \mM_t,\rho(\mM_t)\rangle^2}{\|\rho(\mM_t)\|^2_\F}-\frac{1}{2L\varepsilon^{\prime}}\|\mM_t-\nabla f(\mW_t)\|^2_\F \bigr] \\ 
& \geq \E  \bigl[ \frac{1-\varepsilon^{\prime}}{2L}\frac{\langle  \mM_t,\rho(\mM_t)\rangle^2}{\|\rho(\mM_t)\|^2_\F} -\frac{\|\mM_t-\nabla f(\mW_t)\|^2_\F}{2L\varepsilon^{\prime}}\bigr]
\end{align*}
By Holder's Inequality, for $p\in (0,1)$ and we set $s=\frac{1+p}{1-p}$ and $m=\frac{1+p}{2p}$, we have 
\begin{align*}
\|\rho(\mM_t)\|^2_\F=&\sum_{i=1}^r \sigma_i^{2p}\leq  r^{\frac{1}{s}}\left(\sum_{i=1}^r(\sigma_i^{2p})^m\right)^{\frac{1}{m}} \leq r^{\frac{1}{s}}\left(\sum_{i=1}^r\sigma_i^{1+p}\right)^{\frac{2p}{p+1}}.
\end{align*}
Thus we have 
\begin{align*}
    \frac{\langle  \mM_t,\rho(\mM_t)\rangle^2}{\|\rho(\mM_t)\|^2_\F}&\geq \frac{\left(\sum_{i=1}^r \sigma_i^{1+p}\right)^2}{r^{\frac{1}{s}}\left(\sum_{i=1}^r \sigma_i^{1+p}\right)^{\frac{2p}{p+1}}}=\frac{\|\mM_t\|^2_{1+p}}{r^{\frac{1}{s}}}.
\end{align*} 

Hence, we have 
\begin{align*}
&\E \bigl[f(\mW_t)-f(\mW_{t+1})\bigr] \geq \E  \bigl[ \frac{1-\varepsilon^{\prime}}{2Lr^{\frac{1}{s}}}\|\mM_t\|^2_{1+p} -\frac{\|\mM_t-\nabla f(\mW_t)\|^2_\F}{2L\varepsilon^{\prime}}\bigr]
\end{align*}

By Lemma ~\ref{lemma: bound_for_momentum}, we let $\eta=\max\{\eta_t\}_{t=1}^T$, we have  
\begin{align*}
 \E \|\nabla &f(\mW_t)-\mM_t\|^2_\F  \leq  \left( \sqrt{\frac{1-\beta}{1+\beta}}\frac{\sigma}{\sqrt{B}}+\frac{\beta^t\sigma}{\sqrt{B}}+\frac{\sqrt{r}l^{p}\beta L\eta}{1-\beta} \right)^2 \leq \frac{2(1-\beta)\sigma}{B(1+\beta)}+\frac{2\beta^{2t}\sigma^2}{B}+\frac{2rl^{2p}\beta^2L^2\eta^2}{(1-\beta)^2}
\end{align*}
Thus we have 
\begin{align*}
    &\E \bigl[  \frac{1-\varepsilon^{\prime}}{2Lr^{\frac{1}{s}}}\|\nabla f(\mW_t)\|^2_{1+p} \bigr]\\
    & \leq \E \bigl[  \frac{1-\varepsilon^{\prime}}{Lr^{\frac{1}{s}}}\|\nabla f(\mW_t)-\mM_t\|^2_{1+p} \bigr]+\E  \bigl[\frac{1-\varepsilon^{\prime}}{Lr^{\frac{1}{s}}}\|\mM_t\|^2_{1+p}\bigr]\\
    & \leq  \E \bigl[ \frac{1-\varepsilon^{\prime}}{L}\|\nabla f(\mW_t)-\mM_t\|^2_\F \bigr] +\E \bigl[ \frac{1-\varepsilon^{\prime}}{Lr^{\frac{1}{s}}}\|\mM_t\|^2_{1+p} \bigr] \\
    &  \leq \E\bigl[\left( \frac{1-\varepsilon^{\prime}}{L}+\frac{1}{L\varepsilon^{\prime}}\right)\|\nabla f(\mW_t)-\mM_t\|^2_\F  + f(\mW_t)-f(\mW_{t+1})\bigr]
\end{align*}

Thus we have  
\begin{align*}
    &\frac{1}{T}\sum_{t=1}^T\E \bigl[ \|\nabla f(\mW_t)\|^2_{1+p} \bigr] \\
    & \leq \frac{1}{T}\sum_{t=1}^T \E\bigl[\frac{2Lr^{\frac{1}{s}}}{1-\varepsilon^{\prime}}\left( \frac{1-\varepsilon^{\prime}}{L }+\frac{1}{L\varepsilon^{\prime}}\right) \|\nabla f(\mW_t)-\mM_t\|^2_\F \bigr]+ \frac{2Lr^{\frac{1}{s}}}{1-\varepsilon^{\prime}}\frac{\bigl[f(\mW_0)-f(\mW_{T})\bigr]}{T}\\
    & \leq \frac{2Lr^{\frac{1}{s}}}{1-\varepsilon^{\prime}}\left( \frac{1-\varepsilon^{\prime}}{L }+\frac{1}{L\varepsilon^{\prime}}\right)\bigl[\frac{2(1-\beta)\sigma}{B(1+\beta)} +\frac{2\beta^{2}\sigma^2}{(1-\beta^2)BT}+\frac{2rl^{2p}\beta^2L^2\eta^2}{(1-\beta)^2} \bigr]+ \frac{2Lr^{\frac{1}{s}}}{1-\varepsilon^{\prime}}\frac{\bigl[f(\mW_0)-f(\mW_{T})\bigr] }{T}
\end{align*}
We define $B=T,\beta=\frac{1}{\sqrt{T}}$, $\Delta=f(\mW_0)-f^*$, thus we have 
\begin{align*}
       &\frac{1}{T}\sum_{t=1}^T\E \bigl[ \|\nabla f(\mW_t)\|^2_{1+p} \bigr] \leq \frac{2Lr^{\frac{1}{s}}}{1-\varepsilon^{\prime}}\left( \frac{1-\varepsilon^{\prime}}{L }+\frac{1}{L\varepsilon^{\prime}}\right)\bigl[\frac{2\sigma}{T}+\frac{2\sigma^2}{(1-\beta^2)T^3} +\frac{2rl^{2p}L^2\eta^2}{T(1-\beta)^2} \bigr]+ \frac{2Lr^{\frac{1}{s}}}{1-\varepsilon^{\prime}}\frac{\Delta }{T}.
\end{align*}

When $\beta=\frac{1}{\sqrt{T}}$, it is easy to get $1-\beta\geq\frac{1}{2}$, thus we have

\begin{align*}
       &\frac{1}{T}\sum_{t=1}^T\E \bigl[ \|\nabla f(\mW_t)\|^2_{1+p} \bigr] \leq \frac{2Lr^{\frac{1}{s}}}{1-\varepsilon^{\prime}}\left( \frac{1-\varepsilon^{\prime}}{L }+\frac{1}{L\varepsilon^{\prime}}\right)\bigl[\frac{2\sigma}{T}+\frac{4\sigma^2}{T^3} +\frac{4rl^{2p}L^2\eta^2}{T} \bigr]+ \frac{2Lr^{\frac{1}{s}}}{1-\varepsilon^{\prime}}\frac{\Delta }{T}.
\end{align*}
Based on  Theorem~\ref{thm: htmuon_nonconvex} and Definition~\ref{Def:nuclear norm stationary point}, when $B=T$, the sample complexity upper bound is  $O(\epsilon^{-4})$, which matches \texttt{Muon}'s sample complexity in Theorem~\ref{thm: muon_nonconvex} and \citep{chang2025convergence} and \texttt{SGDM}'s sample complexity  in \citet{garrigos2023handbook,arjevani2023lower}. .\hfill $\square$

\subsection{Lemma for PL exponent $\alpha$ }\label{app:proof_PL_alpha}
\begin{lemma}
    Suppose  the singular values of  matrix $\mW \in \mathbb{R}^{n\times m}$ follow $s_k = s_1k^{-s}, 1\leq k\leq n$, we have PL exponent $\alpha$ of  $\mW$   satisfies $\alpha=1+\frac{1}{2s}.$
\end{lemma}

\paragraph{Proof.} Since we have $s_k = s_1k^{-s}, 1\leq k\leq n$, which means  eigenvalues $\lambda_k = \lambda_1k^{-2s}, 1\leq k\leq n$. Here we suppose $\Lambda$ is a random variable distributed according to
the empirical distribution from these eigenvalues, we can see that the distribution function takes the following form:
\begin{equation*}
    \mathbb{P}(\Lambda > \lambda_1 k^{-2s}) = \frac{k}{n}.
\end{equation*}
By changing variables $\lambda_1 k^{-s}=\lambda$, we get the cumulative distribution function of $\Lambda$:
\begin{equation*}
    \mathbb{P}(\Lambda > \lambda) \sim \lambda^{-\,\frac{1}{2s}}.
\end{equation*}

After that, we take the derivative with respect to $\lambda$, and we get the ESD:
\begin{equation*}
    p(\lambda) \sim \lambda^{-\left(\frac{1}{2s}+1\right)}.
\end{equation*}
So we have $\alpha=1+\frac{1}{2s}.$ \hfill $\square$

\section{More Experimental Details}\label{app:more_results}

In this section, we present more experimental details.

\subsection{\texttt{Muon}  and \texttt{HTMuon} Algorithms }
In Algorithm~\ref{algo:MUON} and~\ref{algo:MUON_SVD}, we give the implementations of \texttt{Muon\_NS} and \texttt{Muon\_SVD}.  \texttt{Muon\_NS} corresponds to the commonly used five-step \texttt{Newton Schulz} implementation of \texttt{Muon}, whereas \texttt{Muon\_SVD} represents the theoretically exact SVD-based implementation. Empirically, we find that due to the numerical approximation in \texttt{Muon\_NS}, the singular values of the update matrix do not always equal 1 exactly. Such deviations may suppress weights along noise-dominated eigenvector directions, which could explain why \texttt{Muon\_NS}  performs better than \texttt{Muon\_SVD} in practice.

In Algorithm~\ref{algo:HTMUON_NS}, we give the implementations of \texttt{HTMuon\_NS}. As discussed in Section~\ref{sec:More_Efficient_Implementations}, we note that the \texttt{HTMuon} update $\mO_t=\mU_t \mSigma_t^{p}\mV_t^{\top}$ admits the factorization
\[
\mO_t = (\mU_t\mV_t^{\top})\,(\mV_t\mSigma_t^{p}\mV_t^{\top}).
\]
For $\mU_t\mV_t^{\top}$ part, this is \texttt{Muon}'s update, we can consider  \texttt{Newton-Schulz} with 5 steps. To compute the symmetric factor $\mV_t\mSigma_t^{p}\mV_t^{\top}$ efficiently, we apply the \texttt{NS\_root} routine to $\mM_t^{\top}\mM_t$ in Algorithm~\ref{algo:NS_2p_root}, since $$\mM_t^{\top}\mM_t=\mV_t\mSigma_t^{2}\mV_t^{\top},$$
we have $\mV_t\mSigma_t^{p}\mV_t^{\top}=(\mM_t^{\top}\mM_t)^{\frac{p}{2}}$. Accordingly, \texttt{NS\_root} uses \texttt{NewtonSchulz} iterations to approximate matrix square roots and applies successive square-root operations for $[1-\log_2 p]$ rounds.

\subsection{Training Loss Curve}
\begin{figure*}[htbp]
     \begin{subfigure}[b]{0.45\textwidth}
         \centering
         \includegraphics[width=\textwidth, trim=0 8mm 0 0 crop]{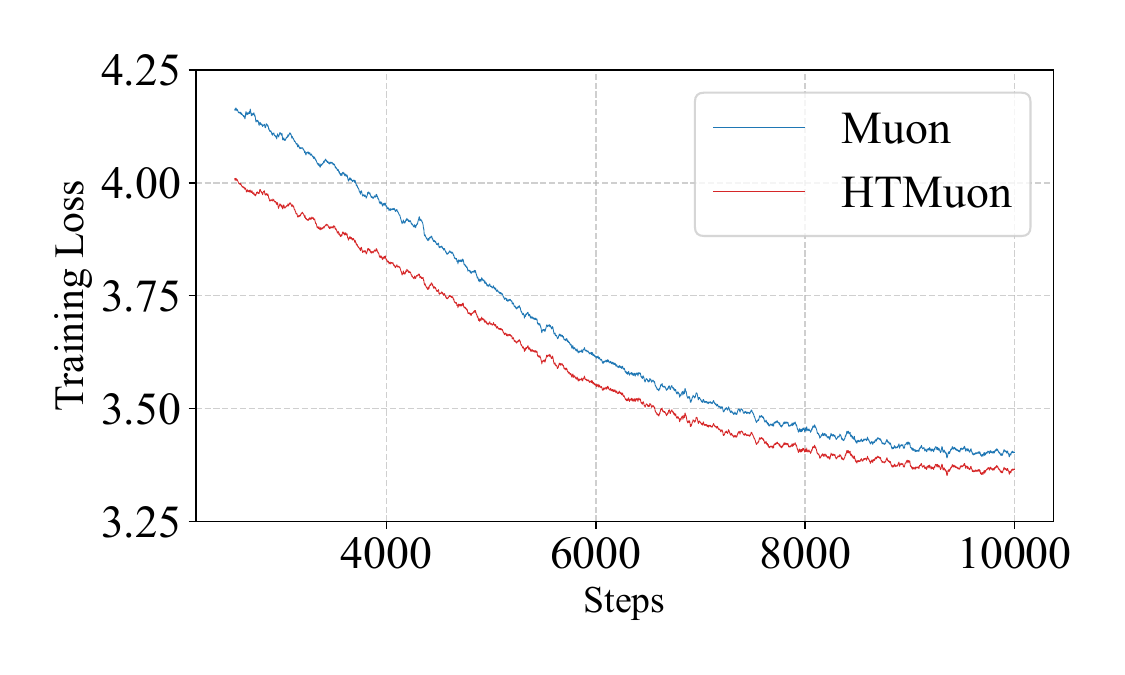}
         \captionsetup{justification=centering}
         \caption{ LLaMa-60M on C4 }
         \label{fig:loss_60M}
     \end{subfigure}
     \begin{subfigure}[b]{0.45\textwidth}
         \centering
         \includegraphics[width=\textwidth, trim=0 8mm 0 0 crop]{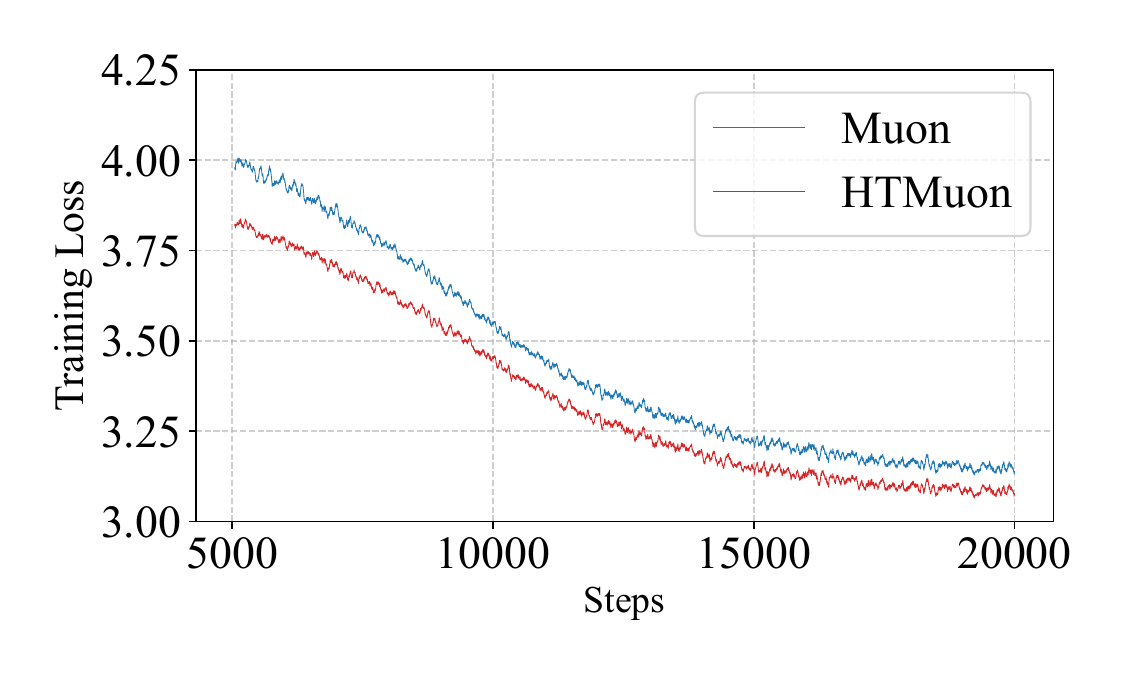}
         \captionsetup{justification=centering}
         \caption{ LLaMA-135M on C4 }
         \label{fig:loss_130M}
     \end{subfigure}
     \caption{Training loss curves for LLaMA-60M and LLaMA-135M. Learning rate for both models and optimizers is 0.03. Both curves are smoothed via a simple moving average (uniform weights, window = 50).}
     \label{fig:loss}
\end{figure*}

In Figure~\ref{fig:loss}, we show the smoothed training loss curves for LLaMA-60M and LLaMA-135M on C4 dataset. We find in the later stages of training, \texttt{HTMuon} consistently achieves a noticeably lower training loss than \texttt{Muon}, demonstrating the effectiveness and training stability.

\subsection{More Experiments Results}
In Table~\ref{table: muon_variants}, we conduct evaluations on LLaMA-60M and LLaMA-135M on the C4 datatset, comparing \texttt{HTMuon} with multiple \texttt{Muon}-variant optimizers as well as other state-of-the-art optimizers. \texttt{HTMuon} consistently achieves the best results. Furthermore, combining \texttt{HTMuon} with other \texttt{Muon}-variant optimizers will achieve better performance.  Detailed hyperparameters are reported in Appendix~\ref{app:hyperparameter}.

\begin{table*}[!t]
\centering
\caption{We evaluate \texttt{HTMuon} on LLaMA-60M and LLaMA-135M, comparing it against several \texttt{Muon} variant optimizers and other state-of-the-art optimizers. Red indicates the best  value, and blue denotes the second-best value. We find that \texttt{HTMuon} consistently outperforms these baselines. More hyperparameter details are provided in Appendix~\ref{app:hyperparameter}.}
\scalebox{0.8}{
\begin{tabular}{@{}lccccccccccc@{}}
\toprule
 & \texttt{Muon} & \texttt{HTMuon}  & \texttt{NorMuon} & \texttt{HTMuon+NorMuon} & \texttt{AdaMuon} & \texttt{MARS}  & \texttt{SOAP} & \texttt{Cautious} & \texttt{COSMOS} &\texttt{GaLore} &\texttt{Sophia}\\
\midrule
LLaMa-60M & 28.80 & \textcolor{blue}{27.88} & 28.17  & \textcolor{red}{\textbf{27.55}}&   28.67&29.13 &28.96 &  29.71& 28.62 &34.26 &34.02\\
LLaMa-135M &  22.23& \textcolor{blue}{21.25} &  21.99 & \textcolor{red}{\textbf{21.11}}&   22.43 &  22.42& 22.43 &  23.15&22.15 &25.11&25.63\\
\bottomrule
\end{tabular}}
\label{table: muon_variants}
\vspace{-5pt}
\end{table*}

\begin{figure*}[!htb]
    \centering

    \begin{subfigure}{\linewidth}
        \centering
        \includegraphics[width=0.78\linewidth]{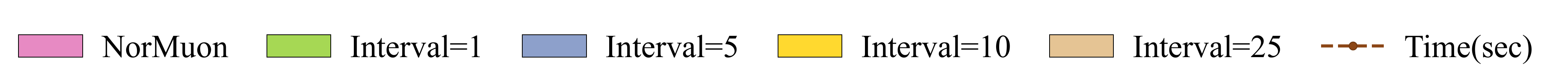}
    \end{subfigure}

    \vspace{2mm}

    \makebox[\linewidth][c]{%
        \begin{subfigure}[b]{0.40\linewidth}
            \centering
            \includegraphics[width=\linewidth]{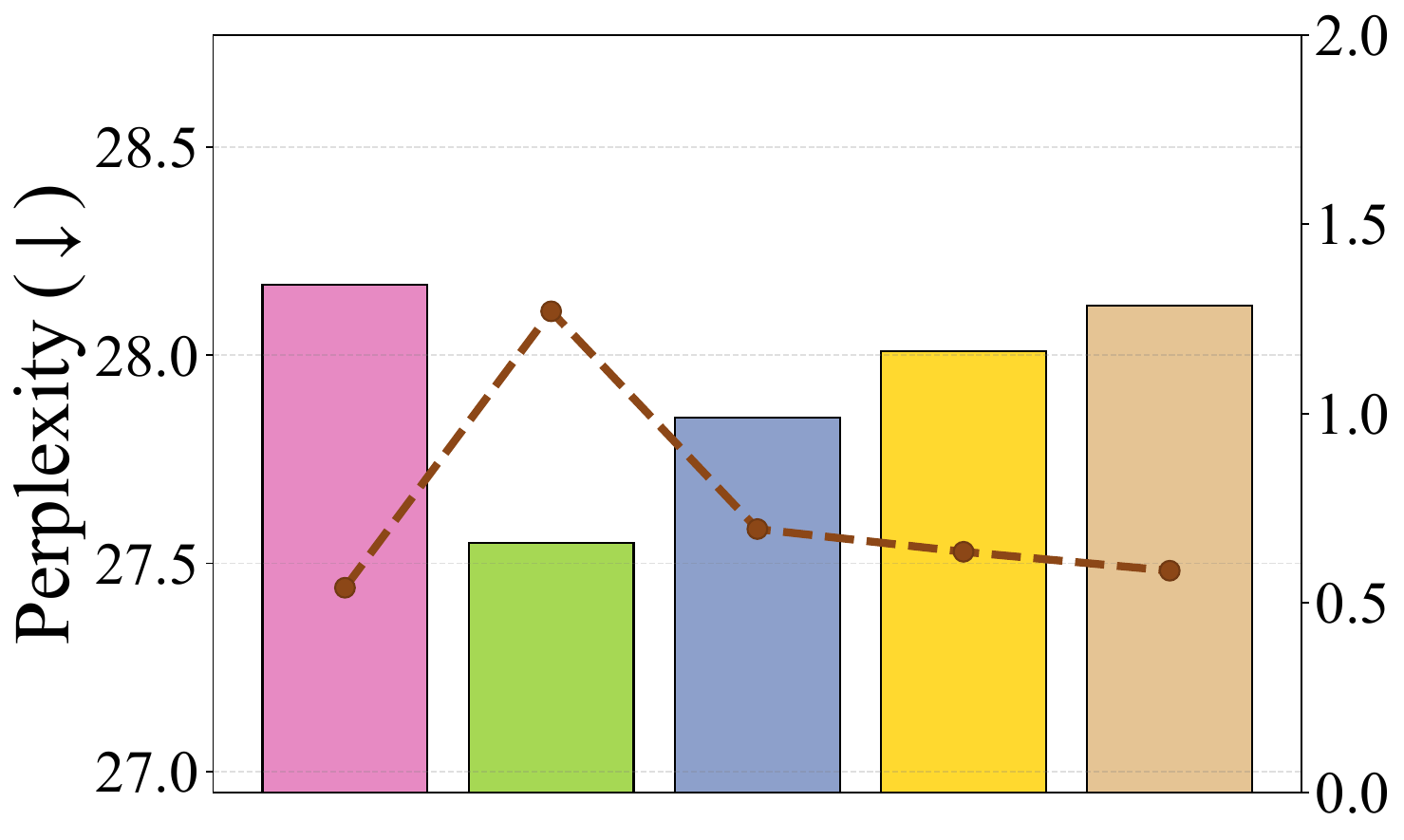}
            \caption{\shortstack{\texttt{HTMuon}$+$\texttt{NorMuon}\\LLaMA-60M}}
            \label{fig:LLaMA-60M_norhtmuon_interval}
        \end{subfigure}
        \begin{subfigure}[b]{0.40\linewidth}
            \centering
            \includegraphics[width=\linewidth]{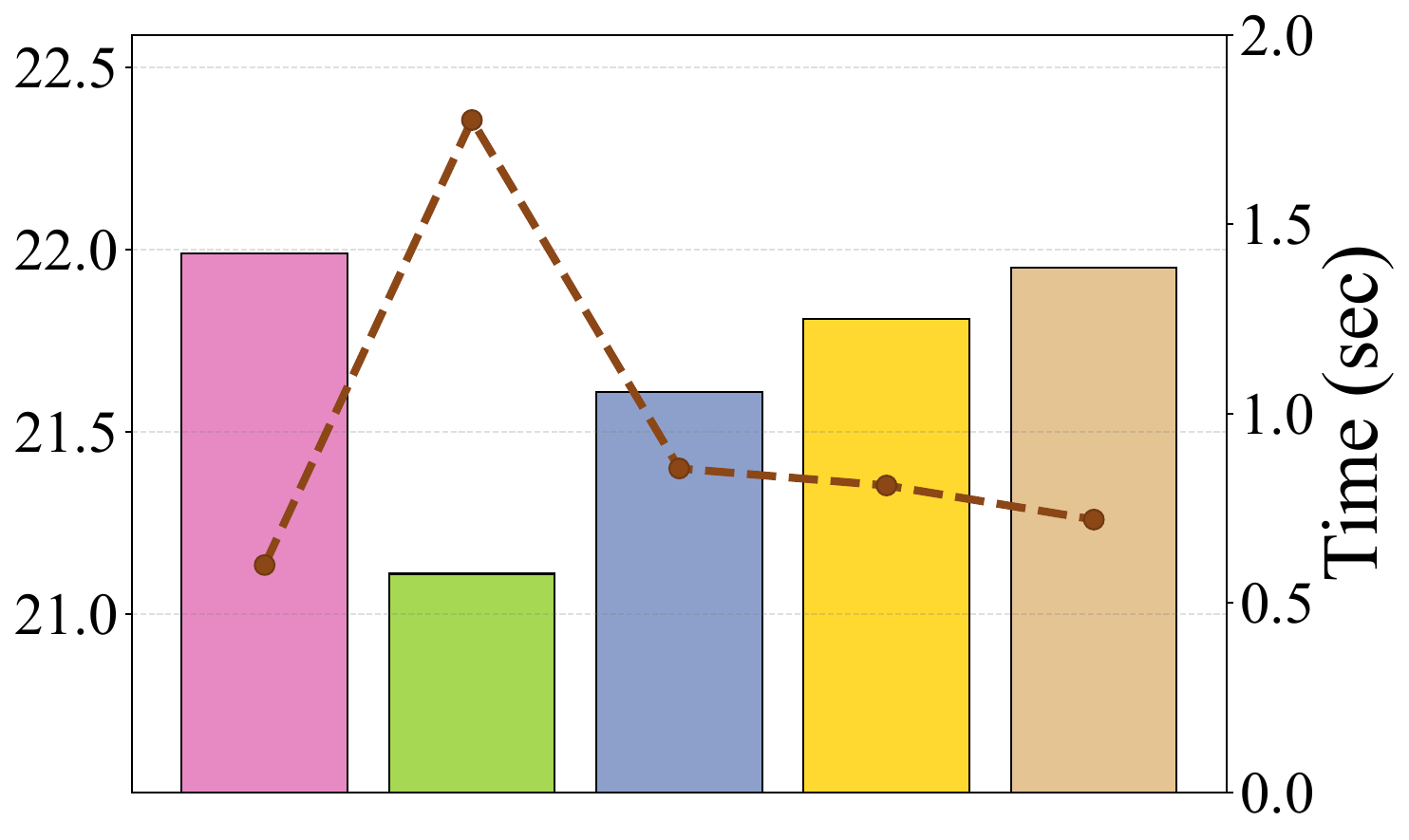}
            \caption{\shortstack{\texttt{HTMuon}$+$\texttt{NorMuon}\\LLaMA-135M}}
            \label{fig:LLaMA-130M_norhtmuon_interval}
        \end{subfigure}%
    }

    \caption{we evaluate applying \texttt{HTMuon}$+$ \texttt{NorMuon} on LLaMA-60M and LLaMA-135M every 1, 5, 10, and 25 steps (while other steps applying \texttt{NorMuon}) . We report the average per-step runtime overhead for all methods. Detailed results and hyperparameter settings are provided in Table~\ref{table:ppl_intervals} and~\ref{table:time_intervals} in Appendix~\ref{app:hyperparameter}.}
    \label{fig:Normuon_interval}
\end{figure*}

In Table~\ref{table:ppl_intervals} and~\ref{table:time_intervals}, 
we report the perplexity  and average per-step runtime for \texttt{HTMuon}, \texttt{HTMuon\_NS}, and \texttt{HTMuon+NorMuon} under different update intervals. We measure per-step runtime on 4 NVIDIA A6000 GPUs. We find that increasing the interval substantially reduces the average overhead of \texttt{HTMuon}, while it still outperforms \texttt{Muon}. Under the same interval, \texttt{HTMuon\_NS} further reduces the average overhead and also remains better than \texttt{Muon}. Similarly, for \texttt{HTMuon+NorMuon}, larger intervals make its average overhead approach that of \texttt{NorMuon}, while consistently outperforming \texttt{NorMuon}. For more visualizations, please refer to Figure~\ref{fig:interval} and~\ref{fig:Normuon_interval}.

\begin{table*}[t]
\centering
\caption{We report the exact PPL obtained when training LLaMA-60M and LLaMA-135M using \texttt{HTMuon}, \texttt{HTMuon\_NS}, and \texttt{HTMuon+NorMuon} under different update intervals. For \texttt{Muon} and \texttt{NorMuon}, interval updates are not applicable; therefore, we only report their PPL after training.}

\setlength{\tabcolsep}{9pt}

\begin{tabular}{l|cccc|cccc}
\toprule
 & \multicolumn{4}{c}{LLaMA-60M} & \multicolumn{4}{|c}{LLaMA-135M} \\
\cmidrule(lr){2-5} \cmidrule(lr){6-9}
Optimizer & 1& 5& 10 & 25& 1 & 5& 10& 25  \\
\midrule
\texttt{Muon} &  &28.80  & & &  & 22.23 & &\\
\texttt{HTMuon}      & 27.88 & 28.36& 28.58 & 28.64 & 21.25& 22.01& 22.13 &22.20 \\
\texttt{HTMuon\_NS}      & 28.03 & 28.42& 28.67 & 28.74 & 21.35& 22.05&22.14 &22.20 \\
\texttt{NorMuon}  &  &28.17 & & &  &21.99  & &\\
\texttt{HTMuon}$+$\texttt{NorMuon} & 27.55&  27.85& 28.01 & 28.12 & 21.11  &21.61&21.81&21.95  \\
\bottomrule
\end{tabular}
\label{table:ppl_intervals}
\end{table*}

\begin{table*}[t]
\centering
\caption{  We report the  average per-step runtime overhead obtained when training LLaMA-60M and LLaMA-135M using \texttt{HTMuon}, \texttt{HTMuon\_NS}, and \texttt{HTMuon+NorMuon} under different update intervals. We measure per-step runtime on 4 NVIDIA A6000 GPUs. For \texttt{Muon} and \texttt{NorMuon}, interval updates are not applicable; therefore, we only report average per-step runtime overhead.}

\setlength{\tabcolsep}{11pt}

\begin{tabular}{l|cccc|cccc}
\toprule
 & \multicolumn{4}{c}{LLaMA-60M} & \multicolumn{4}{|c}{LLaMA-135M} \\
\cmidrule(lr){2-5} \cmidrule(lr){6-9}
Optimizer & 1& 5& 10 & 25& 1 & 5& 10& 25  \\
\midrule
\texttt{Muon} &  &0.51 & & &  & 0.59& &\\
\texttt{HTMuon}      & 1.26 & 0.67& 0.61 & 0.54 & 1.70& 0.82& 0.70&0.63\\
\texttt{HTMuon\_NS}      & 0.66 & 0.59& 0.56 & 0.54 & 0.92& 0.68&0.64 &0.63 \\
\texttt{NorMuon}  &  &0.54 & & &  &0.60 & &\\
\texttt{HTMuon}$+$\texttt{NorMuon} &1.27&  0.70& 0.64 & 0.59 &1.78  &0.86 &0.81&0.72 \\
\bottomrule
\end{tabular}
\label{table:time_intervals}
\end{table*}

In Table~\ref{table:wall_time_clock}, we summarize the PPL and wall-clock time of \texttt{Muon}, \texttt{HTMuon}, \texttt{HTMuon\_NS}, \texttt{HTMuon} (Interval $= 5$), and \texttt{HTMuon\_NS}(Interval $= 5$). To better report the wall-clock time, we rerun training on NVIDIA RTX PRO 6000  GPUs: LLaMA-60M on C4 using 2 GPUs, and LLaMA-135M on C4 using 4 GPUs. We find that although \texttt{HTMuon} can take longer than \texttt{Muon}, the substantial gains achieved through heavy-tailed spectral correction suggest that \texttt{HTMuon} captures the correct inductive bias for learning. Furthermore, we believe that efficient variants such as \texttt{HTMuon\_NS} with interval-based updates provide a strong balance between performance and efficiency. 

Here we also give the FLOPs analysis for \texttt{Muon} and \texttt{HTMuon\_NS}, we consider initial weights $\mW_0 \in \mathbb{R}^{m\times n}$. 
\begin{itemize}
    \item FLOPs analysis for \texttt{Muon}(Algorithm~\ref{algo:MUON}):
    $$F_{Muon} = 20mnr+O(mn).$$
    \item FLOPs analysis for \texttt{HTMuon\_NS}(Algorithm~\ref{algo:HTMUON_NS}):
    $$F_{HTMuon\_NS} = 20mnr+4mn^2+6LTn^3 +O(n^2),$$  where $r=\min(m,n), L=\left[ \log_2\left(\frac{2}{p}\right)\right]$ and T=ns\_steps in Algorithm~\ref{algo:HTMUON_NS}.
    \end{itemize}
Compared to \texttt{Muon}, the additional FLOPs in \texttt{HTMuon\_NS} mainly come from the \texttt{Newton–Schulz} iterations used for the fractional power approximation.

\begin{table*}[t]
\centering
\caption{  We report the  perplexity and wall-clock time of when training LLaMA-60M and LLaMA-135M using \texttt{Muon}, \texttt{HTMuon}, \texttt{HTMuon\_NS}, \texttt{HTMuon}(Interval=5), \texttt{HTMuon\_NS}(Interval=5). We measure wall-clock time on 4 NVIDIA RTX PRO 6000 GPUs.}
\setlength{\tabcolsep}{12pt}

\begin{tabular}{l|cc|cc}
\toprule
 & \multicolumn{2}{c}{LLaMA-60M} & \multicolumn{2}{|c}{LLaMA-135M} \\
\cmidrule(lr){2-3} \cmidrule(lr){4-5}
Optimizer & PPL& Time& PPL& Time(hour)  \\
\midrule
\texttt{Muon} & 28.84 & 0.67  & 22.27 & 1.40\\
\texttt{HTMuon\_NS}(Interval=5) & 28.50 & 0.71& 22.02 &1.55 \\
\texttt{HTMuon}(Interval=5)   & 28.36 &1.02 &22.04 &2.40\\
\texttt{HTMuon\_NS}  & 28.06&0.87& 21.36 &2.17 \\
\texttt{HTMuon}  & 27.87&  2.46& 21.24 & 6.57 \\
\bottomrule
\end{tabular}
\label{table:wall_time_clock}
\end{table*}

\begin{figure*}[!htb]
    \centering

    \begin{subfigure}[b]{0.24\linewidth}
        \centering
        \includegraphics[width=\linewidth]{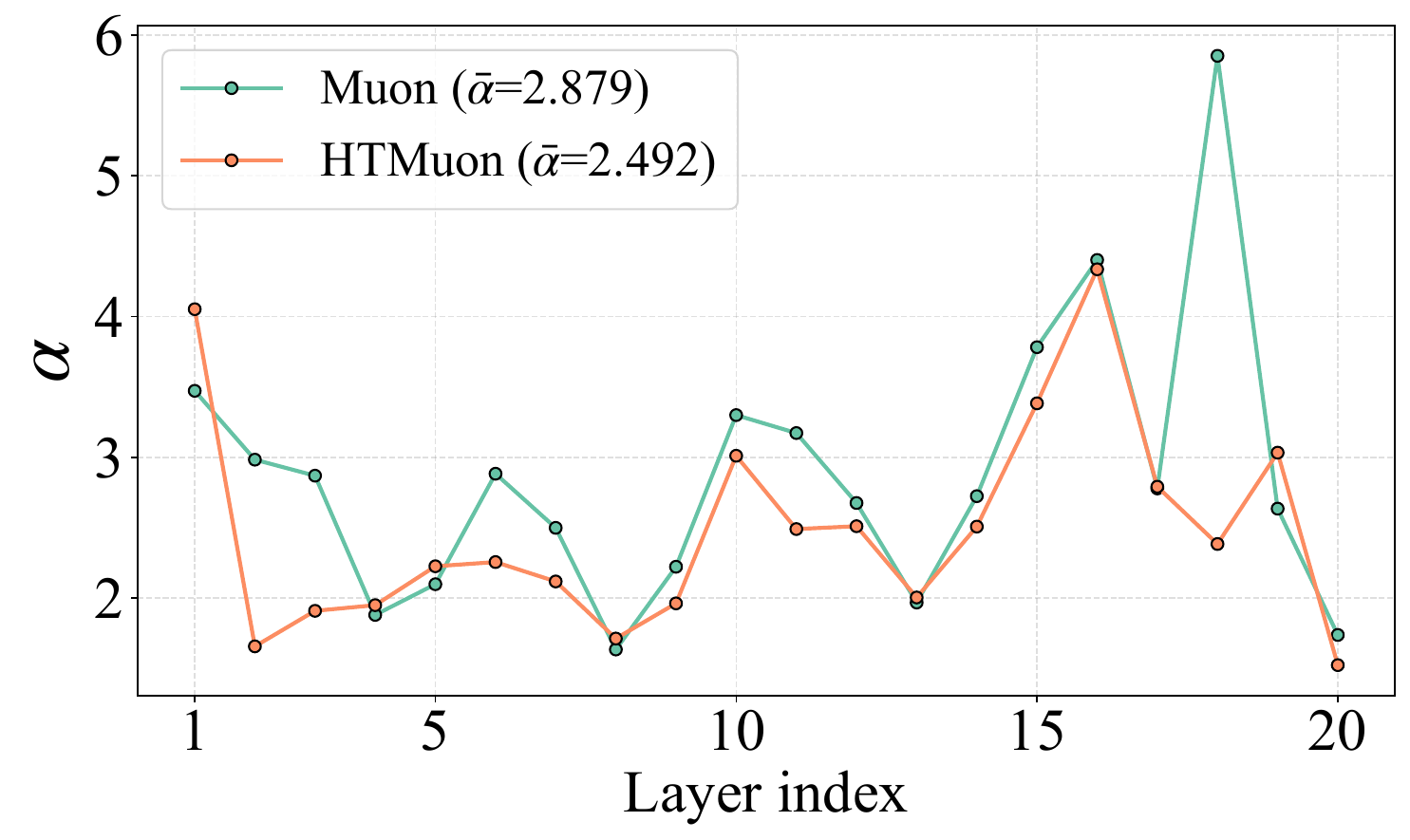}
        \caption{ResNet18 on CIFAR-100}
        \label{fig:Resnet18-alpha-cifar100-alpha}
    \end{subfigure}\hfill
    \begin{subfigure}[b]{0.24\linewidth}
        \centering
        \includegraphics[width=\linewidth]{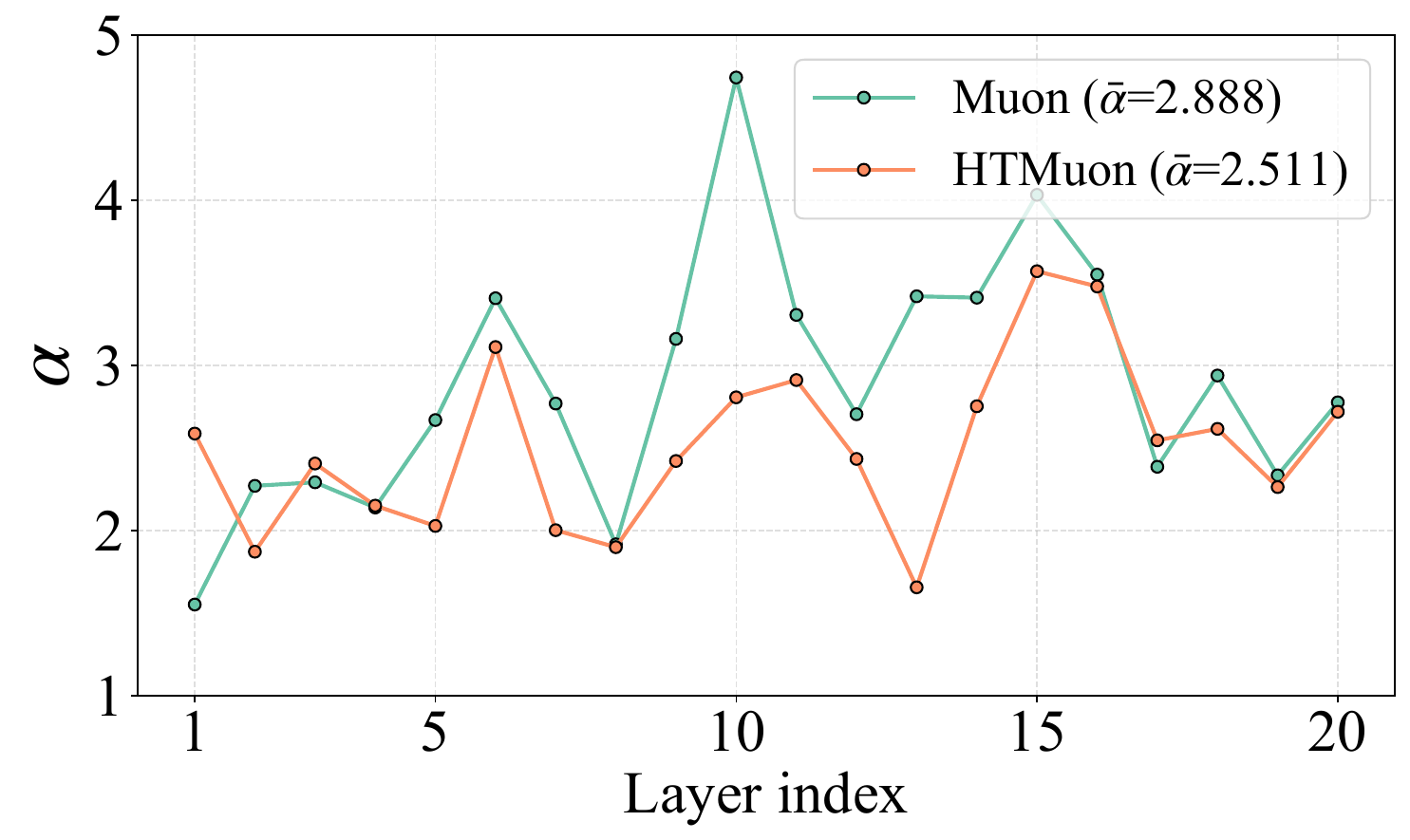}
        \caption{ResNet18 on CIFAR-10}
        \label{fig:Resnet18-alpha-cifar10-alpha}
    \end{subfigure} \hfill
  \begin{subfigure}[b]{0.24\linewidth}
        \centering
        \includegraphics[width=\linewidth]{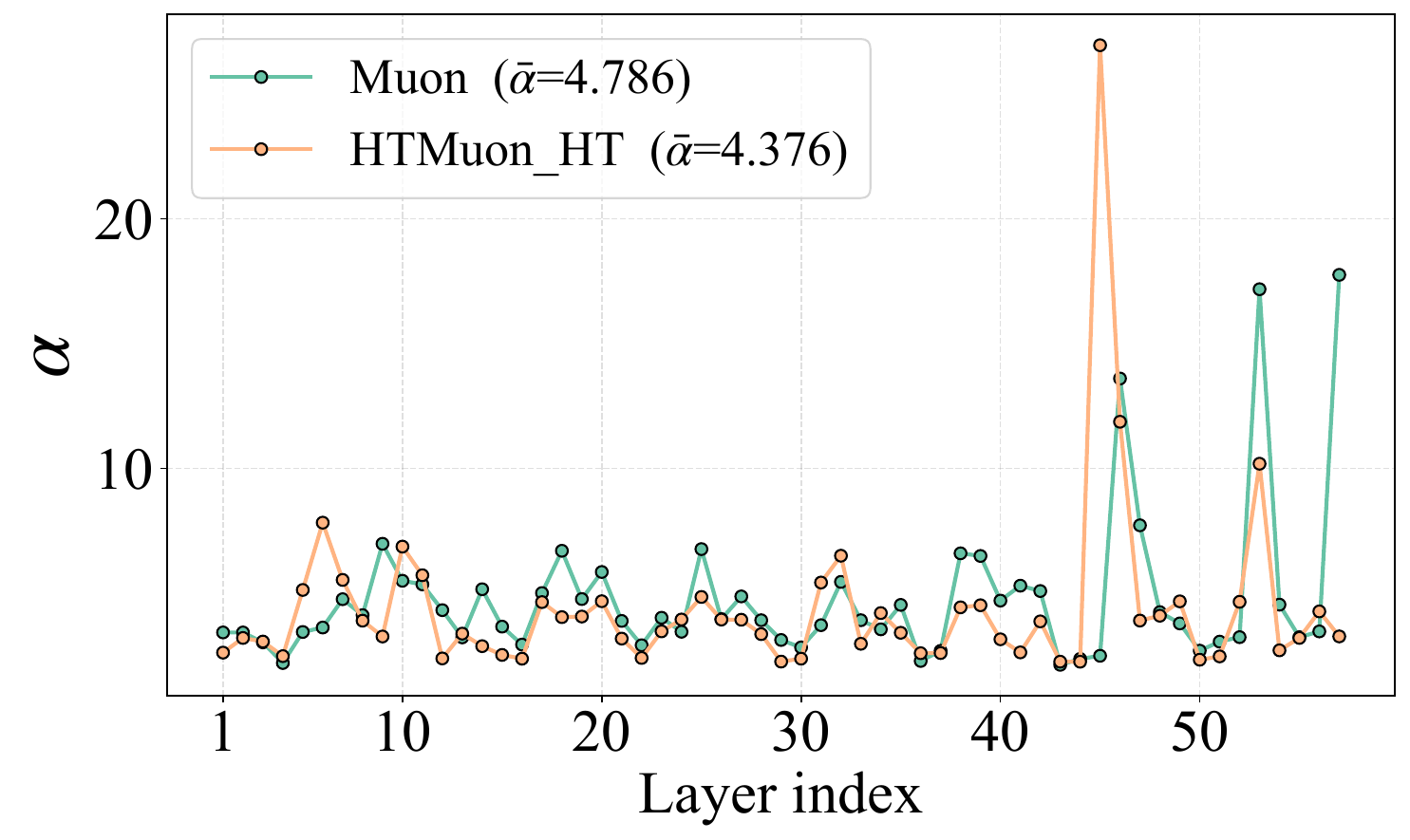}
        \caption{HTMUON\_HT\_60M}
        \label{fig:HTMUON_HT_60M-alpha}
    \end{subfigure}\hfill
    \begin{subfigure}[b]{0.24\linewidth}
        \centering
        \includegraphics[width=\linewidth]{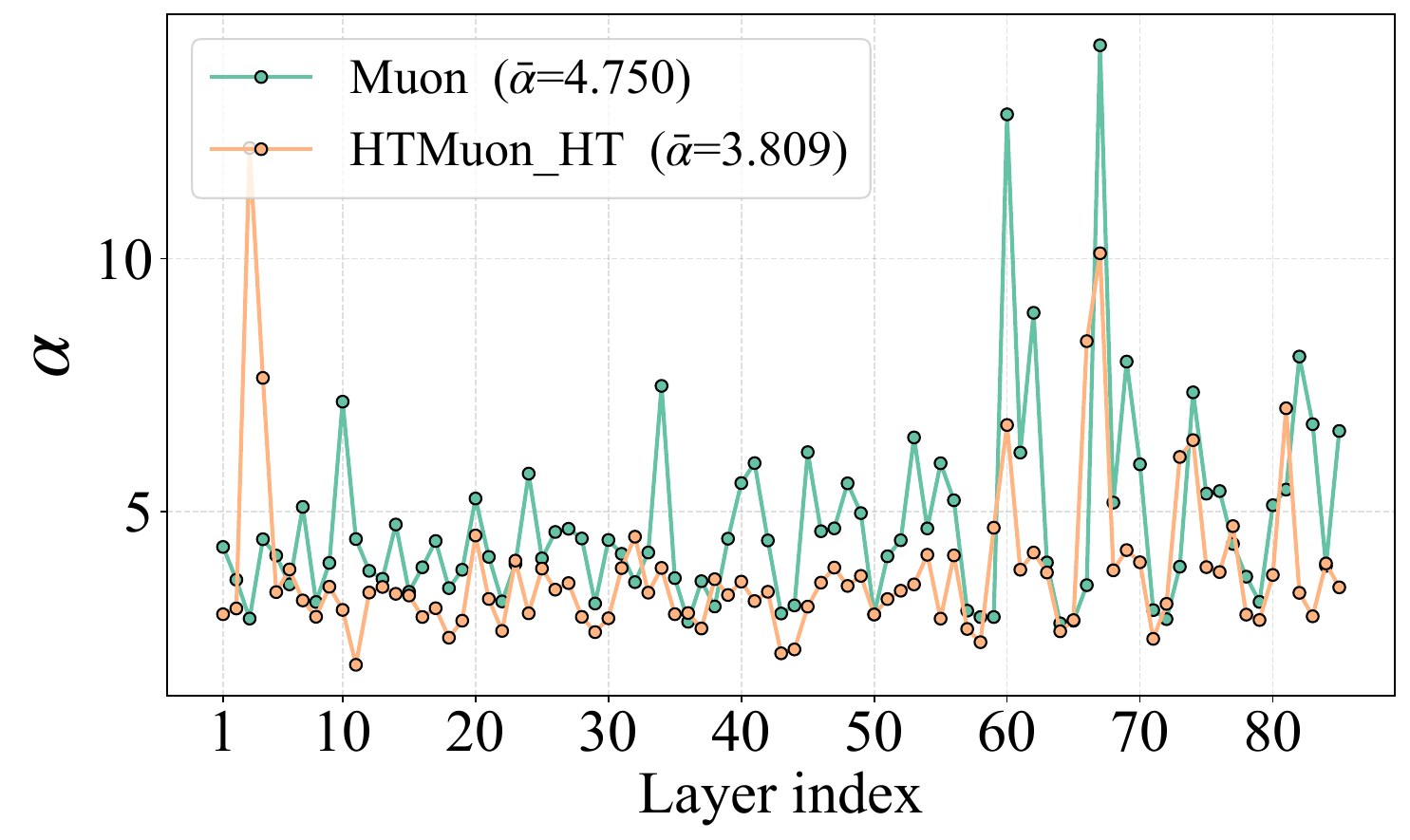}
        \caption{HTMUON\_HT\_135M}
        \label{fig:HTMUON_HT_135M-alpha}
    \end{subfigure}

    \caption{(a)(b): Layer-wise PL $\alpha$   for ResNet-18 model weight ESDs on CIFAR-100 and CIFAR-10 trained with \texttt{Muon} and \texttt{HTMuon}.  (c)(d): Layer-wise PL $\alpha$   for LLaMA model weight ESDs  trained with \texttt{Muon} and \texttt{HTMuon\_HT}. All models used for visualization are trained using each optimizer’s best-performing hyperparameter configuration. For hyperparameter configurations, please refer to Appendix~\ref{app:hyperparameter}.}
\end{figure*}

In Figure~\ref{fig:Resnet18-alpha-cifar100-alpha},~\ref{fig:Resnet18-alpha-cifar10-alpha}, we visualize the layer-wise PL exponent $\alpha$ for ResNet18 on CIFAR-100 and CIFAR-10 dataset to further support that \texttt{HTMuon} makes the updated matrices more heavy-tailed than \texttt{Muon}.

\subsection{More Ablation Study}
\subsubsection{Varying learning rates}
\label{More_abalation_study_lr}
\begin{figure*}[!htb]
    \centering

    \begin{subfigure}{0.8\linewidth}
        \centering
        \includegraphics[width=\linewidth]{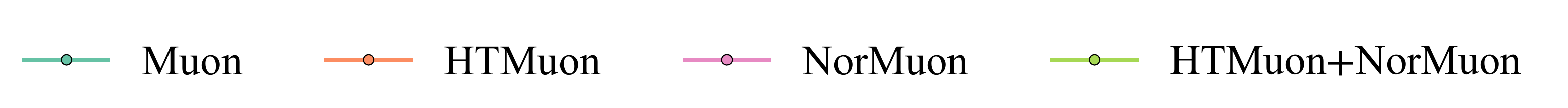}
    \end{subfigure}

    \vspace{2mm} %

    \begin{subfigure}[b]{0.33\linewidth}
        \centering
        \includegraphics[width=\linewidth]{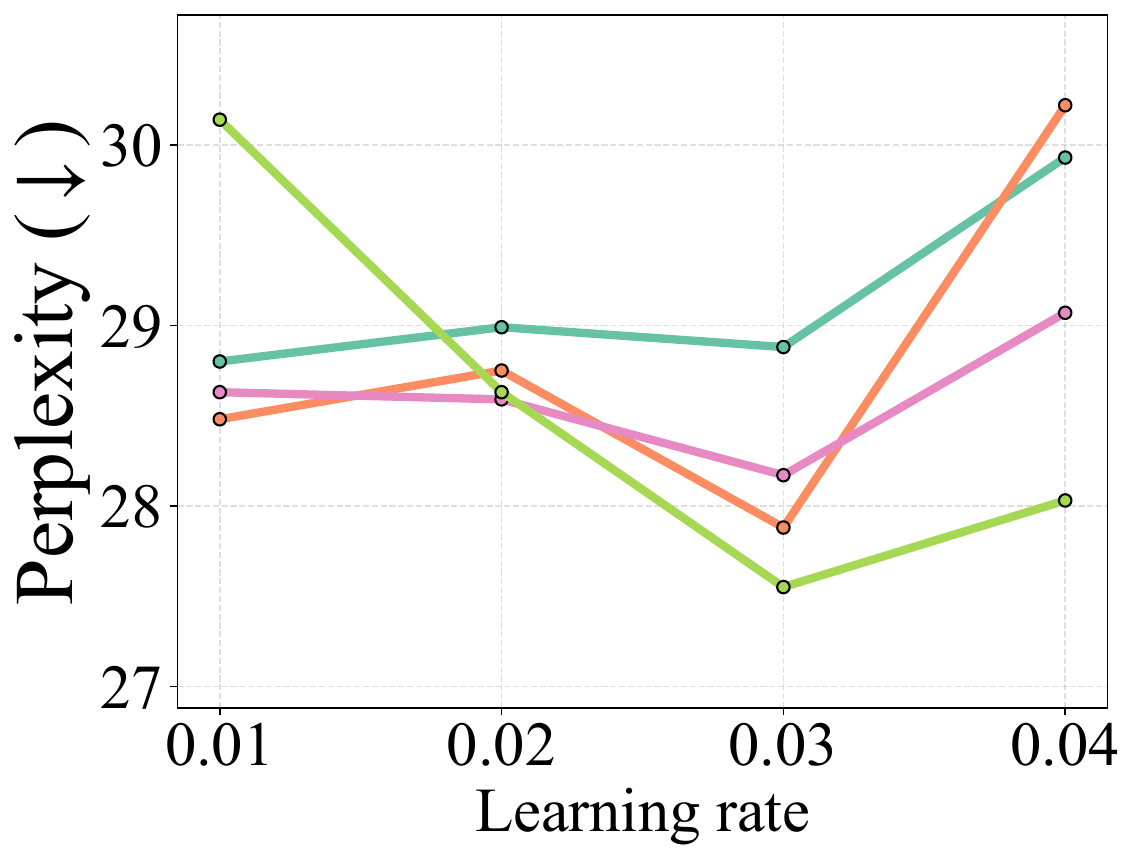}
        \caption{\shortstack{LLaMA-60M on C4}}
        \label{fig:LLaMA-60M_lr}
    \end{subfigure}\hfill
    \begin{subfigure}[b]{0.3\linewidth}
        \centering
        \includegraphics[width=\linewidth]{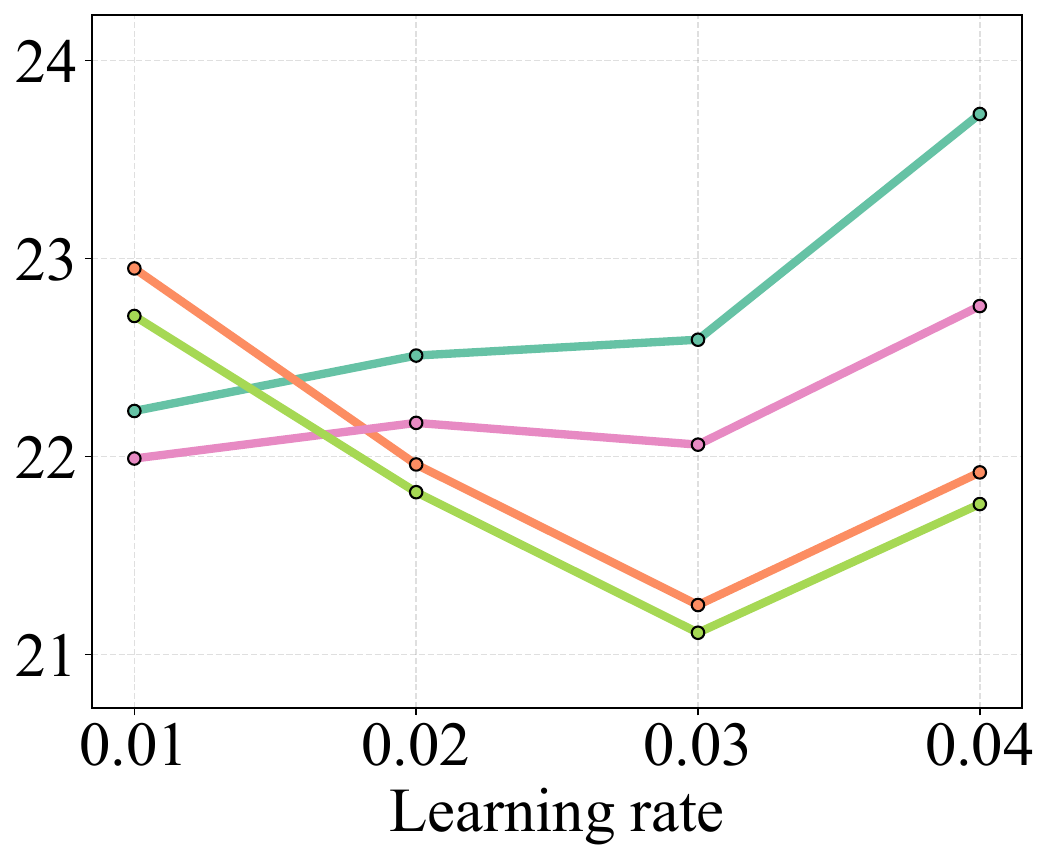}
        \caption{\shortstack{LLaMA-135M on C4}}
        \label{fig:LLaMA-135M_lr}
    \end{subfigure}\hfill
    \begin{subfigure}[b]{0.33\linewidth}
        \centering
        \includegraphics[width=\linewidth]{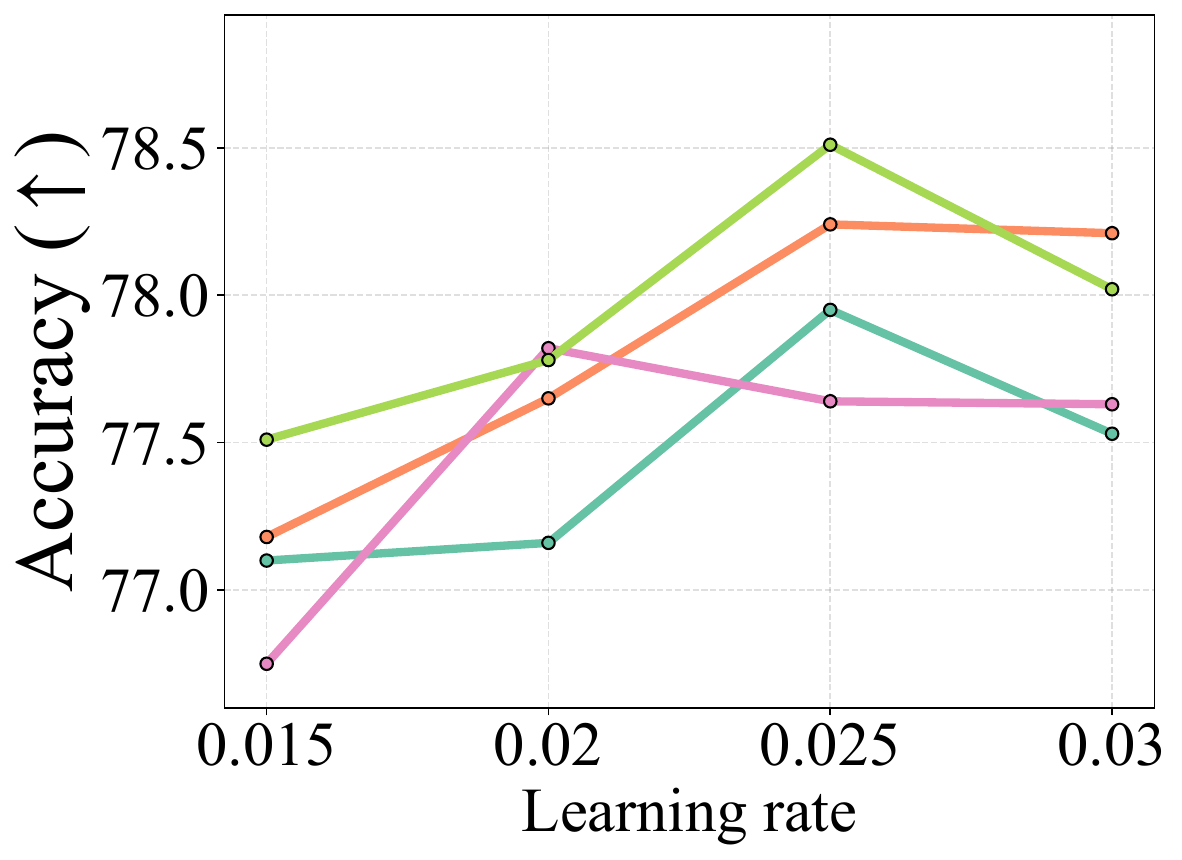}
        \caption{\shortstack{ResNet18 on CIFAR-100}}
        \label{fig:Resnet18-CIFAR100}
    \end{subfigure}\hfill
        \begin{subfigure}[b]{0.33\linewidth}
        \centering
        \includegraphics[width=\linewidth]{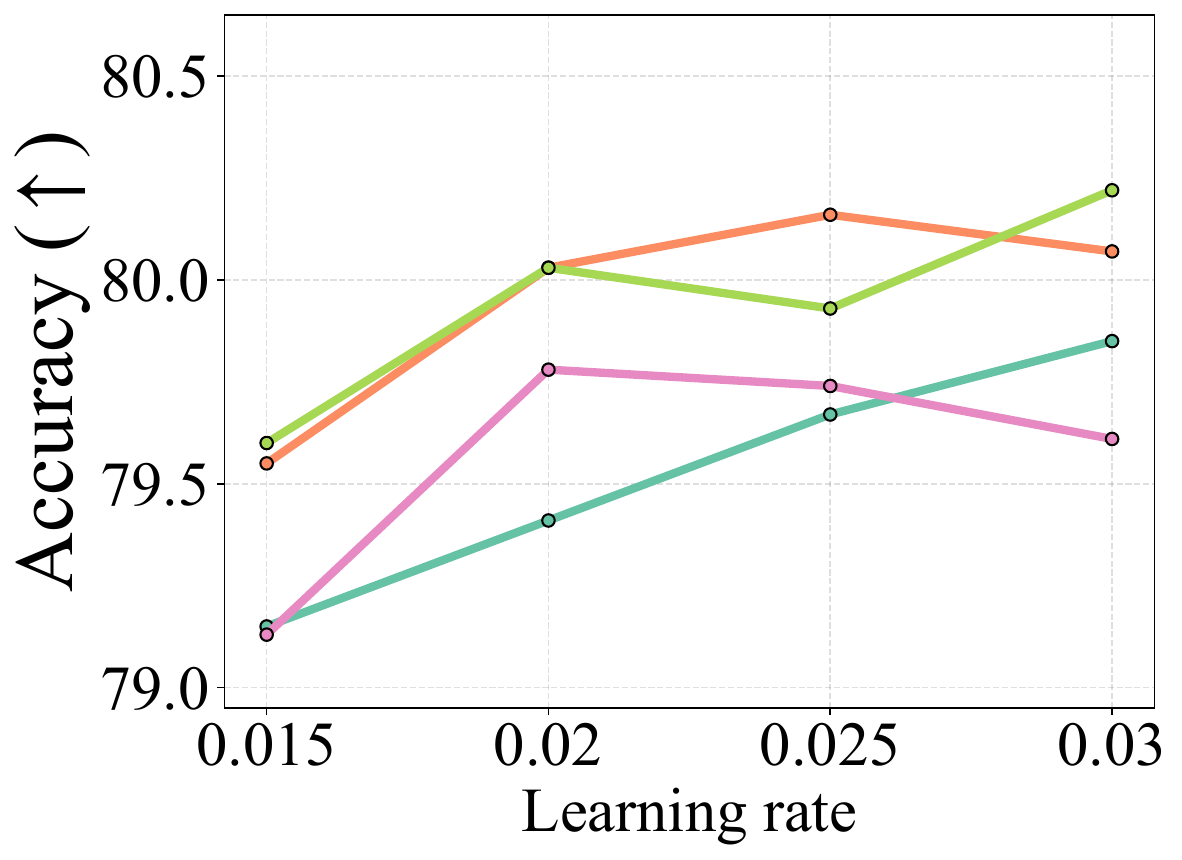}
        \caption{\shortstack{ResNet50 on CIFAR-100}}
        \label{fig:Resnet50-CIFAR100}
    \end{subfigure}\hfill
    \begin{subfigure}[b]{0.3\linewidth}
        \centering
        \includegraphics[width=\linewidth]{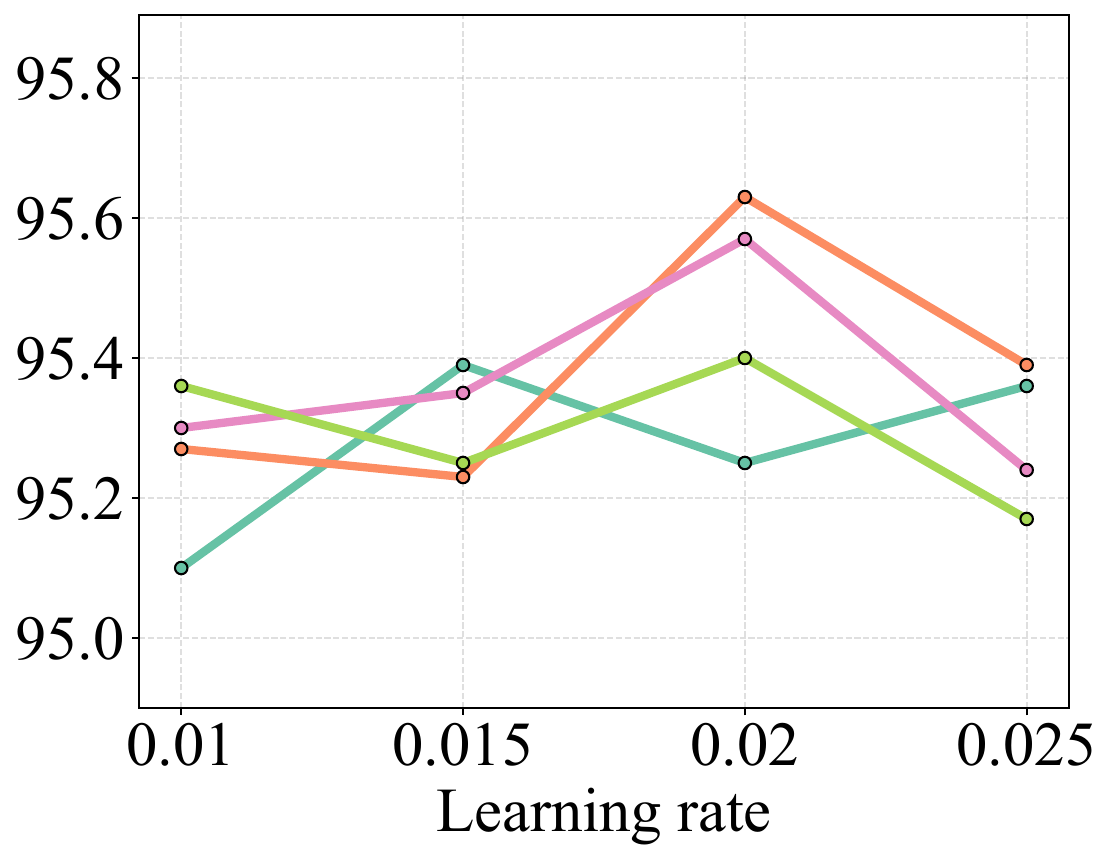}
        \caption{\shortstack{ResNet18 on CIFAR-10}}
        \label{fig:Resnet18-CIFAR10}
    \end{subfigure}\hfill
    \begin{subfigure}[b]{0.3\linewidth}
        \centering
        \includegraphics[width=\linewidth]{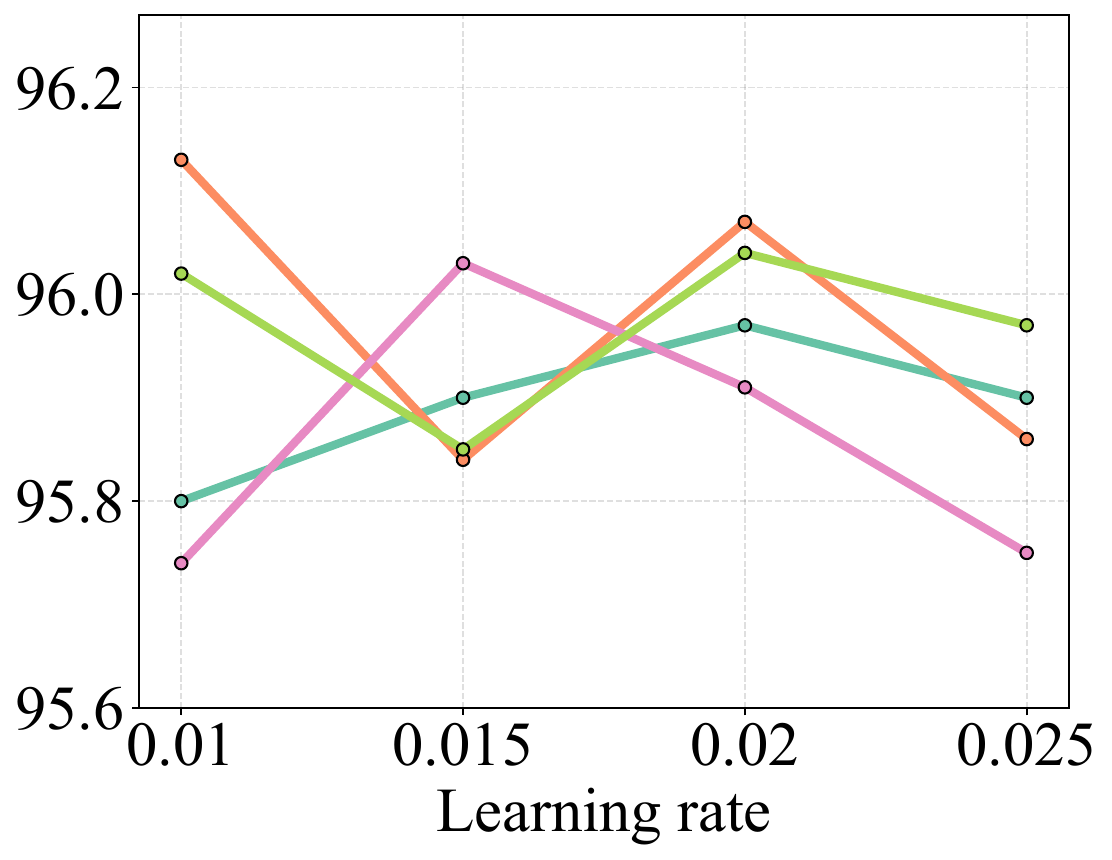}
        \caption{\shortstack{ResNet50 on CIFAR-10}}
        \label{fig:Resnet50-CIFAR100}
    \end{subfigure}\hfill

    \caption{We conduct learning-rate grid searches for \texttt{Muon}, \texttt{NorMuon}, \texttt{HTMuon}, and \texttt{HTMuon+NorMuon} across various datasets and architectures, including LLaMA models on C4 and ResNet models on CIFAR-100 and CIFAR-10. We find that \texttt{HTMuon} outperforms \texttt{Muon} and \texttt{NorMuon} across most learning rates, demonstrating the effectiveness and robustness of our method.
}
    \label{fig:lr}
\end{figure*}

In Figure~\ref{fig:lr}, Table~\ref{Table:Lr grid_C4},~\ref{Table:Lr grid_Cifar100} and~\ref{Table:Lr grid_Cifar10}, we conduct learning grid search on LLaMA  and ResNet models, our experiments show that for most learning rates, \texttt{HTMuon} outperforms \texttt{Muon} and \texttt{NorMuon}, which exhibits our method's effectiveness and robustness. 

\begin{table*}[t]
\centering
\caption{Detailed results for learning rate grid search on C4. Red indicates the best  value, and blue denotes the second-best value.}

\setlength{\tabcolsep}{9pt}

\begin{tabular}{l|cccc|cccc}
\toprule
 & \multicolumn{4}{c}{LLaMA-60M} & \multicolumn{4}{|c}{LLaMA-135M} \\
\cmidrule(lr){2-5} \cmidrule(lr){6-9}
Optimizer & 0.01& 0.02 & 0.03 & 0.04& 0.01 & 0.02 & 0.02 & 0.04  \\
\midrule
\texttt{Muon}        & 28.80 & 28.99 & 28.88 & 29.93 & 22.23&22.51 & 22.59& 23.73\\
\texttt{NorMuon}     & 28.63 & 28.59 & 28.17 &29.07 &21.99&22.17 & 22.06 &22.76 \\
\rowcolor[HTML]{FFF9E5}
\texttt{HTMuon}      & 28.48  & 28.75& \textcolor{blue}{\textbf{27.88}} & 30.22 & 22.95& 21.96& \textcolor{blue}{\textbf{21.25}} & 21.92\\
\rowcolor[HTML]{FFF9E5}
\texttt{HTMuon}$+$\texttt{NorMuon} & 30.14&  28.63& \textcolor{red}{\textbf{27.55}} & 28.03 & 22.71  &21.82 &\textcolor{red}{\textbf{21.11}}  &21.76\\
\bottomrule
\end{tabular}
\label{Table:Lr grid_C4}
\end{table*}

\begin{table*}[t]
\centering
\caption{Detailed results for learning rate grid search  on CIFAR-100. Red indicates the best  value, and blue denotes the second-best value.}

\setlength{\tabcolsep}{9pt}

\begin{tabular}{l|cccc|cccc}
\toprule
 & \multicolumn{4}{c}{ResNet18} & \multicolumn{4}{|c}{ResNet50} \\
\cmidrule(lr){2-5} \cmidrule(lr){6-9}
Optimizer & 0.015 & 0.02 & 0.025 & 0.03 & 0.015 & 0.02 & 0.025 & 0.03 \\
\midrule
\texttt{Muon}        & 77.10 & 77.16 & 77.95 & 77.53 & 79.15 &79.41 & 79.67 & 79.85\\
\texttt{NorMuon}     & 76.75 & 77.82 & 77.64 & 77.63 &79.13 &79.78 &79.74 &79.61 \\
\rowcolor[HTML]{FFF9E5}
\texttt{HTMuon}      & 77.18  & 77.65 & \textcolor{blue}{\textbf{78.24}} & 78.21 & 79.55&80.03& \textcolor{blue}{\textbf{80.16}}  & 80.07\\
\rowcolor[HTML]{FFF9E5}
\texttt{HTMuon}$+$\texttt{NorMuon} & 77.51&  77.78& \textcolor{red}{\textbf{78.51}} & 78.02 & 79.60  &80.03 &79.93 &\textcolor{red}{\textbf{80.22}}\\
\bottomrule
\end{tabular}
\label{Table:Lr grid_Cifar100}
\end{table*}

\begin{table*}[t]
\centering
\caption{Detailed results for learning rate grid search  on  CIFAR-10. Red indicates the best  value, and blue denotes the second-best value.}

\setlength{\tabcolsep}{9pt}

\begin{tabular}{l|cccc|cccc}
\toprule
 & \multicolumn{4}{c}{ResNet18} & \multicolumn{4}{|c}{ResNet50} \\
\cmidrule(lr){2-5} \cmidrule(lr){6-9}
Optimizer & 0.01& 0.015 & 0.02 & 0.025 & 0.01 & 0.015 & 0.02 & 0.025  \\
\midrule
\texttt{Muon}        & 95.10 & 95.39 & 95.25 & 95.36 & 95.8 &95.90 & 95.97& 95.90\\
\texttt{NorMuon}     & 95.30 & 95.35 & 95.57 &95.24 &95.74 &96.03 &95.91 &95.75 \\
\rowcolor[HTML]{FFF9E5}
\texttt{HTMuon}      & 95.27  & 95.23& \textcolor{red}{\textbf{95.63}} & 95.39 & \textcolor{red}{\textbf{96.13}} & 95.84& 96.07 & 95.86\\
\rowcolor[HTML]{FFF9E5}
\texttt{HTMuon}$+$\texttt{NorMuon} & 95.36&  95.25& \textcolor{blue}{\textbf{95.40}} & 95.17 & 96.02  &95.85 &\textcolor{blue}{\textbf{96.04}}  &95.97\\
\bottomrule
\end{tabular}
\label{Table:Lr grid_Cifar10}
\end{table*}

\subsubsection{Varying different $p$}
\label{More_abalation_study_p}
In table~\ref{Table:p}, we give the exact results for ablation study on $p$ in Figure~\ref{fig:p}, we test multiple values of $p$ on LLaMA-60M and LLaMA-135M on the C4 dataset, and find that $p=0.125$ is a robust choice. We therefore use $p=0.125$ in most experiments.

Here we discuss more about hyperparameter sensitivity for $p$. Although introducing $p$ adds a tuning dimension and add hyperparameter search burden, also we only do grid search on LLaMA models to obtain the optimal $p=0.125$, we will clarify that $p=0.125$ can serve as a recommended default for other tasks.

\begin{itemize}
    \item \textbf{$p=0.125$ generalizes well across tasks and architectures.} In our paper, we also use $p=0.125$ for \texttt{HTMuon} when training GPT-2 small on OpenWebText dataset, ResNet-18 and ResNet-50 on CIFAR datasets, Tables~\ref{table:OpenWebText_results}, ~\ref{Table:Lr grid_Cifar100}and ~\ref{Table:Lr grid_Cifar10}  in our paper show that \texttt{HTMuon} consistently outperforms \texttt{Muon}.
    \item \textbf{$p=0.125$ is often optimal or near-optimal across tasks and architectures.} To explore whether $p=0.125$ is also close to optimal on other tasks, we performed grid searches over $p$ on the CIFAR and OpenWebText datasets in Table~\ref{Table:p_CIFAR} and~\ref{Table:p_Openwebtext}. We find that on CIFAR-10, $p=0.125$ achieves the best performance for both ResNet-18 and ResNet-50, and on OpenWebText, p=0.125 also yields the best result for GPT-2 small. While on CIFAR-100 a larger value (e.g., $p=0.2$) can perform slightly better, this does not change our overall conclusion that p=0.125 is a strong and practical recommended default.
\end{itemize}

\begin{table*}[t]
\centering
\caption{Detailed results for  ablation study on $p$ on C4 dataset. Red indicates the best  value, and blue denotes the second-best value.}

\setlength{\tabcolsep}{5pt}

\begin{tabular}{l|ccccc|ccccc}
\toprule
 & \multicolumn{5}{c}{LLaMA-60M} & \multicolumn{5}{|c}{LLaMA-135M} \\
\cmidrule(lr){2-6} \cmidrule(lr){7-11}
Optimizer & 0& 0.1& 0.125 & 0.2& 0.25 & 0& 0.1& 0.125 & 0.2& 0.25  \\
\midrule
HTMuon      & 28.80 & 27.91& \textcolor{blue}{\textbf{27.88}} & 28.56 & 29.97& 22.23&21.33 &\textcolor{blue}{\textbf{21.25}} & 21.66 &22.43\\
HTMuon$+$NorMuon & 28.17&  \textcolor{red}{\textbf{27.48}}& 27.55 & 28.15 & 29.54  &21.99 &21.21&\textcolor{red}{\textbf{21.11}}  &21.33&22.1\\
\bottomrule
\end{tabular}
\label{Table:p}
\end{table*}

\begin{table*}[t]
\centering
\caption{Detailed results for  ablation study on $p$ for \texttt{HTMuon} on CIFAR datasets. Red indicates the best  value.}

\setlength{\tabcolsep}{5pt}

\begin{tabular}{l|ccccc|ccccc}
\toprule
 & \multicolumn{5}{c}{ResNet18} & \multicolumn{5}{|c}{ResNet50} \\
\cmidrule(lr){2-6} \cmidrule(lr){7-11}
Dataset & 0& 0.1& 0.125 & 0.2& 0.25 & 0& 0.1& 0.125 & 0.2& 0.25  \\
\midrule
CIFAR-10     & 95.39 & 95.14& \textcolor{red}{\textbf{95.63}} & 95.46&  95.49&95.97&95.88 &\textcolor{red}{\textbf{95.95}} & 95.95 &95.95\\
CIFAR-100 & 77.95&  77.65& 78.24 & \textcolor{red}{\textbf{78.58}} & 78.32 &79.67 &79.40&80.16  &\textcolor{red}{\textbf{80.30}}&80.27\\
\bottomrule
\end{tabular}
\label{Table:p_CIFAR}
\end{table*}

\begin{table*}[t]
\centering
\caption{Detailed results for  ablation study on $p$ for \texttt{HTMuon} on OpenWebText dataset. Red indicates the best  value.}

\setlength{\tabcolsep}{5pt}

\begin{tabular}{l|ccccc}
\toprule
Optimizer & 0& 0.1& 0.125 & 0.2& 0.25\\
\midrule
\texttt{HTMuon}    & 22.46 & 22.40& \textcolor{red}{\textbf{22.20}} & 22.49 &22.83\\
\bottomrule
\end{tabular}
\label{Table:p_Openwebtext}
\end{table*}

\subsubsection{Varying different ways to make Muon update more heavy-tailed}
\label{app:more_ablation_study_on_HTMuon_HT}
In Algorithm~\ref{algo:HTMUON_HT}, we  explore another way \texttt{HTMuon\_HT} to make \texttt{Muon} updates more heavy-tailed , which directly replaces the momentum matrix’s singular values with a pre-specified heavy-tailed distribution.  We use $\alpha=0.25$ in Algorithm~\ref{algo:HTMUON_HT} for experiments. In Table~\ref{Table:HTMuon_HT}, we  find that, although \texttt{HTMuon\_HT} is not as strong as \texttt{HTMuon}, it consistently outperforms \texttt{Muon}. In Figure~\ref{fig:HTMUON_HT_60M-alpha} and~\ref{fig:HTMUON_HT_135M-alpha}, we visualize the layer-wise PL exponent $\alpha$ for LLaMA-60M and LLaMA-135M, and observe that \texttt{HTMuon\_HT} also yields consistently smaller $\alpha$ than \texttt{Muon} across layers. This suggests that the proposed heavy-tailed spectral correction can indeed improve \texttt{Muon}.

Since some readers maybe confused that the performance improvement  attribution of our method \texttt{HTMuon} is from inducing heavy tails or preserving singular value information.  Here we discuss more about the performance improvement attribution for our algorithm \texttt{HTMuon} based on  \texttt{HTMuon\_HT}. Our motivation for heavy-tail spectral correction is that \texttt{Muon} enforces unit singular values for update matrices. This induces a light-tailed spectrum and can over-emphasize noise-dominated directions, which may limit performance. Guided by HT-SR theory, we therefore consider heavy-tailed spectral correction. We study two ways of introducing heavy-tailed spectra: \texttt{HTMuon\_HT}(which merely induces heavy tails) and \texttt{HTMuon}(which introduces heavy tail spectral correction on top of preserved spectral information). Below, we discuss the benefits of merely inducing heavy tails versus preserving singular value information.
\begin{itemize}
    \item \textbf{Merely inducing a heavy-tailed spectrum already yields non-trivial gains}. Our \texttt{HTMuon\_HT} ablation enforces a rigid, pre-specified heavy-tailed spectral distribution. In Table 12 of our paper, although this design discards the data-driven singular-value structure, \texttt{HTMuon\_HT} still outperforms \texttt{Muon} in perplexity by 0.28 on LLaMA-60M and by 0.58 on LLaMA-135M. We would like to kindly emphasize that this is not "poor performance." First, \texttt{Muon} is a strong baseline: in Table 1, \texttt{Muon} outperforms the widely used Adam/AdamW by up to 3.21 PPL on LLaMA-60M and 135M. Under such a strong baseline, simply replacing the spectrum with a pre-defined heavy-tailed distribution that ignores data information yet still yields gains provides direct evidence that heavy-tail correction itself is effective. Second, the improvement of \texttt{HTMuon\_HT} is not insignificant: as suggested by Table 2 in \citep{liu2025cosmos} and \citep{he2025alphadecay} and Figure 2 in \citep{wen2025fantastic}, after carefully tune the hyperparameters of the baselines on LLaMA/C4, an improvement of $\geq 0.2$ PPL over \texttt{Muon} is generally regarded as non-negligible. For example, \texttt{COSMOS} outperforms \texttt{Muon} by 0.15 PPL for LLaMA-135M in \citep{liu2025cosmos} and \texttt{AlphaDecay} outperforms Adam by 0.11 PPL for LLaMA-1B  in \citep{he2025alphadecay}. Therefore, these results strongly indicate that merely inducing heavy-tail correction is already useful. Moreover, in Figure~\ref{fig:teaser}   of our motivating experiment, we observe that \texttt{Muon\_NS} outperforms \texttt{Muon\_SVD}, this further suggests that strictly enforcing a light-tailed, unit-singular-value spectrum may be overly restrictive, and that relaxing it via heavy-tail correction can be beneficial.
    \item \textbf{Heavy tail spectral correction on top of preserved spectral information is more effective.} We acknowledge  that the data-driven spectral information encoded in singular values is important, as it reflects meaningful structure learned from data. Our method \texttt{HTMuon} therefore performs heavy-tail spectral correction based on the current spectral information: specifically, we raise singular values to the power $p$ (as motivated by Lemma~\ref{lemma:alpha_vs_index}) to give the spectrum a heavier tail than \texttt{Muon}'s unit-spectrum, without discarding the learned geometry. This combination preserves directional information while attenuating noise-dominated directions more than signal-aligned ones compared to \texttt{Muon}, leading to stronger gains than using a rigid, pre-specified heavy-tailed spectrum alone.
\end{itemize}

\begin{table*}[!t]
\centering
\caption{Detailed results for \texttt{HTMuon\_HT}. Red indicates the best  value, and blue denotes the second-best value. We use $\alpha=0.25$ and learning rate $=0.03$ in Algorithm~\ref{algo:HTMUON_HT} for experiments.}
\scalebox{0.99}{
\begin{tabular}{@{}lcccccc@{}}
\toprule
 & \texttt{Muon} & \texttt{HTMuon}  & \texttt{NorMuon} & \texttt{HTMuon+NorMuon} & \texttt{HTMuon\_HT} & \texttt{HTMuon\_HT+NorMuon}  \\
\midrule
LLaMa-60M & 28.8 & \textcolor{blue}{27.88} & 28.17  & \textcolor{red}{\textbf{27.55}}&  28.52 & 28.02\\
LLaMa-135M &  22.23& \textcolor{blue}{21.25} &  21.99 & \textcolor{red}{\textbf{21.11}}&   21.65 & 21.26 \\
\bottomrule
\end{tabular}}
\vspace{-5pt}
\label{Table:HTMuon_HT}
\end{table*}

\section{Detailed Hyperparameters }
\label{app:hyperparameter}
Since in practice our \texttt{Muon} is a hybrid \texttt{MuonwithAdamW} variant\citep{jordan2024muon}, we use the same hyperparameter settings as \texttt{AdamW} for the small subset of parameters updated with \texttt{AdamW}.

\subsection{Hyperparameter settings for LLM pretraining}
For the C4 task, we adopt the LLaMA architecture specifications listed in Table~\ref{Table:Hyperparameters of LLaMA} to ensure reproducibility and consistency with prior work. All model
variants are trained with a uniform maximum sequence length of 256, a batch size of 512, and an
aggregate of 13K tokens per batch. For LLaMA-60M, we run all  experiments on two NVIDIA L40 GPUs or two NVIDIA A6000 GPUs without gradient accumulation; for LLaMA-135M, we run all experiments on four NVIDIA L40  GPUs or four NVIDIA A6000 GPUs without gradient accumulation; for LLaMA-350M, we run all experiments on four NVIDIA RTX PRO 6000 Blackwell without gradient accumulation; for LLaMA-1B, we run all experiments on four NVIDIA RTX PRO 6000 Blackwell with gradient accumulation. 

For the OpenWebText task, we consider  a uniform maximum sequence length of 1024 , a batch size of 480 and 10000 iterations for GPT-2 small.  we run all experiments on four NVIDIA RTX PRO 6000 Blackwell with gradient accumulation=8. We do learning rate grid search on \{0.02, 0.03, 0.04, 0.05, 0.06, 0.07, 0.08, 0.09, 0.1\} for \texttt{Muon} and \texttt{HTMuon}, we set learning rate$=6e-4$ for \texttt{AdamW}. We consider weight decay$=0.1$ for all optimizers.

We apply $p=0.125$ for all LLM pretraining experiments. \textbf{In our experiments, we apply Muon updates to the embedding and output layers during training.}  For more detailed hyperparameter settings, please refer to Table~\ref{table:hyperparam_C4} and ~\ref{table:hyperparam_C4_2}.

\begin{table*}[!t]
\centering
\caption{Hyperparameters of LLaMA models.}
\scalebox{0.99}{
\begin{tabular}{@{}lccccccc@{}}
\toprule
 Params & Hidden  & Intermediate &  Heads & Blocks &  Steps  & Data amount & Batch Size\\
\midrule
60M & 512 & 1376 & 8  & 8&  10K & 1B & 512\\
135M &  768& 2048 &  12& 12&   20K & 2B & 512  \\
350M &  1024& 2736 &  16 & 24&   60K & 6B & 512  \\
1B &  2048& 5461 &  32 & 24&   90K& 9B & 512 \\
\bottomrule
\end{tabular}}
\vspace{-5pt}
\label{Table:Hyperparameters of LLaMA}
\end{table*}

\begin{table*}[t]
\centering
\caption{Hyperparameters for LLaMA-60M and LLaMA-135M On C4 Dataset. The boldfaced values denote the optimal hyperparameters.}

\setlength{\tabcolsep}{11pt}
\renewcommand{\arraystretch}{0.9}

\begin{tabular}{l|cc|cc}
\toprule
 & \multicolumn{2}{c}{LLaMA-60M} & \multicolumn{2}{|c}{LLaMA-135M} \\
\cmidrule(lr){2-3} \cmidrule(lr){4-5}
Optimizer & LR & WD & LR & WD \\
\midrule
\texttt{Adam}       & \textbf{1e-3} & 1e-5 & \textbf{1e-3} & 1e-5 \\
\texttt{AdamW}       & \textbf{1e-3} & 0.1 & \textbf{1e-3} & 0.1\\
\texttt{Muon}     & \{\textbf{0.01},0.02,0.03,0.04\} & 0.1& \{\textbf{0.01},0.02,0.03,0.04\} & 0.1 \\
\texttt{NorMuon} & \{0.01,0.02, \textbf{0.03},0.04\} & 0.1& \{\textbf{0.01},0.02,0.03,0.04\} & 0.1 \\
\texttt{AdaMuon} & \{6e-4, 1e-3, \textbf{5e-3}, 1e-2\} & 0.1& \{ 1e-3, 5e-3, \textbf{1e-2}, 2e-2\}& 0.1 \\
\texttt{MARS} & \{1e-3, 2e-3, \textbf{3e-3}, 4e-3\} & 0.1& \{1e-3, \textbf{2e-3}, 3e-3, 4e-3\}& 0.1 \\
\texttt{SOAP} & \{2e-3, \textbf{3e-3}, 4e-3\} & 0.01& \{ 2e-3, \textbf{3e-3}, 4e-3\}& 0.01 \\
\texttt{Cautious} & \{1e-3, 2e-3, 3e-3, \textbf{4e-3}\} & 0& \{1e-3, \textbf{2e-3}, 3e-3, 4e-3\}& 0 \\
\texttt{COSMOS} & \{1e-3, 2e-3, \textbf{3e-3}, 4e-3\} & 0.1& \{1e-3, \textbf{2e-3}, 3e-3, 4e-3\}& 0.1 \\
\texttt{GaLore} & \{5e-3, 1e-2, \textbf{2e-2}, 3e-2, 4e-2\} & 0& \{5e-3, 1e-2, \textbf{2e-2}, 3e-2, 4e-2\}& 0 \\
\texttt{Sophia} & \{1e-4, 2e-4,3e-4, \textbf{4e-4}, 5e-4, 6e-4\} & 0.1& \{1e-4, \textbf{2e-4},3e-4, 4e-4, 5e-4, 6e-4\} & 0.1 \\
\texttt{HTMuon}      & \{0.01,0.02,\textbf{0.03},0.04\} & 0.1& \{0.01,0.02,\textbf{0.03},0.04\} & 0.1 \\
\texttt{HTMuon}+\texttt{NorMuon}      & \{0.01,0.02,\textbf{0.03},0.04\} & 0.1& \{0.01,0.02,\textbf{0.03},0.04\} & 0.1 \\
\bottomrule
\end{tabular}
\label{table:hyperparam_C4}
\end{table*}

\begin{table*}[t]
\centering
\caption{Hyperparameters for LLaMA-350M and LLaMA-1B On C4 Dataset. The boldfaced values denote the optimal hyperparameters.}

\setlength{\tabcolsep}{11pt}
\renewcommand{\arraystretch}{0.9} 

\begin{tabular}{l|cc|cc}
\toprule
 & \multicolumn{2}{c}{LLaMA-350M} & \multicolumn{2}{|c}{LLaMA-1B} \\
\cmidrule(lr){2-3} \cmidrule(lr){4-5}
Optimizer & LR & WD & LR & WD \\
\midrule
\texttt{Adam}       & \textbf{1e-3} & 1e-5 & \textbf{6e-4} & 1e-6 \\
\texttt{AdamW}       & \textbf{1e-3} & 0.1 & \textbf{6e-4} & 0.1\\
\texttt{Muon}     & \{0.0025, \textbf{0.005},0.01,0.015\} & 0.1& \{\textbf{0.005},0.01\} & 0.1 \\
\texttt{HTMuon}      & \{0.0025, \textbf{0.005},0.01,0.015\} & 0.1& \{\textbf{0.005},0.01\} & 0.1 \\
\bottomrule
\end{tabular}
\label{table:hyperparam_C4_2}
\end{table*}

\subsection{Hyperparameter settings for Image  classification}
For training on CIFAR datasets, we set batch size $=512$ and we set $p=0.125$. We run all the experiments on one NVIDIA L40 GPU. For more detailed hyperparameter settings, please refer to Table~\ref{table: hyperparam_Cifar100} and~\ref{table: hyperparam_Cifar10}.

For training on ImageNet-1K datasets, we set batch size $=1024$ and we set $p=0.03125$. We run all the experiments on one NVIDIA RTX PRO 6000 Blackwell. We set learning rate for \{0.003, 0.004, 0.005\} for \texttt{Adam}, \texttt{Muon}, \texttt{HTMuon}.

\begin{table*}[t]
\centering
\caption{Hyperparameters for ResNet18 and  ResNet50 On CIFAR-100 Dataset. The boldfaced values denote the optimal hyperparameters.}

\setlength{\tabcolsep}{11pt}
\renewcommand{\arraystretch}{0.9} 

\begin{tabular}{l|cc|cc}
\toprule
 & \multicolumn{2}{c}{ResNet18} & \multicolumn{2}{|c}{ResNet50} \\
\cmidrule(lr){2-3} \cmidrule(lr){4-5}
Optimizer & LR & WD & LR & WD \\
\midrule
\texttt{SGDM}      & \{0.1,0.2,\textbf{0.3},0.4\}  & 5e-4 & \{0.1,0.2,\textbf{0.3},0.4\}  & 5e-4 \\
\texttt{Muon}     & \{0.015,0.02,\textbf{0.025},0.03\} & 0.1&  \{0.015,0.02,0.025,\textbf{0.03}\}  & 0.1 \\
\texttt{NorMuon} & \{0.015,\textbf{0.02},0.025,0.03\} & 0.1&  \{0.015,\textbf{0.02},0.025,0.03\}  & 0.1 \\
\texttt{HTMuon}     & \{0.015,0.02,\textbf{0.025},0.03\} & 0.1& \{0.015,0.02,\textbf{0.025},0.03\}  & 0.1 \\
\texttt{HTMuon}+\texttt{NorMuon}     & \{0.015,0.02,\textbf{0.025},0.03\} & 0.1& \{0.015,0.02, 0.025, \textbf{0.03}\}  & 0.1 \\
\bottomrule
\end{tabular}
\label{table: hyperparam_Cifar100}
\end{table*}

\begin{table*}[t]
\centering
\caption{Hyperparameters for ResNet18 and  ResNet50 On CIFAR-10 Dataset. The boldfaced values denote the optimal hyperparameters.}

\setlength{\tabcolsep}{11pt}
\renewcommand{\arraystretch}{0.9} 

\begin{tabular}{l|cc|cc}
\toprule
 & \multicolumn{2}{c}{ResNet18} & \multicolumn{2}{|c}{ResNet50} \\
\cmidrule(lr){2-3} \cmidrule(lr){4-5}
Optimizer & LR & WD & LR & WD \\
\midrule
\texttt{SGDM}      & \{0.1,\textbf{0.2},0.3,0.4\}  & 5e-4 & \{0.1,0.2,\textbf{0.3},0.4\}  & 5e-4 \\
\texttt{Muon}     & \{0.01,\textbf{0.015},0.02,0.025\} & 0.1&  \{0.01,0.015,\textbf{0.02},0.025\}  & 0.1 \\
\texttt{NorMuon} & \{0.01,0.015,\textbf{0.02},0.025\} & 0.1&  \{0.01,\textbf{0.015},0.02,0.025\}  & 0.1 \\
\texttt{HTMuon}    & \{0.01,0.015,\textbf{0.02},0.025\} & 0.1&  \{\textbf{0.01},0.015,0.02,0.025\}  & 0.1 \\
\texttt{HTMuon}+\texttt{NorMuon}    & \{0.01,0.015,\textbf{0.02},0.025\} & 0.1&  \{0.01,0.015,\textbf{0.02},0.025\}  & 0.1 \\
\bottomrule
\end{tabular}
\label{table: hyperparam_Cifar10}
\end{table*}

\section{Baseline Optimizers}
\label{app:optimizerdesign}
In this section, we provide the algorithms for all optimizers evaluated in our study.
We adopt the following notation: $w_t$ denotes the model parameters at step $t$,
$g_t$ the corresponding gradient, $\eta$ the learning rate, $\lambda$ the weight decay,
$\beta_1$ and $\beta_2$ the moment decay rates, $\epsilon$ a numerical stability constant,
$g_{\text{norm}}$ the gradient norm, and $m$ and $v$ the first and second moments, respectively.
All operations are element-wise unless stated otherwise.

\noindent\textbf{\texttt{NorMuon}} \citep{li2025normuon}: A \texttt{Muon}-based optimizer that applies normalized and orthogonalized gradient updates with shape-aware scaling. We put the implementations in Algorithm~\ref{alg:normuon}.

\noindent\textbf{\texttt{AdaMuon}} \citep{si2025adamuon}: An adaptive variant of \texttt{Muon} that incorporates second-moment information into orthogonalized updates. We put the implementations in Algorithm~\ref{alg:adamuon}.

\noindent\textbf{\texttt{MARS}} \citep{yuan2024mars}: A momentum-based adaptive optimizer included as a representative adaptive baseline. We put the implementations in Algorithm~\ref{alg:mars}.

\noindent\textbf{\texttt{SOAP}} \citep{vyas2024soap}: An optimizer that applies stochastic orthogonalization or projection to gradient updates. We put the implementations in Algorithm~\ref{alg:soap}.

\noindent\textbf{\texttt{Cautious}} \citep{liang2024cautious}: An optimizer that modifies update application in a conservative manner based on gradient information. We put the implementations in Algorithm~\ref{alg:cautious}.

\noindent\textbf{\texttt{COSMOS}} \citep{liu2025cosmos}: An optimizer that incorporates geometry-aware scaling into its parameter updates. We put the implementations in Algorithm~\ref{alg:COSMOS}.

\noindent\textbf{\texttt{GaLore}} \citep{zhao2024galore}: A memory-efficient optimizer that performs low-rank gradient projection to reduce optimizer-state and update costs, enabling large-model training under limited GPU memory. We put the implementations in Algorithm~\ref{alg:GaLore}.

\noindent\textbf{\texttt{Sophia}} \citep{liu2024sophiascalablestochasticsecondorder}: : A second-order optimizer that uses a Hessian-based (diagonal) curvature estimate to precondition gradients and applies clipped updates for stability and efficiency in large-scale training. We put the implementations in Algorithm~\ref{alg:Sophia}.

\begin{algorithm*}[ht]
  \caption{\texttt{NorMuon}}
  \label{alg:normuon}
  \begin{algorithmic}[1]

  \State \textbf{Input:} Initial weights $\mathbf{W}_0 \in \mathbb{R}^{m\times n}$, loss function $L$, learning rate $\eta$, momentum parameters $(\beta_1,\beta_2)$, 
        perturbation parameter $\varepsilon$, weight decay $\lambda$.

  \State Initialize momentum $\mathbf{M}_0 \in \mathbb{R}^{m\times n} \leftarrow \mathbf{0}$, 
        second moment $\mathbf{v}_0 \in \mathbb{R}^{m} \leftarrow \mathbf{0}$

  \For{$t = 1,2,\ldots$}

    \State $\mathbf{G}_t \leftarrow \nabla_{\mathbf{W}} L(\mathbf{W}_t)$

    \State $\mathbf{M}_t \leftarrow 
           \beta_1 \mathbf{M}_{t-1} + (1-\beta_1)\mathbf{G}_t$

    \State $\mathbf{O}_t \leftarrow \mathrm{NS5}(\mathbf{M}_t)$

    \State $\mathbf{v}_t \leftarrow 
           \beta_2 \mathbf{v}_{t-1} 
           + (1-\beta_2)\operatorname{mean}_{\text{cols}}
             (\mathbf{O}_t \odot \mathbf{O}_t)$

    \State $\mathbf{V}_t \leftarrow \mathrm{ExpandRows}(\mathbf{v}_t)$ 
           \hfill ($\mathbf{V}_t \in \mathbb{R}^{m\times n}$)

    \State $\widehat{\mathbf{O}}_t \leftarrow 
           \mathbf{O}_t \oslash (\sqrt{\mathbf{V}_t} + \varepsilon)$

    \State $\hat{\eta} \leftarrow 
           0.2\,\eta\,\sqrt{mn} \big/ \|\widehat{\mathbf{O}}_t\|_F$

    \State $\mathbf{W}_{t+1} \leftarrow 
           \mathbf{W}_t - \eta\lambda\mathbf{W}_t 
           - \hat{\eta}\,\widehat{\mathbf{O}}_t$

  \EndFor

  \end{algorithmic}
\end{algorithm*}

\begin{algorithm*}[ht]
  \caption{\texttt{AdaMuon}}
  \label{alg:adamuon}
  \begin{algorithmic}[1]

    \State \textbf{Input:} Initial 2D-weights $\mathbf{W}_0 \in \mathbb{R}^{n \times m}$, 
           loss function $\mathcal{L}$, learning rate $\eta$, weight decay $\lambda$, 
           momentum $\beta$, Newton--Schulz steps $T$, small constant $\varepsilon$.

    \State \textbf{Output:} Updated weights $\mathbf{W}$

    \State Initialize first-momentum $\mathbf{M}_0 \leftarrow \mathbf{0}$, 
           second-momentum $\mathbf{V}_0 \leftarrow \mathbf{0}$

    \For{each iteration $t = 1,2,\ldots$}

        \State Compute gradient:
               $\mathbf{G}_t = \nabla_{\mathbf{W}_t}\mathcal{L}(\mathbf{W}_t)$

        \State Update first momentum:
               $\mathbf{M}_t = \beta \cdot \mathbf{M}_{t-1} + \mathbf{G}_t$

        \State Compute sign-stabilized orthogonal direction: $\mathbf{O}_t = \text{Newton--Schulz}(\mathrm{Sign}(\mathbf{M}_t),\,T)$

        \State Update second momentum: $\mathbf{V}_t = \beta \cdot \mathbf{V}_{t-1}
               + (1-\beta) \cdot (\mathbf{O}_t \odot \mathbf{O}_t)$

        \State Apply second momentum update: $\hat{\mathbf{O}}_t 
               = \mathbf{O}_t \oslash (\sqrt{\mathbf{V}_t} + \varepsilon \cdot \mathbf{1})$

        \State RMS-aligned: $\gamma_t = \frac{0.2 \cdot \sqrt{mn}}{\|\hat{\mathbf{O}}_t\|_F}$

        \State Update weights: $\mathbf{W}_{t+1} = \mathbf{W}_t - \eta \left(\gamma_t \hat{\mathbf{O}}_t + \lambda \mathbf{W}_t\right)$

    \EndFor

  \end{algorithmic}
\end{algorithm*}

\begin{algorithm*}[ht]
  \caption{\texttt{MARS}}
  \label{alg:mars}
  \begin{algorithmic}[1]
  \State \textbf{Hyperparameters:} $\beta_1,\;\beta_2,\;\gamma,\;\epsilon,\;\eta,\; \lambda,\;g_{norm}$

  \State \textbf{State:} $m,\;v,\;g_{t-1}$

  \For{$t=1,2,\ldots$}
    \State $c_t = g_t + \gamma\,\frac{\beta_1}{1-\beta_1}\,(g_t - g_{t-1})$ 
    \State $\hat c_t = c_t \max\{1, \frac{g_{norm}}{\|c_t\|_2}\},$ 
    \State $ m_t = \beta_1\,m_{t-1} + (1-\beta_1)\, \hat c_t$
    \State $v_t = \beta_2\,v_{t-1} + (1-\beta_2)\, \hat c_t^2$
    \State $\hat m_t = \frac{m_t}{1-\beta_1^t},\quad \hat v_t = \frac{v_t}{1-\beta_2^t}$ 
    \State $w_{t+1} = w_t - \eta\,\frac{\hat m_t}{\sqrt{\hat v_t} + \epsilon} - \eta\,\lambda\,w_t.$
 \EndFor
 \end{algorithmic}
\end{algorithm*}

\begin{algorithm*}[ht]
  \caption{\texttt{SOAP}}
  \label{alg:soap}
  \begin{algorithmic}[1]
  \State \textbf{Hyperparameters:} $\beta_1,\;\beta_2,\;\mu,\;k,\;\epsilon,\text{block\_size}, g_{norm}$
  \State Partition all parameters into \text{block\_size} × \text{block\_size}
  \State \textbf{Update Rules for Each Block:}
  \For{$t=1,2,\ldots$} 
    \State $\hat g_t = g_t \max\{1, \frac{g_{norm}}{\|g_t\|_2}\} $
    \State $\hat g_t = Q_A\,\hat g_t\,Q_B$
    \State $m_t = \beta_1\,m_{t-1} + (1-\beta_1)\,\hat g_t,\quad v_t = \beta_2\,v_{t-1} + (1-\beta_2)\,\hat g_t^2$
    \State$\hat m_t = \frac{m_t}{1-\beta_1^{t}}, \hat v_t = \frac{v_t}{1-\beta_2^{t}}$
    \State $w_{t+1} = w_t - \eta_t\;Q_A^\top\! \Bigl(\frac{\hat m_t}{\sqrt{\hat v_t} + \epsilon}\Bigr)\!Q_B^\top$
   \State  $G_A = \mu\,G_A + (1-\mu)\,\hat g_t\,\hat g_t^\top + \epsilon I,\quad G_B = \mu\,G_B + (1-\mu)\,\hat g_t^\top\,\hat g_t+ \epsilon I$
   \State $\text{if }t \bmod k = 0: \quad Q_A = \mathrm{QR}(G_A\,Q_A),\quad Q_B = \mathrm{QR}(G_B\,Q_B)$
  \EndFor
  \end{algorithmic}
\end{algorithm*}

\begin{algorithm*}[ht]
  \caption{\texttt{Cautious}}
  \label{alg:cautious}
  \begin{algorithmic}[1]
  \State \textbf{Hyperparameters:} $\beta_1,\;\beta_2,\;\epsilon,\;\eta,\;\lambda,\;g_{norm}$
  \State \textbf{State:} $m,\;v$
  \For{$t=1,2,\ldots$} 
    \State $\hat g_t = g_t \max\{1, \frac{g_{norm}}{\|g_t\|_2}\}$
    \State $m_t = \beta_1\,m_{t-1} + (1-\beta_1)\,\hat g_t$
    \State $ v_t = \beta_2\,v_{t-1} + (1-\beta_2)\,\hat g_t^2$
    \State $\hat m_t = \frac{m_t}{1-\beta_1^t},\quad \hat v_t = \frac{v_t}{1-\beta_2^t}$
    \State $u_t = \frac{\hat m_t}{\sqrt{\hat v_t} + \epsilon},\quad s_t = \mathbb{I}\bigl(u_t\cdot \hat g_t > 0\bigr)$
    \State $\hat u_t = \frac{u_t \cdot s_t}{\mathrm{mean}(s_t)},\quad w_{t+1} = w_t - \eta\,\hat u_t - \eta\,\lambda\,w_t$
  \EndFor
  \end{algorithmic}
\end{algorithm*}

\begin{algorithm*}[ht]
  \caption{\texttt{COSMOS}}
  \label{alg:COSMOS}
  \begin{algorithmic}[1]

    \State For an $m \times n$ layer $W$, the algorithm maintain four matrices:
            $U \in \mathbb{R}^{n \times r}$, $S \in \mathbb{R}^{r \times r}$,
            $V \in \mathbb{R}^{m \times r}$, and $M \in \mathbb{R}^{m \times n}$ per layer.

    \State \textbf{Input:} Learning rate $\eta$, combination weight $\gamma$, 
          projection rank $r \ll n$, momentum parameters $(\beta_1,\beta_2)$, 
          perturbation parameter $\epsilon$. 
          For simplicity, we omit the initialization.

    \For{$t = 0, \ldots$}

        \State Sample batch $\mathcal{M}_t$

        \State $G_t \gets \nabla_W \phi_{\mathcal{M}_t}(W_t)$

        \State $M_t \gets \beta_1 M_{t-1} + (1 - \beta_1) G_t$

        \State $U_t \gets 
        \texttt{QR}\!\big(\beta_2 U_{t-1} S_{t-1} 
        + (1 - \beta_2) G_t^\top G_t U_{t-1}\big)$

        \State $S_t \gets 
        U_t^\top 
        \big(\beta_2 U_{t-1} S_{t-1} U_{t-1}^\top 
        + (1 - \beta_2) G_t^\top G_t\big)\,
        U_t$

        \State $V_t \gets 
        \beta_2 V_{t-1} 
        + (1 - \beta_2) (G_t U_t) \odot (G_t U_t)$

        \State $\displaystyle 
        A_t = 
        \left(
          \frac{
            M_t U_t / (1 - \beta_1^t)
          }{
            \sqrt{(V_t + \epsilon) / (1 - \beta_2^t)}
          }
        \right) 
        U_t^\top$

        \State $\displaystyle 
        B_t \gets 
        \texttt{NORM}\!\left(
          \texttt{NS5}\!\left(
            \frac{
              M_t - M_t U_t U_t^\top
            }{
              \|M_t - M_t U_t U_t^\top\|_{\mathrm{F}}
            }
          \right)
        \right)$

        \State $\displaystyle 
        \tilde{G}_t \gets 
        A_t + \gamma \cdot B_t \cdot \sqrt{m}$

        \State $\displaystyle 
        W_{t+1} \gets 
        W_t - \eta \cdot \texttt{NORM}(\tilde{G}_t) \cdot \sqrt{m}$

    \EndFor

  \end{algorithmic}
\end{algorithm*}

\begin{algorithm*}[ht]
  \caption{\texttt{GaLore}}
  \label{alg:GaLore}
  \begin{algorithmic}[1]

    \State For an $m \times n$ layer $W$, the algorithm maintains a rank-$r$ subspace projector
          (left or right, depending on the shape) and Adam moments in the projected space.

    \State \textbf{Input:} Learning rate $\eta$, scale factor $\alpha$,
          projection rank $r$, subspace change frequency $T$,
          Adam parameters $(\beta_1,\beta_2)$, numerical constant $\epsilon$.
          For simplicity, we omit the initialization.

    \For{$t = 0, 1, \ldots$}

        \State Sample batch $\mathcal{B}_t$

        \State $G_t \gets \nabla_W \phi_{\mathcal{B}_t}(W_t)$

        \If{$t \bmod T = 0$}
            \State $U,\Sigma,V \gets \texttt{SVD}(G_t)$
            \If{$m \le n$}
                \State $P_t \gets U_{[:,1:r]}$ \hfill \{left projector\}
            \Else
                \State $Q_t \gets V_{[:,1:r]}$ \hfill \{right projector\}
            \EndIf
        \Else
            \If{$m \le n$}
                \State $P_t \gets P_{t-1}$ \hfill \{reuse projector\}
            \Else
                \State $Q_t \gets Q_{t-1}$ \hfill \{reuse projector\}
            \EndIf
        \EndIf

        \If{$m \le n$}
            \State $R_t \gets P_t^\top G_t$ \hfill \{project to compact space, $R_t \in \mathbb{R}^{r\times n}$\}
        \Else
            \State $R_t \gets G_t Q_t$ \hfill \{project to compact space, $R_t \in \mathbb{R}^{m\times r}$\}
        \EndIf

        \State \textbf{UPDATE}$(R_t)$ by Adam in the projected space:
        \State $M_t \gets \beta_1 M_{t-1} + (1-\beta_1) R_t$
        \State $V_t \gets \beta_2 V_{t-1} + (1-\beta_2) (R_t \odot R_t)$
        \State $\hat{M}_t \gets M_t / (1-\beta_1^{t+1})$
        \State $\hat{V}_t \gets V_t / (1-\beta_2^{t+1})$
        \State $N_t \gets \hat{M}_t / (\sqrt{\hat{V}_t} + \epsilon)$

        \If{$m \le n$}
            \State $\tilde{G}_t \gets \alpha \cdot P_t N_t$ \hfill \{project back to original space\}
        \Else
            \State $\tilde{G}_t \gets \alpha \cdot N_t Q_t^\top$ \hfill \{project back to original space\}
        \EndIf

        \State $W_{t+1} \gets W_t - \eta \cdot \tilde{G}_t$

    \EndFor

  \end{algorithmic}
\end{algorithm*}

\begin{algorithm*}[ht]
  \caption{\texttt{Sophia}}
  \label{alg:Sophia}
  \begin{algorithmic}[1]

    \State \textbf{Input:} Parameters $\theta_1$, learning rates $\{\eta_t\}_{t=1}^T$,
          hyperparameters $\lambda, \gamma, \beta_1, \beta_2, \epsilon$,
          curvature estimator choice
          $\texttt{Estimator} \in \{\texttt{Hutchinson}, \texttt{Gauss\text{-}Newton\text{-}Bartlett}\}$.
    \State Set $m_0 \gets 0$, $v_0 \gets 0$, and $h_{1-k} \gets 0$.

    \For{$t = 1, \ldots, T$}

        \State Compute minibatch loss $L_t(\theta_t)$.

        \State $g_t \gets \nabla L_t(\theta_t)$.

        \State $m_t \gets \beta_1 m_{t-1} + (1-\beta_1)\, g_t$.

        \If{$t \bmod k = 1$}
            \State $\hat{h}_t \gets \texttt{Estimator}(\theta_t)$.
            \State $h_t \gets \beta_2 h_{t-k} + (1-\beta_2)\, \hat{h}_t$.
        \Else
            \State $h_t \gets h_{t-1}$.
        \EndIf

        \State $\theta_t \gets \theta_t - \eta_t \lambda \theta_t$ \hfill \{weight decay\}

        \State $\theta_{t+1} \gets \theta_t - \eta_t \cdot
        \texttt{clip}\!\Big(
            \frac{m_t}{\max\{\gamma \cdot h_t,\ \epsilon\}},\ 1
        \Big)$.

    \EndFor

  \end{algorithmic}
\end{algorithm*}

\end{document}